\documentclass[12pt]{article}
\usepackage[margin=1.25in]{geometry}
\usepackage[square,sort,comma,numbers]{natbib}
\usepackage[T1]{fontenc}
\usepackage{times}

\usepackage[utf8]{inputenc} 
\usepackage{amsmath}
\usepackage{amssymb}
\usepackage{amsthm}
\usepackage[T1]{fontenc}    
\usepackage{hyperref}       
\usepackage{url}            
\usepackage{booktabs}       
\usepackage{amsfonts}       
\usepackage{nicefrac}       
\usepackage{microtype}      
\usepackage{xcolor}         

\usepackage{amsmath,amsthm,amssymb}
\usepackage{graphicx}
\usepackage{cleveref}
\usepackage{subcaption}
\usepackage{cleveref}
\usepackage{comment}
\usepackage{bbm}
\usepackage{algorithm}
\usepackage{physics}

\usepackage[noend]{algpseudocode}
\usepackage{textcomp}
\usepackage{wrapfig}
\usepackage{float}

\usepackage{pgfplots}
\usepackage{multirow}
\pgfplotsset{compat=1.8}
\tikzset{elegant/.style={smooth,thick,samples=500,magenta}}

\theoremstyle{plain}
\newtheorem{theorem}{Theorem}[section]
\newtheorem{lemma}[theorem]{Lemma}
\newtheorem{claim}[theorem]{Claim}
\newtheorem{remark}[theorem]{Remark}
\newtheorem{corollary}[theorem]{Corollary}
\theoremstyle{definition}
\newtheorem{definition}[theorem]{Definition}

\newtheorem{assumption}[theorem]{Assumption}

\newtheorem{example}[theorem]{Example}

\crefname{assumption}{Assumption}{Assumptions}

\usepackage{amsmath,amsfonts,bm}

\def\floor#1{\lfloor #1 \rfloor}
\def\1{\bm{1}}

\def\cB{{\mathcal{B}}}

\def\veta{{\bm{\eta}}}

\def\vtheta{{\bm{\theta}}}

\def\va{{\bm{a}}}
\def\vb{{\bm{b}}}

\makeatletter
\newcommand{\ve}{\@ifnextchar\bgroup{\velong}{{\bm{e}}}}
\newcommand{\velong}[1]{{\bm{#1}}}
\makeatother

\def\vg{{\bm{g}}}

\def\vu{{\bm{u}}}
\def\vv{{\bm{v}}}
\def\vw{{\bm{w}}}
\def\vx{{\bm{x}}}
\def\vy{{\bm{y}}}
\def\vz{{\bm{z}}}

\def\mA{{\bm{A}}}

\def\mG{{\bm{G}}}

\def\mI{{\bm{I}}}

\def\mM{{\bm{M}}}
\def\mN{{\bm{N}}}

\def\mR{{\bm{R}}}
\def\mS{{\bm{S}}}
\def\mT{{\bm{T}}}
\def\mU{{\bm{U}}}
\def\mV{{\bm{V}}}

\def\mX{{\bm{X}}}
\def\mY{{\bm{Y}}}

\def\mSigma{{\bm{\Sigma}}}

\DeclareMathAlphabet{\mathsfit}{\encodingdefault}{\sfdefault}{m}{sl}
\SetMathAlphabet{\mathsfit}{bold}{\encodingdefault}{\sfdefault}{bx}{n}
\newcommand{\tens}[1]{\bm{\mathsfit{#1}}}

\def\tW{{\tens{W}}}
\def\tX{{\tens{X}}}

\def\gA{{\mathcal{A}}}

\def\gF{{\mathcal{F}}}

\def\gR{{\mathcal{R}}}

\def\gV{{\mathcal{V}}}

\def\sA{{\mathbb{A}}}

\def\sC{{\mathbb{C}}}

\def\sP{{\mathbb{P}}}

\def\sR{{\mathbb{R}}}

\newcommand{\E}{\mathbb{E}}
\newcommand{\R}{\mathbb{R}}

\newcommand{\KL}{D_{\mathrm{KL}}}

\DeclareMathOperator*{\argmax}{arg\,max}
\DeclareMathOperator*{\argmin}{arg\,min}

\newcommand{\wh}{\widehat}
\newcommand{\wt}{\widetilde}

\newcommand{\PP}{\mathbb{P}}

\renewcommand{\tilde}{\wt}
\renewcommand{\hat}{\wh}

\newcommand{\T}{{\cal T}}
\newcommand{\F}{{\cal F}}

\newcommand{\supp}{\mathrm{supp}}

\newcommand{\states}{\mathcal{S}}

\newcommand{\actions}{\mathcal{A}}

\newcommand{\mdps}{\mathcal{M}}

\newcommand{\hatvtheta}{\hat{\vtheta}}

\newcommand{\poly}{\mathrm{poly}}

\newcommand{\vecz}{\mathrm{vec}} 

\newcommand{\normsm}[1]{\| #1 \|}
\newcommand{\normtwo}[1]{\norm{#1}_2}
\newcommand{\norminf}[1]{\norm{#1}_\infty}
\newcommand{\normtwosm}[1]{\normsm{#1}_2}
\newcommand{\normF}[1]{\left\| #1 \right\|_{\mathrm{F}}}

\newcommand{\Ac}{\mathcal{A}}
\newcommand{\Fc}{\mathcal{F}}
\newcommand{\Reg}{\mathrm{Reg}}

\def\cA{\mathcal{A}}
\def\cC{\mathcal{C}}
\def\cM{\mathcal{M}}
\def\cN{\mathcal{N}}
\def\cX{\mathcal{X}}
\def\cT{\mathcal{T}}
\def\cP{\mathcal{P}}

\newcommand{\indict}{\mathbb{I}}

\title{Optimal Gradient-based Algorithms for Non-concave Bandit Optimization}

\author{
	Baihe Huang\textsuperscript{1$\ast$}, Kaixuan Huang\textsuperscript{2$\ast$}, Sham M. Kakade\textsuperscript{3,4$\ast$}, Jason D. Lee\textsuperscript{2$\ast$}\\ Qi Lei\textsuperscript{2$\ast$}, Runzhe Wang\textsuperscript{2$\ast$}, Jiaqi Yang\textsuperscript{5}\thanks{Alphabetical order. Correspondence to: Qi Lei, \url{qilei@princeton.edu}, Jason D. Lee, \url{jasonlee@princeton.edu}.}\\
	\\
	\textsuperscript{1}Peking University \quad \textsuperscript{2}Princeton University \quad \textsuperscript{3}University of Washington\\
	\textsuperscript{4}Microsoft Research \quad \textsuperscript{5}{Tsinghua University}
}

\begin{document}

\maketitle

\begin{abstract}
Bandit problems with linear or concave reward have been extensively studied, but relatively few works have studied bandits with non-concave reward. This work considers a large family of bandit problems where the unknown underlying reward function is non-concave, including the low-rank generalized linear bandit problems and two-layer neural network with polynomial activation bandit problem.
For the low-rank generalized linear bandit problem, we provide a minimax-optimal algorithm in the dimension, refuting both conjectures in \cite{lu2021low,jun2019bilinear}. Our algorithms are based on a unified zeroth-order optimization paradigm that applies in great generality and attains optimal rates in several structured polynomial settings (in the dimension). We further demonstrate the applicability of our algorithms in RL in the generative model setting, resulting in improved sample complexity over prior approaches.
Finally, we show that the standard optimistic algorithms (e.g., UCB) are sub-optimal by dimension factors. In the neural net setting (with polynomial activation functions) with noiseless reward, we provide a bandit algorithm with sample complexity equal to the intrinsic algebraic dimension. Again, we show that optimistic approaches have worse sample complexity, polynomial in the extrinsic dimension (which could be exponentially worse in the polynomial degree).
\end{abstract}

\section{Introduction}

Bandits~\citep{lattimore2020bandit} are a class of online decision-making problems where an agent interacts with the environment, only receives a scalar reward, and aims to maximize the reward. In many real-world applications, bandit and RL problems are characterized by large or continuous action space. To encode the reward information associated with the action, function approximation for the reward function is typically used, such as linear bandits~\cite{dani2008stochastic}.
Stochastic linear bandits assume the mean reward to be the inner product between the unknown model parameter and the feature vector associated with the action. This setting has been extensively studied, and algorithms with optimal regret are known~\citep{dani2008stochastic,lattimore2020bandit,abbasi2011improved,bubeck2012regret}.

However, linear bandits suffer from limited representation power unless the feature dimension is prohibitively large. A comprehensive empirical study \citep{riquelme2018deep} found that real-world problems required non-linear models and thus non-concave rewards to attain good performance on a testbed of bandit problems. To take a step beyond the linear setting, it becomes more challenging to design optimal algorithms. Unlike linear bandits, more sophisticated algorithms beyond optimism are necessary. For instance, a natural first step is to look at quadratic~\citep{katariya2017stochastic}  and higher-order polynomial~\citep{hao2020low} reward. In the context of phase retrieval, which is a special case for the quadratic bandit, people have derived algorithms that achieve minimax risks in the statistical learning setting~\citep{arias2012fundamental,lecue2013minimax,cai2016optimal}. However, the straightforward adaptation of these algorithms results in sub-optimal dimension dependency.

In the bandit domain, existing analysis on the nonlinear setting includes eluder dimension~\citep{russo2013eluder}, subspace elimination \citep{lu2021low,jun2019bilinear}, etc. Their results also suffer from a larger dimension dependency than the best known lower bound in many settings. (See Table \ref{tbl:main} and Section \ref{sec:main} for a detailed discussion of these results.)
Therefore in this paper, we are interested in investigating the following question:
\begin{center}
	{\em What is the optimal regret for non-concave bandit problems, including structured polynomials (low-rank etc.)? Can we design algorithms with optimal dimension dependency? }
\end{center}
\paragraph{Contributions:}
In this paper, we answer the questions and close the gap (in problem dimension) for various non-linear bandit problems. 

\begin{enumerate}
	\item First, we design stochastic zeroth-order gradient-like\footnote{Our algorithm estimates the gradient, but with some irreducible bias for the tensor case. Note that our algorithms converge linearly despite the bias. } ascent algorithms to attain minimax regret for a large class of structured polynomials. The class of structured polynomials contains bilinear and low-rank linear bandits and symmetric and asymmetric higher-order homogeneous polynomial bandits with action dimension $d$. Though the reward is non-concave, we combine techniques from two bodies of work, non-convex optimization and numerical linear algebra, to design robust gradient-based algorithms that converge to global maxima. Our algorithms are also computationally efficient, practical, and easily implementable.
	
	In all cases, our algorithms attain the optimal dependence on dimension $d$, which was not previously attainable using existing optimism techniques. As a byproduct, our algorithm refutes\footnote{Both papers conjectured regret of the form $O(\sqrt{d^3 \poly(k) T})$ based on convincing but potentially misleading heuristics. } the conjecture from \cite{jun2019bilinear} on the bilinear bandit, and the conjecture from \cite{lu2021low} on low-rank linear bandit by giving an algorithm that attains the optimal dimension dependence.
	
	\item We demonstrate that our techniques for non-concave bandits extend to RL in the generative setting, improving upon existing optimism techniques.
	
	\item When the reward is a general polynomial without noise, we prove that solving polynomial equations achieves regret equal to the intrinsic algebraic dimension of the underlying polynomial class, which is often linear in $d$ for interesting cases. In general, this complexity cannot be further improved.
	\item Furthermore, we provide a lower bound showing that all UCB algorithms have a sample complexity of $\Omega(d^p)$, where $p$ is the degree of the polynomial. The dimension of \textit{all homogeneous polynomials} of degree $p$ in dimension $d$ is $d^p$, showing that UCB is oblivious to the polynomial class and highly sub-optimal even in the noiseless setting.
	
\end{enumerate}

\subsection{Related Work}


\paragraph{Linear Bandits.} Linear bandit problems and their variants are studied in \citep{dani2008stochastic,lattimore2020bandit,abbasi2014linear,bubeck2012regret,abbasi2011improved,dani2008stochastic,rusmevichientong2010linearly,hao2020high,hao2021online}. The matching upper bound and minimax lower bound achieves $O(d \sqrt{T})$ regret. Structured linear bandits, including sparse linear bandit \citep{abbasi2012online} which developed an online-to-confidence-set technique. This technique yields the optimal $O(\sqrt{sd T})$ rate for sparse linear bandit. However \cite{lu2021low} employed the same technique for low-rank linear bandits giving an algorithm with regret $O(\sqrt{d^3 \poly(k) T})$ which we improve to $O(\sqrt{d^2 \poly(k) T})$, which meets the lower bound given in \cite{lu2021low}.

\paragraph{Eluder Dimension.} \cite{russo2013eluder} proposed the eluder dimension as a general complexity measure for nonlinear bandits. However, the eluder dimension is only known to give non-trivial bounds for linear function classes and monotone functions of linear function classes. For structured polynomial classes, the eluder dimension simply embeds into an ambient linear space of dimension $d^p$, where $d$ is the dimension and $p$ is the degree. This parallels the linearization/NTK line in supervised learning \citep{wei2019regularization,ghorbani2019linearized,allen2019can} which show that linearization also incurs a similarly large penalty of $d^p$ sample complexity, and more advanced algorithm design is need to circumvent linearization~\citep{bai2019beyond,chen2020towards,fang2020modeling,woodworth2019kernel,gao2019convergence,nacson2019lexicographic,ge2018learning,moroshko2020implicit,haochen2020shape,wang2020beyond,damian2021label}.


\paragraph{Neural Kernel Bandits.} 
 \cite{valko2013finite} initiated the study of kernelized linear bandits, showing regret dependent on the information gain. \citep{xu2020neural} specialized this to the Neural Tangent Kernel (NTK)~\citep{li2018learning,du2019gradient,jacot2018neural,gao2019convergence,bai2019beyond}, where the algorithm utilizes gradient descent but remains close to initialization and thus remains a kernel class. Furthermore NTK methods require $d^p$ samples to express a degree $p$ polynomial in $d$ dimensions~\cite{ghorbani2021linearized}, similar to eluder dimension of polynomials, and so lack the inductive biases necessary for real-world applications of  decision-making problems~\cite{riquelme2018deep}.
  
  \paragraph{Concave Bandits.} There has been a rich line of work on concave bandits starting with \cite{flaxman2004online,kleinberg2004nearly}. \cite{agarwal2011stochastic} attained the first $\sqrt{T}$ regret algorithm for concave bandits though with a large $\poly(d)$ dependence. In the adversarial setting, a line of work \cite{hazan2016optimal,bubeck2017kernel,lattimore2020improved} have attained polynomial-time algorithms with $\sqrt{T}$ regret with increasingly improved dimension dependence. The sharp dimension dependence remains unknown.

\paragraph{Non-concave Bandits.} To our knowledge, there is no general study of non-concave bandits, likely due to the difficulty of globally maximizing non-concave functions. A natural starting point of studying the non-concave setting are quadratic rewards such as the Rayleigh quotient, or namely bandit PCA~\cite{kotlowski2019bandit,garber2015online}. In the bilinear setting \citep{jun2019bilinear} and the low-rank linear setting \citep{lu2021low} can be made in the low rank (rank $k$) case to achieve $\tilde{O}(\sqrt{d^3 \poly(k) T})$ regret. Other literature consider related but different settings that are not comparable to our results \citep{katariya2017stochastic,johnson2016structured,gopalan2016low,lale2019stochastic}.
We note that the regret of all previous work is at least $O(\sqrt{d^3 T})$. This includes the subspace exploration and refinement algorithms from \cite{jun2019bilinear,lu2021low}, or from eluder dimension \citep{russo2013eluder}. Recently \citep{hao2020low} considers online problem with a low-rank tensor associated with  axis-aligned set of arms, which corresponds to finding the largest entry of the tensor. Finally, \cite{kotlowski2019bandit} study the bandit PCA problem in the adversarial setting, attaining regret of $O(\sqrt{d^3 T})$. We leave adapting our results to the advesarial setting as an open problem.



In the noiseless setting, there is some investigation in phase retrieval borrowing the tools from algebraic geometry (see e.g. \citep{wang2019generalized}). In this paper, we will study the bandit problem with more general reward functions: neural nets with polynomial activation (structured polynomials). \cite{kileel2019expressive} study similar structured polynomials, also using tools from algebraic geometry, but they only study the expressivity of those polynomials and do not consider the learning problems. \cite{dong2021provable} study noiseless bandits, but only attain local optimality.

In concurrent work, \cite{lattimore2021bandit} address the phase retrieval bandit problem which is equivalent to a symmetric rank $1$ variant of the bilinear bandit of \cite{jun2019bilinear} and attain $\tilde{O}(\sqrt{d^2 T})$ regret. Our work in Section \ref{sec:bilinear} specialized to the rank $1$ case attains the same regret.

\paragraph{Matrix/Tensor Power Method.} Our analysis stems from noisy power methods for matrix/tensor decomposition problems. Robust power method, subspace iteration, and tensor decomposition that tolerate noise first appeared in \citep{hardt2014noisy,anandkumar2014tensor}.
Follow-up work attained the optimal rate for both gap-dependence and gap-free settings for matrix decomposition \citep{musco2015randomized,allen2017first}. An improvement on the problem dimension for tensor power method is established in \citep{wang2016online}. \citep{sanjabi2019does} considers the convergence of tensor power method in the non-orthogonal case.


\section{Preliminaries}
\subsection{Setup: Structured Polynomial Bandit}
We study structured polynomial bandit problems where the reward function is from a class of structured polynomials (the precise settings and structures are discussed below). A player plays the bandit for $T$ rounds, and at each round $t\in[T]$, the player chooses one action $\va_{t}$ from the feasible action set $\Ac$ and receives the reward $r_t$ afterward.

We consider both the stochastic case where $r_t=f_{\vtheta}(\va_{t})+\eta_t$ where $\eta_t$ is the random noise, and the noiseless case $r_t=f_{\vtheta}(\va_{t})$.  Specifically the function $f_{\vtheta}$ is unknown to the player, but lies in a known function class $\Fc$. We use the notation $\vecz(\mM)$ to denote the vectorization of a matrix or a tensor $\mM$, and $\vv^{\otimes p}$ to denote the $p$-order tensor product of a vector $\vv$. For vectors we use $\|\cdot\|_2 $ or $\|\cdot\|$ to denote its $\ell_2$ norm. For matrices $\|\cdot\|_2$ or $\|\cdot\|$ stands for its spectral norm, and $\|\cdot\|_F$ is Frobenius norm. For integer $n$, $[n]$ denotes set $\{1,2,\cdots n\}$. For cleaner presentation, in the main paper we use $\tilde O$, $\tilde \Theta$ or $\tilde\Omega$ to hide universal constants, polynomial factors in $p$ and polylog factors in dimension $d$, error $\varepsilon$, eigengap $\Delta$, total round number $T$ or failure rate $\delta$.

We now present the outline with the settings considered in the paper:

\textbf{The stochastic bandit eigenvector case}, $\Fc_\text{EV}$, considers action set $\Ac=\{\va\in\R^d:\normtwosm{\va}\leq1\}$ as shown in Section \ref{sec:bilinear}, and
\[\Fc_{\text{EV}}=\left\{\begin{array}{l}
f_{\vtheta}(\va) = \va^T \mM\va, \mM=\sum_{j=1}^k \lambda_j\vv_j\vv_j^\top, \text{ for orthonormal }\vv_j  \\
\mM\in\R^{d\times d}, 1\geq \lambda_1 \geq |\lambda_2|\geq \cdots \geq |\lambda_k|
\end{array}\right\}.
\]

\textbf{The stochastic low-rank linear reward case}, $\Fc_{\text{LR}}$ considers action sets on bounded matrices $\Ac=\{\mA\in\R^{d\times d}:\normF{\mA}\leq1\}$ in Section \ref{sec:low-rank}, and
\[\Fc_{\text{LR}}=\left\{\begin{array}{l}
f_{\vtheta}(\mA) = \langle\mM, \mA\rangle = \vecz(\mM)^\top \vecz(\mA),\\
\mM\in\R^{d\times d}, \rank(\mM)=k, \mM=\mM^\top,  \normF{\mM}\leq 1
\end{array}\right\}.
\]

We illustrate how to apply the established bandit oracles to attain a better sample complexity for RL problems with the simulator in Section \ref{sec:bellmen}.

\textbf{The stochastic homogeneous polynomial reward case } is presented in Section \ref{sec:tensor}. For the symmetric case in Section \ref{sec:symmetric_tensor}, the action sets are $\Ac=\{\va\in\R^d:\normtwosm{\va}\leq1\}$, and
\[\Fc_{\text{SYM}}=\left\{
\begin{array}{l}
f_{\vtheta}(\va) = \sum_{j=1}^k \lambda_j (\vv_j^\top\va)^p \text{ for orthonormal } \vv_j,\\
1\geq r^*=\lambda_1 > |\lambda_2|\geq\cdots\geq |\lambda_k|
\end{array}\right\};
\]
in the asymmetric case in Section \ref{sec:asymmetric_tensor}, the action sets are \\$\Ac=\{\va=\va(1)\otimes \va(2)\otimes\cdots\otimes \va(p)\in\R^{d^p}: \forall q\in [p], \normtwosm{\va(q)}\leq 1\}$, and
\[\Fc_{\text{ASYM}}=\left\{
\begin{array}{l}
f_{\vtheta}(\va) = \sum_{j=1}^k \lambda_j \prod_{q=1}^p (\vv_j(q)^\top\va(q)) \text{ for orthonormal } \vv_j(q)\text{ for each }q,\\
1\geq r^*=|\lambda_1|\geq |\lambda_2|\geq\cdots\geq |\lambda_k|
\end{array}\right\}.
\]
In the above settings, there is stochastic noise on the observed rewards. We also consider noiseless settings as below:\\
\textbf{The noiseless polynomial reward case} is presented in Section \ref{sec:noiseless}. The action sets $\Ac$ are subsets of $\R^{d}$, and
\[\Fc_{\text{P}}=\left\{\begin{array}{l}
f_{\vtheta}(\va) = \left\langle \vtheta, \widetilde{\va}^{\otimes p}\right \rangle: \vtheta \in \mathcal{V},    \widetilde{\va} = [1, \va^\top]^\top, \mathcal{V}\subseteq (\mathbb{R}^{d+1})^{\otimes p} \text{ is an algebraic variety}
\end{array}\right\}.
\]
Additionally, $\Fc$ needs to be admissible (\Cref{def:admissible}). This class includes two-layer neural networks with polynomial activations (i.e. structured polynomials). We study the fundamental limits of all UCB algorithms in Section \ref{sec:noiseless:lb} as they are $\Omega(d^{p-1})$ worse than our algorithm presented in Section \ref{sec:noiseless:ub}.

In the above settings, we are concerned with the cumulative regret $ \mathfrak{R}(T)$ for $T$ rounds. Let $f^*_\vtheta=\sup\limits_{\va\in\Ac}f_\vtheta(\va)$,

$$ \mathfrak{R}_\vtheta (T):=\sum\limits_{t=1}^T (f^*_\vtheta - f_\vtheta(\va_t))$$

And since the parameters can be chosen adversarially, we are bounding $\mathfrak{R}(T)=\sup_\vtheta \mathfrak{R}_\vtheta (T)$ in this paper. 

In all the stochastic settings above, we make the standard assumption on stochasticity that $\eta_t$ is conditionally zero-mean 1-sub-Gaussian random variable regarding the randomness before $t$.

\subsection{Warm-up: Adapting Existing Algorithms}

In all of the above settings, the function class can be viewed as a generalized linear function in kernel spaces. Namely, there is fixed feature maps $\psi,\phi$ so that $f_\vtheta(\va)=\psi(\vtheta)^T\phi(\va)$. Thus it is straightforward to adapt linear bandit algorithms like the renowned LinUCB \citep{li2010contextual} to our settings. Furthermore, another baseline is given by the eluder dimension argument \cite{russo2013eluder}\citep{wang2020reinforcement} which gives explicit upper bounds for general function classes. We present the best upper bound by adapting these methods as a baseline in Table \ref{tbl:main}, together with our newly-derived lower bound and upper bound in this paper.

The best-known statistical rates are based on the following result.
\begin{theorem}[Proposition 4 in \cite{russo2013eluder}]\label{thm:prep}
	With $\alpha=O(T^{-2})$ appropriately small, given the $\alpha$-covering-number $N$ (under $\norminf{\cdot}$) and the $\alpha$-eluder-dimension $d_E$ of the function class $\Fc$, Eluder UCB (\Cref{alg:linUCB}) achieves regret $\tilde{O}(\sqrt{ d_E T \log N })$.
\end{theorem}
In the first row of Table \ref{tbl:main}, we further elaborate on the best results obtained from Theorem \ref{thm:prep} in individual settings. More details can be found in Appendix \ref{appendix:prelim}.

\section{Main results}
\label{sec:main}

\begin{table}[]
	\centering
	\resizebox{\textwidth}{!}{
		\begin{tabular}{|c|c|c|c|c|c|c|}
			\hline
			\multicolumn{3}{|c|}{Regret} & $\Fc_{\text{SYM}}$ & $\Fc_{\text{ASYM}}$ & $\Fc_{\text{EV}}$ & $\Fc_{\text{LR}}$ \\
			\hline
			\multicolumn{3}{|c|}{LinUCB/eluder} & $
			\sqrt{d^{p+1}kT} $ & $\sqrt{d^{p+1}k T}$ & $\sqrt{d^{3}kT}$ & $\sqrt{d^{3}kT}$ \\
			\hline
			\multirow{4}{*}{Our Results} &  \multirow{2}{*}{NPM} & Gap & N/A & N/A & $\sqrt{\kappa^3 d^{2}T}$ & $\sqrt{ d^{2}k\lambda_k^{-2}T}$ \\\cline{3-7}
			& & Gap-free & $\sqrt{d^{p}k T}$  & $\sqrt{ k^p d^pT}$ & $k^{4/3}(dT)^{2/3} $ & $(dk T)^{2/3}$\\\cline{2-7}
			& \multicolumn{2}{|c|}{Lower Bound} & $\sqrt{d^pT}$ &$\sqrt{d^pT}$ & $\sqrt{d^2T} $& $\sqrt{d^2 k^2 T}$ (*) \\
			\hline
		\end{tabular}
	}
	\vspace{3mm}
	\caption{Baselines and our main results (for stochastic settings).  Eigengap $\Delta=\lambda_1-|\lambda_2|, $ condition number $\kappa=\lambda_1/\Delta$. The result with $*$ is from \cite{lu2021low}. For simplicity, in this table, we treat $p$ and $r^*:=\max_{\va\in \cA} f_{\vtheta}(\va)$ as constant and ignore all $\poly\log$ factors. 
	}
	\label{tbl:main}
	\vspace{-8pt}
\end{table}

We now present our main results. We consider four different stochastic settings (see Table~\ref{tbl:main}) and one noiseless setting with structured polynomials.

In the cases of stochastic reward, all our algorithms can be unified as gradient-based optimization. At each stage with a candidate action $\va$, we define the estimator $ G_n(\va):= \frac{1}{n}\sum_{i=1}^n(f_{\vtheta}((1-\zeta) \va+\zeta\vz_i)+\eta_i)\vz_i$, with $\vz_i \sim \cN(0,\sigma^2 \mI_d)$ and proper step-size $\zeta$~\cite{flaxman2004online}. Therefore $\E_{\vz}[G(\va)] = \sigma^2 \zeta \nabla f_{\vtheta}((1-\zeta)\va) + O(\zeta^2)=\zeta(1-\zeta)^{p-1}\sigma^2 \nabla f_{\vtheta}(\va) +O(\zeta^2)$ for $p$-th order homogeneous polynomials. Therefore with enough samples, we are able to implement noisy gradient ascent with bias.

In the noiseless setting, our algorithm solves for the parameter $\vtheta$ with randomly sampled actions $\{\va_t\}$ and the noiseless reward $\{ f_\vtheta(\va_t) \}$, and then determines the optimal action by computing $\arg\max_\va f_\vtheta (\va)$.

\subsection{Stochastic Eigenvalue Reward ($\mathcal{F}_{\text{EV}}$)}
\label{sec:bilinear}
Now consider bandits with stochastic reward $r(\va) = \va^\top \mM \va +\eta$ with action set $\cA=\{\va| \|\va\|_2\leq 1\}$, where $\mM=\sum_{i=1}^r \lambda_i \vv_i\vv_i^\top$ is symmetric and satisfies $ r^* = \lambda_1 > |\lambda_2| \geq |\lambda_3|\cdots \geq |\lambda_r|, \eta\sim \cN(0,1)$.\\
Denote by $\va^*$ the optimal action ($\pm \vv_1$), the leading eigenvector of $\mM$, $(\va^*)^\top \mM\va^*=\lambda_1$.  Let $\Delta=\lambda_1-|\lambda_2| > 0$ be the eigengap and $\kappa:=\lambda_1/\Delta$ be the condition number.
\begin{remark}[Negative leading eigenvalue]
	\label{remark:negative_leading_eigenvalue}
	For a symmetric matrix $\mM$, we will conduct noisy power method to recover its leading eigenvector, and therefore we require its leading eigenvalue $\lambda_1$ to be positive. It is straightforward to extend to the setting where the nonzero eigenvalues satisfy: $r^*\equiv \lambda_1>\lambda_2\geq \lambda_l >0 >\lambda_{l+1}\cdots \geq \lambda_k,$ and $|\lambda_k|>\lambda_1$. For this problem, we can shift $\mM$ to get $\mM+|\lambda_k|\mI$ and the eigen-spectrum now becomes $\lambda_1+|\lambda_k|, \lambda_2+|\lambda_k|,\cdots 0$; therefore, we can still recover the optimal action with dependence on the new condition number $ (\lambda_1+|\lambda_k|)/(\lambda_1-\lambda_2)$. 	
\end{remark}
\begin{remark}[Asymmetric matrix]
	\label{remark:asymmetric_reduction}
	Our algorithm naturally extends to the asymmetric setting: $f(\va_1,\va_2) = \va_1^\top \tilde \mM \va_2$, where $\tilde \mM = \mU\mSigma \mV^\top$. This setting can be reduced to the symmetric case via defining 
	$$\mM=\begin{bmatrix}0 & \tilde \mM^\top \\ \tilde \mM& 0 \end{bmatrix}=\frac{1}{2}\begin{bmatrix} \mV & \mV \\ \mU & -\mU\end{bmatrix}\begin{bmatrix} \mSigma& 0 \\0& -\mSigma \end{bmatrix} \begin{bmatrix} \mV& \mV \\ \mU&-\mU\end{bmatrix}^\top, $$ which is a symmetric matrix, and its eigenvalues are $\pm \sigma_i (\tilde\mM)$, the singular values of $\tilde \mM$. Therefore our analysis on symmetric matrices also applies to the asymmetric setting and will equivalently depend on the gap between the top singular values of $\tilde \mM$. A formal asymmetric to symmetric conversion algorithm is presented in Algorithm 1 in \cite{garber2015online}.
\end{remark}

\textbf{Algorithm.} We note that by conducting zeroth-order gradient estimate $1/n\sum_{i=1}^{n}(f(\va/2+\vz_i/2)+\eta_i)\vz_i$ with step-size $1/2$~\cite{flaxman2004online} and sample size $n$, we get an estimate for $\E_{\eta,\vz}[(f(\va/2+\vz/2)+\eta)\vz] = \frac{\sigma^2}{2} \mM\va $ when $\vz \sim \cN(0,\sigma^2)$. Therefore we are able to use noisy power method to recover the top eigenvector. We present a more general result with Tensor power method in Algorithm \ref{alg:npm} and attain a gap-dependent risk bound: 

\begin{algorithm}
	\caption{Noisy power method for bandit eigenvalue problem.}
	\label{alg:npm}	
	\begin{algorithmic}[1]
		\State {\bf \underline{Input:} }  Quadratic function $f:\cA\rightarrow \R$ with noisy reward, failure probability $\delta$, error $\varepsilon$.
		\State {\bf \underline{Initialization:} } Initial action $\va_0 \in\R^{d}$ randomly sampled on the unit sphere $\mathbb{S}^{d-1}$. We set $\alpha = |\lambda_2/\lambda_1|$, sample size per iteration $n=C_nd^2\log(d/\delta)(\lambda_1 \alpha)^{-2}\varepsilon^{-2} $, sample variance $m=C_m d\log(n/\delta)$, 
		total iteration $L=\lfloor C_L \log(d/\varepsilon) \rfloor + 1$.
		\For {Iteration $l$ from $1$ to $L$}
		\State Sample $\vz_i\sim \cN(0,1/m I_d), i=1,2,\cdots n$. (Re-sample the whole batch if exists $\vz_i$ with norm greater than 1.)		
		\State {\bf \underline{Noisy power method:} }
		\State Take actions $\tilde \va_i = \frac{\va_{l-1} + \vz_i}{2}$ and observe $r_i = f(\va_i)+\eta_i, \forall i \in [n] $
		\State Update normalized action $\va_l \leftarrow \frac{m}{n}\sum_{i=1}^n r_i \vz_i$, and normalize $\va_{l} \leftarrow \va_l/\|\va_l\|_2$.
		\EndFor
		\State {\bf \underline{Output:}} $\va_{L}$.
	\end{algorithmic}
\end{algorithm}

\begin{theorem}[Regret  bound for noisy power method (NPM)]
	\label{thm:bilinear_npm}
	In Algorithm \ref{alg:npm}, we set $\varepsilon\in(0,1/2)$, $\delta=0.1/(L_0S)$ and let $C_L,C_S,C_n,C_m$ be large enough universal constants. Then with high probability $0.9$ we have: the output $\va_L$ satisfies $\tan\theta(\va^*,\va_L)\leq \varepsilon$ and yields $r^*\varepsilon^2$-optimal reward; and the total number of actions we take is $\tilde O(\frac{\kappa d^2}{\Delta^2\varepsilon^2})$. 
By explore-then-commit (ETC) the cumulative regret is at most $\tilde O(\sqrt{\kappa^3 d^2 T})$.

\end{theorem}
All proofs in this subsection are in Appendix \ref{appendix:bandit_eigenvector}.
\begin{remark}[Intuition of \cite{jun2019bilinear} and how to overcome the conjectured lower bound via the design of adaptive algorithms] Let us consider the rank $1$ case of $r(\va) = (\va^T \vtheta^*)^2 +\eta$. A random action $\va \sim \text{Unif}( \mathbb{S}^{d-1}) $ has $f(\va) \asymp {1/d^2}$, and the noise has standard deviation $O(1)$. Thus the signal-to-noise-ratio is $O(1/d^2)$ and the optimal action $\theta^*$ requires $d$ bits to encode. If we were to play non-adaptively, this would require $O(d^3)$ queries and result in regret $\sqrt{d^3T}$ which matches the result of \cite{jun2019bilinear}.
	
	To go beyond this, we must design algorithms that are adaptive, meaning the information in $f(\va) +\eta$ is \textbf{strictly larger than} $\frac{1}{d^2}$. As an illustration of why this is possible, consider batching the time-steps into $d$ stages so that each stage decode $1$ bit of $\vtheta^\ast$. At the first stage, random exploration $\va \sim \text{Unif}( \mathbb{S}^{d-1}) $ gives signal-to-noise-ratio $O(1/d^2)$. Suppose $k$ bits of $\vtheta$ are decoded at $k$-th stage by $\hat \vtheta$, adaptive algorithms can boost the signal-to-noise-ratio to $O(k/d^2)$ by using $\hat \vtheta$ as bootstrap (e.g. exploring with $\hat \vtheta \pm a$ where $a$ is random exploration in the unexplored subspace). In this way adaptive algorithms only need $d^2/k$ queries in ($k+1$)-th stage and so the total number of queries sums up to $\sum_{k=1}^d d^2/k \approx d^2 \log d$.

	Gradient descent and power method offer a computationally efficient and seamless way to implement the above intuition. For every iterate action $\va$, we estimate $\mM \va$ from noisy observations and take it as our next action $\va^+$. With $d^2/(\Delta^2\varepsilon^2)$ samples, noisy power method enjoys linear progress $\tan\theta(\va^+, \va^*) \leq \max\{ c \tan(\va,\va^*), \varepsilon\} $, where $c<1$ is a constant that depends on $\lambda_1,\lambda_2,$ and $\varepsilon$. Therefore even though every step costs $d^2$ samples, overall we only need logarithmic (in $d,\varepsilon,\lambda_1,\lambda_2$) iterations to find an $\varepsilon$-optimal action.   
\end{remark}

\begin{remark}[Connection to phase retrieval and eluder dimension]
	For rank-1 case  $\mM=\vx\vx^\top$, the bilinear bandits can be viewed as phase retrieval, where one observes $y_r = (\va_r^\top \vx)^2 = \va_r^\top \mM \va_r $ plus some noise $\eta_r \sim \cN(0,\sigma^2)$. The optimal (among non-adaptive algorithms) sample complexity to recover $\vx$ is $\sigma^2 d/\epsilon^2$ \cite{candes2015phase,cai2016optimal} where they play $\va$ from random Gaussian $\cN(0, \mI)$. However, in bandit, we need to set the variance of $\va$ to at most $1/d$ to ensure $\|\va\|\leq 1$. Our problem is equivalent to observing $y_r/d$ where their $\va/\sqrt{d}\sim \cN(0,1/d \mI)$ and noise level $\eta_r/d\sim \cN(0,1)$, i.e., $\sigma^2=d^2$. Therefore one gets $ d^3/\epsilon^2$ even for the rank-1 problems. On the other hand, for all rank-1 $\mM$ the condition number $\kappa=1$; and thus our results match the lower bound (see Section \ref{sec:lb}) up to logarithmic factors, and also have fundamental improvements for phase retrieval problems by leveraging adaptivity.
	
	For UCB algorithms based on eluder dimension, the regret upper bound is $O(\sqrt{d_E \log(N)T}) = O( \sqrt{d^3T}) $ as presented in Theorem \ref{thm:prep}, where the dependence on $d$ is consistent with \citep{jun2019bilinear,lu2021low} and is non-optimal.
\end{remark}

The previous result depends on the eigen-gap. When the matrix is ill-conditioned, i.e., $\lambda_1$ is very close to $\lambda_2$, we can obtain gap-free versions with a modification: The first idea stems from finding higher reward instead of recovering the optimal action. Therefore when $\lambda_1$ and $\lambda_2$ are very close (gap being smaller than desired accuracy $\varepsilon$), it is acceptable to find any direction in the span of $(\vv_1,\vv_2)$. More formally, we care about the convergence speed of identifying any action in the space spanned by any top eigenvectors (whose associated eigenvalues are higher than $\lambda_1-\varepsilon$). Therefore the convergence speed will depend on $\varepsilon$ instead of $\lambda_1-\lambda_2$.
\begin{corollary}[Gap-free regret bound]
	\label{lemma:gap_free_matrix}
	For positive semi-definite matrix $\mM$, by setting $\alpha = 1-\varepsilon^2/2$ in Algorithm \ref{alg:npm} and performing ETC afterwards, one can obtain cumulative regret  of $ \tilde O( \lambda_1^{3/5} d^{2/5}  T^{4/5} )$.
\end{corollary}
Again, the PSD assumption is not essential. For general symmetric matrices with $\lambda_1\geq \lambda_2\geq\cdots > 0 > \cdots \lambda_k. $ We can still conduct shifted power method on $\mM -\lambda_k \mI$, yielding a cumulative regret of $ (\lambda_1 + |\lambda_k|)^{3/5} d^{2/5}T^{4/5}$.

Another novel gap-free algorithm requires to identify any top eigenspace $\mV_{1:l}, l\in[k]$: $\mV_{1:l}$ is the column span of $\{\vv_1,\vv_2,\cdots \vv_l\}$. Notice that in traditional subspace iteration, the convergence rate of recovering $\mV_l$ depends on the eigengap $\Delta_l:=|\lambda_l|-|\lambda_{l+1}| $. Meanwhile, since $\sum_{l=1}^k \Delta_l = \lambda_1$, at least one eigengap is larger or equal to $\lambda_1/k$. Suppose $\Delta_{l^*}\geq \lambda_1/k$; we can, therefore, set $\alpha=1-1/k$ and recover the top $l^*$ subspace up to $\lambda_1 \epsilon$ error, which will give an $\epsilon$-optimal reward in the end. We don't know $l^*$ beforehand and will try recovering the top subspace $\mV_1,\mV_2,\cdots \mV_k$  respectively, which will only lose a $k$ factor. With the existence of $l^*$, at least one trial (on recovering $\mV_{l^*}$) will be successful with the parameters of our choice, and we simply output the best action among all trials.

\begin{theorem}[Informal statement: (gap-free) subspace iteration]
	\label{lemma:bilinear_subspace_iteration}
	By running subspace iteration (Algorithm \ref{alg:gap-free_subspace_iteration}) with proper choices of parameters, we attain a cumulative regret of $\tilde O(\lambda_1^{1/3}k^{4/3} (dT)^{2/3})$. Algorithm \ref{alg:gap-free_subspace_iteration}) with another set of parameters can also recover the whole eigenspace, and achieve cumulative regret of $\tilde O((\lambda_1k)^{1/3} (\tilde\kappa dT)^{2/3})$, where $\tilde \kappa=\lambda_1/|\lambda_k|$.
\end{theorem}

\begin{table}
	\resizebox{\textwidth}{!}{
		\begin{tabular}{|c||c|c|c|c|c|}
			\hline
			$\mathcal{F}_\text{EV}$ &  LB ($k=1$)  & \cite{jun2019bilinear} & NPM & Gap-free NPM & Subspace Iteration \\
			\hline
			Regret		 &  {\color{red} $\sqrt{d^2T}$} &  $\sqrt{d^3 k\lambda_k^{-2}T}$ & {\color{red} $\sqrt{\kappa^3 d^2 T}$ } & {\color{red} $ d^{2/5}  T^{4/5}  $ } &  {\color{red} $ \min(				
				k^{4/3} (dT)^{2/3}, k^{1/3}(\tilde\kappa dT)^{2/3})  $ }\\
			\hline
		\end{tabular}
	}
	\resizebox{\textwidth}{!}{
		\begin{tabular}{|c||c|c|c|}
			\hline
			$\mathcal{F}_\text{LR}$  &
			LB (\cite{lu2021low}) & UB (\cite{lu2021low}) & Subspace Iteration \\
			\hline
			Regret & $ \Omega( \sqrt{d^2k^2T}  )$  & $ \sqrt{d^3kT} ^*$  or $\sqrt{d^3k\lambda_k^{-2} T}$  & {\color{red} $\min( \sqrt{d^2k\lambda_k^{-2}T}, (d k T)^{2/3}) $ } \\
			\hline
	\end{tabular} }
	\caption{Summary of results for quadratic reward. All red expressions are our results. LB, UB, NPM stands for lower bound, upper bound, and noisy power method respectively. 
		$\Delta = \lambda_1-|\lambda_2|$ is the eigengap, and $\kappa=\lambda_1/\Delta$, $\tilde \kappa=\lambda_1/|\lambda_k|$ are the condition numbers. The result with $*$ is not computationally tractable. For low-rank setting in this table, we treat $\|\mM\|_F$ as a constant for simplicity and leave its dependence in the theorems.
		Our upper bounds match the lower bound in terms of dimension and substantially improve over existing algorithms that are computationally efficient. }
	\label{tab:quadratic}
\end{table}

\subsection{Stochastic Low-rank Linear Bandits ($\mathcal{F}_{\text{LR}})$}
\label{sec:low-rank}
In the low-rank linear bandit, the reward function is $f(\mA) =\langle \mA, \mM\rangle$, with noisy observations $ r_t =  f(\mA)+\eta_t$, and the action space is  $\{\mA\in \mathbb{R}^{d \times d}: \|\mA\|_F \le 1\}$. Without loss of generality we assume $k$, the rank of $\mM$ satisfies $k\leq \frac{d}{2}$, since when $k$ is of the same order as $d$, the known upper and lower bound are both $\sqrt{d^2k^2 T} = \Theta(\sqrt{d^4 T})$ \citep{lu2021low} and there is no room for improvement. We write $r^*=\|\mM\|_F\leq 1$, therefore the optimal action is $\mA^*=\mM/r^*$. In this section, we write $\mX(s)$ to be the $s$-th column of any matrix $\mX$.

As presented in Algorithm \ref{alg:subspace_iteration}, we conduct noisy subspace iteration to estimate the right eigenspace of $\mM$.  Subspace iteration requires calculating $\mM \mX_t$ at every step. This can be done by considering a change of variable of $g(\mX):=f(\mX\mX^\top) =\langle \mX\mX^\top, \mM\rangle$ whose gradient\footnote{Directly performing projected gradient descent on $f(\mA)$ would not work, since this is not an adaptive algorithm as the gradient of a linear function is constant. This would incur regret $\sqrt{d^3 \poly(k) T}$. } is $\nabla g(\mX) =2 \mM \mX$. The zeroth-order gradient estimator can then be employed to stochastically estimate $\mM \mX$. We instantiate the analysis with symmetric $\mM$ while extending to asymmetric setting is straightforward since the problem can be reduced to symmetric setting (suggested in Remark \ref{remark:asymmetric_reduction}).

With stochastic observations and randomly sampled actions, we achieve the next iterate $\mY_l$ that satisfies $\mM \mX_{l-1}\equiv \E[\mY_l]$ in Algorithm \ref{alg:subspace_iteration}.
Let $\mM=\mV \mSigma \mV^\top$.
With proper concentration bounds presented in the appendix, we can apply the analysis of noisy power method \citep{hardt2014noisy} and get: 
\begin{algorithm}
	\caption{Subspace Iteration Exploration for Low-rank Linear Reward.}
	\label{alg:subspace_iteration}	
	\begin{algorithmic}[1]
		\State {\bf \underline{Input:} }  Quadratic function $f:\cA\rightarrow \R$ with noisy reward, failure probability $\delta$, error $\varepsilon$.
		\State {\bf \underline{Initialization:} } Set $k'=2k$. Initial candidate matrix $\mX_0 \in\R^{d\times k'}$, $\mX_0(j)\in \R^{d}, j=1,2,\cdots k'$ is the $j$-th column of $\mX_0$ and are i.i.d sampled on the unit sphere $\mathbb{S}^{d-1}$ uniformly. Sample variance $m$, 
		\# sample per iteration $n$, 
		total iteration $L$.
		\For {Iteration $l$ from $1$ to $L$}
		\State Sample $\vz_i\sim \cN(0,1/m I_d), i=1,2,\cdots n$.		
		\For {$s$ from $1$ to $k'$}
		\State {\bf \underline{Noisy subspace iteration:} }
		\State Calculate tentative rank-1 actions $\tilde \mA_i = \mX_{l-1}(s)\vz_i^\top $.
		\State Conduct estimation $\mY_l(s) \leftarrow m/n \sum_{i=1}^{n} (\langle \mM,\tilde \mA_i\rangle +\eta_{i,s})\vz_i $. ($\mY_l\in \R^{d\times k'}$)
		\EndFor
		\State Let $\mY_l = \mX_l \mR_l$ be a QR-factorization of $\mY_l$
		\State Update target action $\mA_l\leftarrow \mY_l\mX_l^\top $.
		\EndFor
		\State {\bf \underline{Output:}} $\hat\mA = \mA_L/\|\mA_L\|_F$
	\end{algorithmic}
\end{algorithm}

\begin{theorem}[Informal statement; sample complexity for low-rank linear reward]
	\label{thm:low-rank_subspace_iteration}
	With Algorithm \ref{alg:subspace_iteration}, $\mX_L$ satisfies $\|(\mI -\mX_L\mX_L^\top)\mV\| \leq \varepsilon/4$, and output $\hat\mA$ satisfies $\|\hat \mA-\mA^*\|_F\leq \varepsilon \|\mM\|_F $ with sample size $\tilde O(d^2k\lambda_k^{-2}\varepsilon^{-2}) $.
\end{theorem}
We defer the proofs for low-rank linear reward in Appendix \ref{appendix:low-rank}. 
\begin{corollary}[Regret bound for low-rank linear reward]
	We first call Algorithm \ref{alg:subspace_iteration} with $\varepsilon^4 = \tilde\Theta(d^2k\lambda_k^{-2}T^{-1} )$ to obtain $\hat\mA$; we then play $\hat\mA$ for the remaining steps. The cumulative regret satisfies
	$$ \mathfrak{R}(T) \leq \tilde O(\sqrt{ d^2 k (r^*)^2 \lambda_k^{-2} T }), $$
	with high probability $0.9.$ 
\end{corollary}
To be more precise, we need
$T\geq \tilde\Theta(d^2k/\lambda_k^2)$ for Algorithm \ref{alg:subspace_iteration} to take sufficient actions; however, the conclusion still holds for smaller $T$. Since simply playing $0$ for all $T$ actions will give a sharper bound of $\mathfrak{R}(T) \leq r^*T\leq \tilde O(r^* \sqrt{  d^2k\lambda_k^{-2} T})$). For cleaner presentation, we won't stress this for every statement.
	
\begin{proof}
	The corollary uses a special property of the strongly convex action set that ensures: $\mA^* = \mM/r^*$. With $\hat\mA$ that satisfies $\|\hat\mA\|_F=1$, we have
	\begin{align}
	\notag
	r^*- f_{\mM}(\mA) = & r^* - \langle \hat\mA, \mM\rangle =r^* - \langle \hat\mA, r^* \mA^* \rangle\\
	\notag
	= & \frac{r^*}{2}(2-2\langle \hat\mA, \mA^*\rangle) = \frac{r^*}{2}(\|\hat\mA\|_F^2 + \|\mA^*\|_F^2 - \langle \hat\mA, \mA^*\rangle)\\
	\label{eqn:low_rank_action2regret}
	= & \frac{r^*}{2}\|\hat\mA - \mA^*\|_F^2 \leq \frac{r^* \varepsilon^2}{2}
	\end{align}
	Therefore, with first $T_1 = \tilde O(d^2k\lambda_k^{-2}\varepsilon^{-2})$ exploratory samples we get $r^*- f(\hat\mA) \leq r^*\varepsilon^2/2 = r^* \sqrt{\frac{d^2k}{\lambda_k^2T}} =  \sqrt{\frac{(r^*)^2 d^2k}{\lambda_k^2T}} $.  Together we have:
	\begin{align*}
	\mathfrak{R}(T) = &  \sum_{t=1}^{T_1} r^* - f(\mA_t) + \sum_{t=T_1+1}^T r^* - f(\hat\mA) \\
	< & r^* T_1 + T r^*\varepsilon^2\\
	\leq & \tilde O(\sqrt{ d^2 k (r^*)^2 \lambda_k^{-2} T }).
	\end{align*}

\end{proof}
Notice that $k\leq \|\mM\|_F^2/\lambda_k^2 \leq k \tilde \kappa^2$ is order $k$. Thus for well-conditioned matrices $\mM$, our upper bound of $\tilde O(\sqrt{d^2 k \|\mM\|_F^2/\lambda_k^2 T})$ matches the lower bound $ \sqrt{d^2k^2T}$ except for logarithmic factors. 

In the previous setting with the bandit eigenvalue problem, estimating $\mM$ up to an $\epsilon$-error (measured by \textbf{operator norm}) gives us an $\epsilon$-optimal reward. Therefore the sample complexity for eigenvalue reward with similar subspace iteration is $\|\mM\|_2^2 d^2k/(\lambda_k^2\epsilon^2)$. In this section, on the other hand, we need \textbf{Frobenius norm} bound $\|\mA_L-\mA^*\|_F\leq \epsilon$; naturally the complexity becomes $\|\mM\|_F^2 d^2k/(\lambda_k^2\epsilon^2)$.

\begin{theorem}[Regret bound for low-rank linear reward: gap-free case]
	\label{thm:gap-free_low-rank}
	Set $\varepsilon^6 = \Theta(\frac{d^2k^2}{(r^*)^2 T})$, 	$n= \tilde\Theta(\frac{d^2k^2 }{ (r^*)^2 \varepsilon^4} ),L=\Theta(\log(d/\varepsilon) )$ and $k'=2k$ in Algorithm \ref{alg:subspace_iteration} and get $\hat\mA$. Then we play it for the remaining steps, the cumulative regret satisfies:
	$$ \mathfrak{R}(T) \leq \tilde O((dkT)^{2/3}  (r^*)^{1/3}). $$
\end{theorem}

We summarize all our results and prior work for quadratic reward in Table \ref{tab:quadratic}.

\subsubsection{RL with Simulator: $Q$-function is Quadratic and Bellman Complete}
\label{sec:bellmen}
In this section we demonstrate how our results for non-concave bandits also apply to reinforcement learning. Let $\cT_h$ be the Bellman operator applied to the Q-function $Q_{h+1}$ defined as:
\begin{align*}
\cT_h(Q_{h+1})(s,a) = r_h(s,a) +  \E_{s' \sim \PP(\cdot|s,a)}[\max_{a'}Q_{h+1}(s',a')].
\end{align*}
\begin{definition}[Bellman complete]
	\label{def:bellman_complete}
	Given MDP $\mdps = (\states,\actions,\PP,r,H)$, function class $\F_h: \states \times \actions \mapsto \R ,h \in [H]$ is called Bellman complete if for all $h \in [H]$ and $Q_{h+1} \in \F_{h+1}$, $\T_h(Q_{h+1}) \in \F_h$.
\end{definition}

\begin{assumption}[Bellman complete for low-rank quadratic reward]
	\label{assump:bellman_complete}
	We assume the function class $\F_{h} = \{ f_{\mM}: f_{\mM}(s,a) = \phi(s,a)^\top \mM\phi(s,a) , \rank(\mM) \le k, \text{ and } 0<\lambda_1(\mM)/\lambda_{\min}(\mM)\leq \tilde \kappa.   \}$ is a class of quadratic function and the MDP is Bellman complete. Here $f_{\mM}(\phi(s,a)) = \sum_{j=1}^k \lambda_j (\vv_j^\top \phi(s,a))^2$ when $\mM=\sum_{j=1}^k \lambda_j \vv_j\vv_j^\top$. Write $Q^*_h = f_{\mM_h^*}\in \F_h$. 	
\end{assumption}

{\em Observation: } When querying $s_{h-1},a_{h-1}$, we observe $s_h'\sim \PP(\cdot | s_{h-1},a_{h-1})$ and reward $r_{h-1}(s_{h-1},a_{h-1})$.

{\em Oracle to recover parameter $\hat \mM$:} Given $n\geq \tilde \Theta( d^2k^2\tilde\kappa^2(\mM)/\varepsilon^2)$, if one can play $\geq n$ samples $\va_i$ and observe $y_i \sim \va_i^\top \mM \va_i+\eta_i, i\in [n]$ with $1$-sub-gaussian and mean-zero noise $\eta$, we can recover $\hat \mM=\hat \mM(\{(\va_i,y_i)\})$ such that $\|\hat \mM-\mM\|_2 \leq \varepsilon$. This oracle is implemented via our analysis from the bandit setting.

With the oracle, at time step $H$, we can estimate $\hat \mM_H $ that is $\epsilon/H$ close to $\mM^*_H$ in spectral norm through noisy observations from the reward function with $\tilde O(\tilde\kappa^2 d^2H^2/\epsilon^2)$ samples.
Next, for each time step $h=H-1,H-1,\cdots,1$, sample $s'_i\sim \PP(\cdot |s,a)$, we define $\eta_i = \max_{a'}f_{\hat \mM_{h+1}}(s'_i, a') - \E_{s'\sim \PP(\cdot|s,a)} \max_{a'}f_{\hat \mM_{h+1}}(s',a')$. $\eta_i$ is mean-zero and
$O(1)$-sub-gaussian since it is bounded. Denote $\mM_h$ as the matrix that satisfies $f_{\mM_h}:=\cT f_{\hat \mM_{h+1}},$ which is well-defined due to Bellman completeness. We estimate $\hat \mM_h$ from the noisy observations $y_i = r_h(s,a) + \max_{a'}f_{\hat \mM_{h+1}}(s'_i, a') = \cT f_{\hat \mM_{h+1}} + \eta_i =: f_{\mM_{h}} + \eta_i$. Therefore with the oracle, we can estimate $\hat \mM_h$ such that $\|\hat \mM_h - \mM_{h}\|_2\leq \epsilon/H$ with $\Theta(\tilde\kappa^2 d^2k^2H^2/\epsilon^2)$ bandits. More details are deferred to Algorithm \ref{alg:simulator_tensor_policy_complete} and the Appendix. We state the theorem on sample complexity of finding $\epsilon$-optimal policy here:

\begin{theorem}
	\label{thm:quadratic_rl}
	Suppose $\F$ is Bellman complete associated with parameter $\tilde\kappa$.	With probability $1-\delta$, Algorithm~\ref{alg:simulator_tensor_policy_complete} learns an $\epsilon$-optimal policy $\pi$ with $\tilde\Theta(d^2k^2\tilde\kappa^2H^3/\epsilon^2 )$ samples.
\end{theorem}

Existing approaches require $O(d^3 H^2/\epsilon^2)$ trajectories, or equivalently $O(d^3 H^3/\epsilon^2) $ samples, though they operate in the online RL setting~\cite{zanette2020learning,du2021bilinear,jin2021bellman}, which is worse by a factor of $d$.

It is an open problem on how to attain $d^2$ sample complexity in the online RL setting. The quadratic Bellman complete setting can also be easily extended to any of the polynomial settings of Section \ref{sec:tensor}.

\subsection{Stochastic High-order Homogeneous Polynomial Reward}
\label{sec:tensor}
Next we move on to homogeneous high-order polynomials.
\subsubsection{The symmetric setting}
\label{sec:symmetric_tensor}
Let reward function be a $p$-th order stochastic polynomial function $f:\cA\rightarrow \R$, where the action set $\cA:=\{B_1^d=\{\va\in\R^d,\|\va\|\leq 1. \}$. $f(\va) = \mT(\va^{\otimes p}), r_t = f(\va) +\eta_t$, where $\mT=\sum_{j=1}^k \lambda_j \vv_j^{\otimes p}  $ is an orthogonally decomposable rank-$k$ tensor. $\{\vv_1,\cdots \vv_k\}$ form an orthonormal basis. Optimal reward $r^*$ satisfies $1\geq r^* = \lambda_1\geq  |\lambda_2|\cdots \geq  |\lambda_k|$. Noise $\eta_t\sim \cN(0,1)$.

In this setting, the problem is fundamentally more challenging than quadratic reward functions. On the one hand, it has a higher noise-to-signal ratio with larger $p$. One can tell from the rank-1 setting where $\mT=\lambda_1\vv_1^{\otimes p}$. For a randomly generated action $\va$ on the unit ball, $\E[\|\va^\top \vv_1\|^2]=1/d$. Therefore on average the signal strength is only $(\va^\top \vv_1)^p \sim d^{-p/2}$, much smaller than the noise level $1$. Intuitively this demonstrates why higher complexity is needed for high-order polynomials. On the other hand, it is also technically more challenging. Unlike the matrix case, the expected zeroth-order update is no longer equal to any tensor product. Therefore existing tensor decomposition arguments do not apply. Fortunately, we prove that zeroth-order optimization still pushes the iterated actions toward the optimal action with linear convergence, given a good initialization. We show the bandit optimization procedure in Algorithm \ref{alg:tensor} and
present the result in Theorem \ref{thm:tensor_staged_progress}:


\begin{algorithm}
	\caption{Phased elimination with zeroth order exploration.}
	\label{alg:tensor}
	\begin{algorithmic}[1]
		\State {\bf \underline{Input:} }  Function $f:\cA\rightarrow \R$ of polynomial degree $p$ generating noisy reward, failure probability $\delta$, error $\varepsilon$.
		\State {\bf \underline{Initialization:} }  $L_0 = C_L k\log(1/\delta) $
		; Total number of stages $S = C_S\lceil \log(1/\varepsilon)\rceil+1 $,   $\cA_0 = \{\va_0^{(1)}, \va_0^{(2)}, \cdots \va_0^{(L_0)} \}$ where each $\va_0^{(l)}$ is uniformly sampled on the unit sphere $\mathbb{S}^{d-1}$.  $\tilde\varepsilon_0 = 1$.
		\For {$s$ from $1$ to $S$}
		\State $ \tilde\varepsilon_s \leftarrow \tilde \varepsilon_{s-1}/2$, $n_s \leftarrow C_n d^p\log(d/\delta)/\lambda_1^2\tilde\varepsilon_s^2) $,$n_s\leftarrow n_s\cdot \log^3(n_s/\delta)$, $m_s\leftarrow C_m d\log(n_s/\delta)$, $\cA_s=\varnothing$.
		\For {$ l$ from  $1$ to $L_{s-1}$ }
		\State {\bf \underline{Zeroth-order optimization:} }
		\State Locate current action $\tilde \va = \va_{s-1}^{(l)}$.
		\For {$\lceil (1/(1-\alpha)) \log(2d) \rceil$ times }
		\State Sample $\vz_i\sim \cN(0,1/m_s I_d), i=1,2,\cdots n_s$.
		\State Take actions $\va_i = (1-\frac{1}{2p})\tilde \va + \frac{1}{2p}\vz_i $ and observe $r_i = \mT(\va_i)+\eta_i,i\in [n_s] $; Take actions $\frac{1}{2p}\vz_i$ and observe $r_i' = \mT(\frac{1}{2p}\vz_i)+\eta_i', i\in [n_s]$.
		\State Conduct estimation $\vy \leftarrow 1/n_s \sum_{i=1}^{n_s} (r_i-r_i')\vz_i $.
		\State Update the current action $\tilde \va\leftarrow \vy/\|\vy\|$.
		\EndFor
		\State Estimate the expected reward for $\tilde \va$ through $n_s$ samples: $r_n(\tilde\va) = 1/n_s \sum_{i=1}^{n_s} (\mT(\tilde \va)+\eta_i)$.
		\State {\bf \underline{Candidate Elimination:} }
		\If {$r_n\geq \lambda_1(1-p\tilde \varepsilon_s^2)$}
		\State Keep the action $ \cA_s\leftarrow \cA_s\cup \{\tilde \va \}$
		\EndIf
		\EndFor
		\State Label the actions: $L_s = |\cA_s|, \cA_s =: \{\va_s^{(1)},\cdots  \va_s^{(L_s)} \}$.
		\EndFor
			\State Run UCB (Algorithm \ref{alg:UCB}) with the candidate set $\cA_S$.
	\end{algorithmic}
\end{algorithm}

\begin{theorem}[Staged progress]
	\label{thm:tensor_staged_progress}
	For each stage $s$, with high probability $\cA_s$ is not empty; and at least one action $\va\in \cA_s$ satisfies:
	$\tan \theta(\va, \va^*)\leq  \tilde \varepsilon_s = 2^{-s}$. 
\end{theorem}
We defer the proofs together with formal statements to Appendix \ref{appendix:symmetric_tensor}; we will also present the proof sketch in the next subsection \ref{sec:proof_sketch_tensor}.  
\begin{remark}[Choice of step-size]
	Here we choose step-size $\zeta=1/2p$. Note that $\E[\vy] = \zeta(1-\zeta)^{p}\sigma^2 \nabla_{\va}f(\va)$. The scaling $(1-\zeta)^p\geq 1/\sqrt{e}$ ensures the signal to noise ratio not too small. The choice of weighted action is a delicate balance between making progress in optimization and controlling the noise to signal ratio.
\end{remark}
\begin{corollary}[Regret bound for tensors]
	\label{coro:tensor_regret_bound_symmetric}
	Algorithm \ref{alg:tensor} yields an regret of:
	$ \mathfrak{R}(T) \leq \tilde O\left(\sqrt{ k d^pT } \right)$,
	with high probability.
\end{corollary}

\begin{corollary}[Regret bound with burn-in period]
	\label{coro:burn-in_tensor}
	In Algorithm \ref{alg:tensor}, we first set $\varepsilon = 1/p, n_s= \tilde\Theta (d^p/\lambda_1^2) $. We can first estimate the action $\va$ such that $ \vv_1^\top \va \geq 1-1/p$ with $\tilde O(k d^p / \lambda_1^2 )$ samples. Next we change $\varepsilon = \tilde\Theta(k^{1/4}d^{1/2}\lambda_1^{-1/2} T^{-1/4} ) $, and $n_s = \tilde\Theta(d^2 \varepsilon^{-2}\lambda_1^{-2} )$ in Algorithm \ref{alg:tensor}. This procedure suffices to find $ \lambda_1 \varepsilon^2$-optimal reward with $\tilde O( kd^2 /\lambda_1^2 \varepsilon^2 )$ samples in the candidate set with size at most $\tilde O(k)$. Finally with the UCB algorithm altogether we have a regret bound of:
	\begin{align*}
	\tilde O( \frac{kd^p}{\lambda_1 } +  \sqrt{kd^2 T} ).
	\end{align*} 
\end{corollary}
In Section \ref{sec:lb} we will also demonstrate the necessity of the sample complexity in this burn-in period. 

\subsubsection{Proof Sketch of Theorem \ref{thm:tensor_staged_progress}}
\label{sec:proof_sketch_tensor}
\begin{definition}[Zeroth order gradient function]
	For some scalar $m$, we define an empirical operator $G_n:\cA\rightarrow \cA$ that is similar to the zeroth-order gradient of $f$  through $n$ samples:
	\begin{align*}
	G_n (\va):= 
	&  \frac{m}{n} \sum_{i=1}^n \left(T\left(\left((1-\frac{1}{2p})\va+\frac{1}{2p}\vz_i\right)^{\otimes p}\right)  -\mT(\frac{1}{2p} \vz_i)\right) \vz_i + (\eta_i-\eta_i') \vz_i.
	\end{align*}
	where $\vz_i\sim \cN(0,\frac{1}{m}\mI)$ and $\eta_i,\eta_i'$ are independent zero-mean 1-sub-Gaussian noise. 	Therefore we have:
	\begin{align*}
	\E[	G_n (\va)] = & m\E[ \sum_{l=0}^{p-1} \binom{p}{l} \mT(  (1-\frac{1}{2p})^{p-l} \va^{\otimes (p-l)} \otimes (\frac{1}{2p})^l \vz^{\otimes l} ) \vz  ]  \tag{Due to symmetry of Gaussian only for odd $l=:2s+1$ expectation is nonzero} \\
	= &   (1-\frac{1}{2p})^{p-2s-1} (\frac{1}{2p})^{2s+1} [ \sum_{s=0}^{\lfloor p/2-1 \rfloor } m^{-s}  \binom{p}{2s+1} \mT(  \va^{\otimes (p-2s-1)} \otimes \mI^{\otimes s+1} )  ]
	\end{align*}
	Note that for even $p$ the last term (when $s=p/2-1$) is $ \mT(\va\otimes \mI^{\otimes p/2}) = \sum_{j=1}^k \lambda_j (\va^\top \vv_j)\vv_j $. While all other terms will push the iterate towards the optimal action at a superlinear speed, the last term perform a matrix multiplication and the convergence speed will depend on the eigengap. Therefore for $p\geq 4$  we will remove the extra bias in the last term that is orthogonal to $\vv_1$ and will treat it as noise. (Notice for quadratic function $s=0=p/2-1$ is the only term in $\E[G_n(\va)]$. This is the distinction between $p=2$ and larger $p$, and why its convergence depends on eigengap.)
	
	We further define $G(\va)$ as the population version of $G_n(\va)$ by removing this undesirable bias term that will be treated as noise:
	\begin{align*}
	G(\va) = & \left\{ \begin{array}{l l}
	\E[G_n] - \frac{(\frac{1}{2p})^{p-1}(1-\frac{1}{2p}) p }{ m^{p/2-1} } \sum_{j=2}^k \lambda_j (\vv_j^\top \va)\vv_j , & \text{ when }p\text{ is even}\\
	\E[G_n], 
	& \text{ when }p\text{ is odd}.
	\end{array}	\right.\\
	= &	 \sum_{s=0}^{\floor{ (p-3)/2}} \frac{(\frac{1}{2p})^{2s+1}}{m^s } \binom{p}{2s+1} \mT(\mI^{\otimes s+1}\otimes ((1-\frac{1}{2p}) \va)^{\otimes p-2s-1})\\
	= &  \frac{1}{2}(1-\frac{1}{2p})^{p-1} \mT(\mI,\va^{\otimes p-1}) + O(1/m).
	\end{align*}
\end{definition}

We define $G(\va)$ to push the action $\va$ towards the $\vv_1$ direction with at least linear convergence rate. More precisely, their angle $\tan \theta(G(\va),\vv_1)$ will converge linearly to $0$ for proper initialization with the dynamics $\va\rightarrow G(\va)$. An easy way to see that is when $p=2$ or 3, $G$ is conducting ($3$-order tensor) power iteration. For higher-order problems, this operation $G$ is equivalent to the summation of $p, p-2, p-4, \cdots$-th order tensor product and hence the linear convergence.

The estimation error $G_n(\va)-G(\va)$ will be treated as noise (which is not mean zero when $p$ is even but will be small enough: $O((2p)^{-p}m^{-(p-1)/2})$). Therefore the iterative algorithm with $\va\rightarrow G_n(\va)$ will converge to a small neighborhood of $\vv_1$ depending on the estimation error. This estimation error is controlled by the choice of sample size $n$ in each iteration. We now provide the proof sketch:

\begin{lemma}[Initialization for $p\geq 3$; Corollary C.1 from \cite{wang2016online} ]
	\label{lemma:tensor_initialization}
	For any $\eta\in (0,1/2)$, with $L=\Theta(k\log(1/\eta))$ samples $\cA  = \{\va^{(1)},\va^{(2)},\cdots \va^{(L)}\}$ where each $\va^{(l)}$ is sampled uniformly on the sphere $\mathbb{S}^{d-1}$. At least one sample $\va\in \cA$ satisfies
	\begin{align}
	\label{eqn:good_initial_tensor}
	\max_{j\neq 1} |\vv_j^\top \va| \leq  0.5|\vv_1^\top \va|,\text{ and }|\vv_1^\top \va|\geq 1/\sqrt{d}.
	\end{align}
	with probability at least $1-\eta$.
\end{lemma}

\begin{lemma}[Iterative progress]
	\label{lemma:iterative_progress}
	Let $\alpha = 1/2$ for $p\geq 3$ in Algorithm \ref{alg:tensor}. Consider noisy operation $\va^+\rightarrow G(\va)+\vg$. If the error term $\vg$ satisfies:
	\begin{align*}
	\|\vg\| \leq & \min\{ \frac{0.025}{p} \lambda_1 (\vv_1^\top \va)^{p-2}, 0.1 \lambda_1 \tilde \varepsilon   \}\\
	&  + 0.03 \lambda_1 |\sin\theta(\vv_1,\va) |(\vv_1^\top \va)^{p-2},\\
	|\vv_1^\top \vg| \leq & 0.05\lambda_1 (\vv_1^\top \va)^{p-1}.
	\end{align*}
	
	Suppose $\va$ satisfies $0.5 |\vv_1^\top \va| \geq \max_{j\geq 2} |\vv_j^\top \va|$, 
	we have:
	\begin{align*}
	\tan\theta(\va^+,\vv_1) \leq 0.8\tan\theta(\va,\vv_1) + \tilde\varepsilon.
	\end{align*}
\end{lemma}

We can also bound $\vg$ by standard concentration plus an additional small bias term.
\begin{lemma}[Estimation error bound for $G$]
	\label{lemma:bound_g_tensor}
	For fixed value $\delta\in (0,1)$ and large enough universal constant $c_1,c_2,c_m,c_n$, when $m=c_m d\log(n/\delta), n \geq c_n d\log(d/\delta) $, we have
	\begin{align*}
	\|\vg\|\equiv \|G_n(\va)-G(\va)\|\leq & c_1 \sqrt{\frac{d^2\log^3(n/\delta)\log(d/\delta) }{n}}+e\lambda_2|\sin\theta(\va,\vv_1)| ,\\
	|\vv_1^\top \vg|\equiv |\vv_1^\top G_n(\va) - \vv_1^\top G(\va) | \leq & c_2 \sqrt{\frac{d\log^3(n/\delta)\log(d/\delta) }{n}}.
	\end{align*}
	with probability $1-\delta$. $e=0$ for odd $p$ and $e= (2p)^{-(p-1)}m^{-(p/2-1)}$ for even $p$.
\end{lemma}

Together we are able to prove Theorem \ref{thm:tensor_staged_progress}:
\begin{proof}[Proof of Theorem \ref{thm:tensor_staged_progress}]
	Initially with high probability there exists an $\va_0\in \cA_0$ such that Eqn. \eqref{eqn:good_initial_tensor} holds, i.e.,  $\vv_1^\top \va_0 \geq 1/\sqrt{d}$ and $\vv_1^\top \va_0 \geq 2 |\vv_j^\top \va_0|, \forall j\geq 2$. 
	
	Next, from Lemma \ref{lemma:bound_g_tensor}, the extra bias term is bounded by $e \lambda_2 |\sin\theta(\va,\vv_1)|\\ \leq 0.03\lambda_1(\vv_1^\top \va)^{p-2}|\sin\theta(\va,\vv_1)|$ since $e=(2p)^{-p+1}m_s^{-p/2+1}$ and with our choice of variance $ m_s\geq d\geq (\vv_1^\top \va)^{-2}$, plus $p\geq 3$.
	Next with our setting of $n_s=\tilde\Theta(d^p/(\lambda_1^2\tilde \varepsilon_t^2))$, 
	the error term $\|\E[G(\va)]-G_n(\va)\|$ is upper bounded by
	$\tilde O(\sqrt{\frac{d^2}{n} }) \leq 0.025\lambda_1 d^{-(p-2)/2}\tilde\varepsilon_s/p + 0.1\lambda_1 \tilde\varepsilon_s$. Meanwhile $|\vv_1^\top \vg| \leq \tilde O(\sqrt{\frac{d}{n_s}}) \leq 0.05\lambda_1 (\vv_1^\top \va)^{p-1}$. 
	
	This meets the requirements for Theorem \ref{lemma:iterative_progress} and therefore
	$\tan\theta(G_n(\va_0),\vv_1)\leq 0.8\tan\theta(\va_0,\vv_1)+0.1\lambda_1\tilde\varepsilon_s$. Therefore after $l$ steps will have 
	\begin{align*}
	\tan\theta(G_n^l(\va_0),\vv_1)\leq & 0.8^l \tan\theta(\va_0,\vv_1) + \sum_{i=1}^l 0.8^i \cdot 0.1\tilde\varepsilon_s\\
	 \leq& 0.8^l \tan\theta(\va_0,\vv_1) + 0.5\tilde\varepsilon_s.
	\end{align*}
		
	Notice initially $\tan\theta(\va_0,\vv_1)\leq 1/(\vv_1^\top \va_0)\leq \sqrt{d}$. Therefore after at most\\ $l=O(\log_2(\tan\theta(\va_0,\vv_1))) \leq O(\log_2(d))$ steps, we will have $ \tan(G_n^l(\va_0),\vv_1)\leq \tilde\varepsilon_0/2 =\tilde \varepsilon_{1} $.
	With the same argument, the progress also holds for $s>0$ with even smaller $l$.
\end{proof}

\subsubsection{The asymmetric setting}
\label{sec:asymmetric_tensor}
Now we consider the asymmetric tensor problem with reward $f: \cA\rightarrow \R$. The input space $\cA$ consists of $p$ vectors in a unit ball: $\vec \va=(\va(1),\va(2),\cdots \va(p))\in \cA, \|\va(s)\|\leq 1,\forall s\in [p]$. $f(\vec\va) = \mT(\otimes_{s=1}^p \va(s)) + \eta.$ Tensor $\mT = \sum_{j=1}^k \lambda_j \vv_j(1)\otimes \vv_j(2)\cdots\otimes \vv_j(p)$. For each $s\in [p]$, $\{\vv_1(s),\vv_2(s),\cdots \vv_k(s)\}$ are orthonormal vectors. We order the eigenvalues such that $\lambda_1\geq |\lambda_2|\cdots \geq |\lambda_k|$. Therefore the optimal reward is $\lambda_1$ and can be achieved by $\va^*(s) = \vv_1(s), s\in [p]$. In this section we only consider $p\geq 3$ and leave the quadratic and low-rank matrix setting to the next section.
\begin{theorem}
	\label{thm:asymmetric_tensor_bound}
	For $p\geq 3$, by conducting alternating power iteration, one can get a $\varepsilon$-optimal reward with a total $\tilde O\left( (2k)^p\log^p(p/\delta) d^p\lambda_1^{-1}\varepsilon^{-1} \right)$ actions; therefore the regret bound is at most $\tilde O(\sqrt{k^p d^p T})$.
\end{theorem}
This setting is actually much easier than the symmetric setting. Notice by replacing one slice of $\vec \va$ by random Gaussian $\vz_i\sim\cN(0,2/d\log(d/\delta))$, one directly gets $\mT(\va(1),\cdots \va(s-1),\mI,\va(s+1),\cdots \va(p))$ on each slice with $1/n\sum_i f(\va(1),\cdots \va(s-1),\vz_i,\va(s+1),\cdots \va(p))\vz_i$ which is tensor product. We defer the proof to Appendix \ref{sec:proof_asymmetric_tensor}.

\subsubsection{Lower Bounds for Stochastic Polynomial Bandits}\label{sec:lb}


In this section we show lower bound for stochastic polynomial bandits. 
\begin{align}\label{eq:asym_bandit}
    f(\va) = \prod_{i=1}^p (\vtheta_i^\top \va) + \eta, \text{ where }\eta \sim N(0,1), \|\va\|_2 \leq 1\text{ and }f(\va) \leq 1.
\end{align}

\begin{theorem}\label{thm:lb_asym_tensor}
Define minimax regret as follow
\begin{align*}
    \mathfrak{R}(d,p,T)  = \inf_{\pi} \sup_{(\vtheta_1,\dots,\vtheta_p)} \E_{(\vtheta_1,\dots,\vtheta_p)} \left[T\max_{\va}\prod_{i=1}^p (\vtheta_i^\top \va) - \sum_{t=1}^T \prod_{i=1}^p (\vtheta_i^\top \va^{(t)})\right].
\end{align*}
For all algorithms $\mathcal{A}$ that adaptively interact with bandit (Eq~\eqref{eq:asym_bandit}) for $T$ rounds, we have
$\mathfrak{R}(d,p,T) \geq \Omega(\sqrt{d^pT}/p^{p})$.
\end{theorem}
From the theorem, we can see even when the problem is rank-1, any algorithm incurs at least $\Omega(\sqrt{d^pT}/p^{p})$ regret. This further implies algorithm requires sample complexity of $\Omega((d/p^2)^p/\epsilon^2)$ to attain $\epsilon$-optimal reward. This means our regret upper bound obtained in Corollary \ref{coro:tensor_regret_bound_symmetric} is optimal in terms of dependence on $d$.
 In Appendix~\ref{sec:proof_lb} we also show a $\Omega(\sqrt{d^pT})$ lower bound for asymmetric actions setting, which our upper bound up to poly-logarithmic factors. 

We note that with burn-in period, our algorithm also obtains a cumulative regret of $kd^p/\lambda_1 + \sqrt{kd^2T}$, as shown in Corollary \ref{coro:burn-in_tensor}. Here for a fixed $\lambda_1$ and very large $T$, this is better result than the previous upper bound. We note that there is no contradiction with the lower bound above, since this worst case is achieved with a specific relation between $T$ and $\lambda_1$.

The burn-in period requires $kd^p/\lambda_1^2$ samples to get a constant of $r^*$. We want to investigate whether our dependence on $\lambda_1\equiv r^*$ is optimal.  
Next we show a gap-dependent lower bound for finding an arm that is close to the optimal arm by a constant factor.
\begin{theorem}\label{thm:lb_gap_dep}
For all algorithms $\mathcal{A}$ that adaptively interact with bandit (Eq~\eqref{eq:asym_bandit}) for $T$ rounds and output a vector $\va^{(T)}\in \R^d$, it requires at least $T = \Omega(d^p/\|\vtheta\|^{2p})$ rounds to find an arm $\va^{(T)}\in \R^d$ such that 
\begin{align*}
    \prod_{i=1}^p (\vtheta_i^\top \va^{(T)}) \geq \frac{3}{4} \cdot \max_{\va}\prod_{i=1}^p (\vtheta_i^\top \va).
\end{align*}
\end{theorem}
In the rank-1 setting $r^*=\|\vtheta\|^p$ and the lower bound for achieving a constant approximation for the optimal reward is $\Omega(d^p/(r^*)^2)$. Therefore our burn-in sample complexity is also optimal in the dependence on $d$ and $r^*$. 
\subsection{Noiseless Polynomial Reward}
\label{sec:noiseless}
In this subsection, we study the regret bounds for learning bandits with \textbf{noiseless} polynomial rewards.
First we present the definition of admissible polynomial families.
\begin{definition}[Admissible Polynomial Family]\label{def:admissible}
For $\va \in \mathbb{R}^d$, define $\widetilde{\va} = [1, \va^\top]^\top$. For a algebraic variety $\gV \subseteq (\mathbb{R}^{d+1})^{\otimes p}$, define $\mathcal{R}_{\gV}:=\left \{r_\vtheta(\va)= \left\langle \vtheta, \widetilde{\va}^{\otimes p}\right \rangle: \vtheta \in \mathcal{V}  \right\}$ as the polynomial family with parameters in $\gV$. We define the dimension of the family $\mathcal{R}_{\gV}$ as the algebraic dimension of $\gV$. Next, define $\mathcal{X}: = \left\{ \widetilde{\va}^{\otimes p}: \va \in \mathbb{R}^d\right\}$. An polynomial family $\mathcal{R}_{\gV}$ is said to be admissible\footnote{Intuitively, admissibility means the dimension of $\mathcal{X}$ decreases by one when there is an additional linear constraint $\langle \vtheta, X\rangle =0$} w.r.t. $\mathcal{X}$ if for any $\vtheta \in \gV$, $\mathrm{dim}(\mathcal{X} \cap \{X \in \mathcal{X} : \langle X,  \vtheta\rangle =0\rangle\}  )< \mathrm{dim}(\mathcal{X})=d $.
\end{definition}

%
%
\subsubsection{Upper Bounds via Solving Polynomial Equations}
\label{sec:noiseless:ub}
We show that if the action set $\mathcal{A}$ is of positive measure with respect to the Lebesgue measure $\mu$, then by playing actions randomly, we can uniquely solve for the ground-truth reward function $r_\vtheta(\va)$ almost surely with samples of size that scales with the intrinsic algebraic dimension of $\gV$, provided that $\gV$ is an admissible algebraic variety.


\begin{theorem}
\label{thm:poly_noiseless}
Assume that the reward function class is an admissible polynomial family $\mathcal{R}_{\gV}$, and the maximum reward is upper bounded by $1$. If $\mu(\cA)>0$, where $\mu$ is the Lebesgue meaure, then by randomly sample actions $\va_1,\dots,\va_T$ from $\PP _{\va \sim \cN(0,I_d)} ( \cdot | \va \in \cA ) $, when $T \geq 2 \mathrm{dim}(\gV)$, we can uniquely solve for the ground-truth $\vtheta$ and thus determine the optimal action almost surely. Therefore, the cumulative regret at round $T$ can be bounded as
\[
    \mathfrak{R}(T) \leq \min\{ T, 2\mathrm{dim}(\gV)\}.
\]
\end{theorem}
We state two important examples of admissible polynomial families with $O(d)$ dimensions.
\begin{example}[low-rank polynomials]  \label{ex:nn}
The function class $\mathcal{R}_{\gV}$ of possibly inhomogeneous degree-$p$ polynomials with $k$ summands $\mathcal{R}_{\gV} = \{ r(\va) =  \sum_{i = 1}^k \lambda_i \langle \vv_i, \va \rangle^{p_i}  \mid  \lambda_i \in \mathbb{R} , \vv_i \in \mathbb{R}^d \}$ is admissible with $\mathrm{dim}(\mathcal{R}_{\gV}) \leq dk$, where $p  = \max\{p_i\}$. Neural network with monomial/polynomial activation functions are low-rank polynomials.
\end{example}

\begin{example}[\cite{chen2020learning}] \label{ex:qux}
The function class $ \mathcal{R}_{\gV} = \{ r(\va) = q(\mU \va) \mid \mU \in \mathbb{R}^{k \times d}, \mathrm{deg}\ q(\cdot) \leq p \}$ is admissible with $\mathrm{dim}(\gV) \leq   dk + (k+1)^p$.
\end{example}




\subsubsection{Lower Bounds with UCB Algorithms}
\label{sec:noiseless:lb}
%

In this subsection, we construct a hard bandit problem where the rewards are noiseless degree-$p$ polynomial, and show that any UCB algorithm needs at least $\Omega(d^p)$ actions to learn the optimal action.
On the contrary, Theorem~\ref{thm:poly_noiseless} shows that by playing actions randomly, we only need $2(dk + (p+1)^p) = O(d)$ actions.

\paragraph{Hard Case Construction}

%
Let $\ve_i$ denotes the $i$-th standard orthonormal basis of $\R^{d}$, i.e., $\ve_i$ has only one $1$ at the $i$-th entry and $0$'s for other entries. 
We define a $p$-th multi-indices set $\Lambda$ as 
$
    \Lambda = \{ (\alpha_1,\dots,\alpha_p) | 1 \leq \alpha_1 < \dots < \alpha_p \leq d \}.
$
For an $\alpha = (\alpha_1,\dots,\alpha_p) \in \Lambda$, denote $\mM_\alpha = \ve_{\alpha_1} \otimes \dots \otimes \ve_{\alpha_p}$. Then the model space $\cM$ is defined as $ \cM =   \Big\{ \mM_\alpha | \alpha \in \Lambda \Big\} $, which is a subset of rank-1 $p$-th order tensors.
The action set $\cA$ is defined  as $ \cA = \mathrm{conv}(\{ \ve_{\alpha_1} + \dots + \ve_{\alpha_p} | \alpha \in \Lambda \}) $.
Assume that the ground-truth parameter is $\mM^* = \mM_{\alpha^*} \in \cM$. 
The noiseless reward $r_t = r(\mM^*, \va_t) =  \langle \mM^*, (\va_t)^{\otimes p} \rangle = \prod_{i=1}^p \langle \ve_{\alpha^*_i}, \va_t\rangle$ is a polynomial of $\va_t$ and falls into the case of Example~\ref{ex:qux}.





\paragraph{UCB Algorithms} The UCB algorithms sequentially maintain a confidence set $\cC_t$ after playing actions $\va_1,\dots, \va_t$. Then UCB algorithms play $\va_{t+1} \in \argmax_{\va\in \cA} \mathrm{UCB}_t(\va)$, where
$ \mathrm{UCB}_t(\va) = \max_{\mM \in \cC_t} \langle \mM, (\va)^{\otimes p} \rangle$.

\begin{theorem}
\label{thm:lb_noiseless_ucb}
Assume that for each $t \geq 0$, the confidence set $\cC_t$ contains the ground-truth model, i.e., $\mM^* \in \cC_t$. Then for the noiseless degree-$p$ polynomial bandits, any UCB algorithm needs to play at least ${d\choose p}-1$ actions to distinguish models in $\cM$. Furthermore, the worst-case cumulative regret at round $T$ can be lower bounded by 
\[
    \mathfrak{R}(T) \geq \min\{ T, {d \choose p}  - 1\}.
\]

\end{theorem}

Theorem~\ref{thm:lb_noiseless_ucb} shows the failure of the optimistic mechanism, which forbids the algorithm to play an informative action that is known to be of low reward for all models in the confidence set. On the contrary, the reward function class falls into the form of $q(\mU \va)$, therefore, by playing actions randomly\footnote{Careful readers may notice that $\mathcal{A}$ is of measure zero in this setting. However, since the reward function is a homogenous polynomial of degree $p$, we can actually obtain the rewards on $\mathrm{conv}(\mathcal{A}, \mathbf{0})$, which is of positive measure.}, we only need $O(d)$ actions as Theorem~\ref{thm:poly_noiseless} suggests. 

\section{Conclusion}
In this paper, we design minimax-optimal algorithms for a broad class of bandit problems with non-concave rewards.  For the stochastic setting, our algorithms and analysis cover the low-rank linear reward setting, bandit eigenvector problem, and homogeneous polynomial reward functions.  We improve the best-known regret from prior work and attain the optimal dependence on problem dimension $d$. Our techniques naturally extend to RL in the generative model settings. Furthermore, we obtain the optimal regret, dependent on the intrinsic algebraic dimension, for general polynomial reward without noise. Our regret bound demonstrates the fundamental limits of UCB algorithms, being $\Omega(d^{p-1})$ worse than our result for cases of interest.  

We leave to future work several directions. First, the gap-free algorithms for the low-rank linear and bandit eigenvector problem do not attain $\sqrt{T}$ regret. We believe $\sqrt{d^2 T}$ regret without any dependence on gap is impossible, but do not have a lower bound. Secondly, our study of degree $p$ polynomials only covers the noiseless setting or orthogonal tensors with noise. We believe entirely new algorithms are needed for general polynomial bandits to surpass eluder dimension. As a first step, designing optimal algorithms would require understanding the stability of algebraic varieties under noise, so we leave this as a difficult future problem. Finally, we conjecture our techniques can be used to design optimal algorithms for representation learning in bandits and MDPs~\cite{yang2020provable,hu2021nearoptimal}.


\section*{Acknowledgment}
JDL acknowledges support of the ARO under MURI Award W911NF-11-1-0303,  the Sloan Research Fellowship, NSF CCF 2002272, and an ONR Young Investigator Award. QL
is supported by NSF 2030859 and the Computing Research
Association for the CIFellows Project. SK acknowledges funding from the NSF Award CCF-1703574 and the ONR award N00014-18-1-2247. The authors would like to thank Qian Yu for numerous conversations regarding the lower bound in Theorem 3.23. JDL would like to thank Yuxin Chen and Anru Zhang for several conversations on tensor power iteration, Simon S. Du and Yangyi Lu for explaining to  him to the papers of \cite{jun2019bilinear,lu2021low}, and Max Simchowitz and Chao Gao for several conversations regarding adaptive lower bounds. 

\bibliography{ref,add_ref,beyondntk,jason_bib}

\newcommand{\etalchar}[1]{$^{#1}$}
\begin{thebibliography}{GMMM21}

\bibitem[ACCD12]{arias2012fundamental}
Ery Arias-Castro, Emmanuel~J Candes, and Mark~A Davenport.
\newblock On the fundamental limits of adaptive sensing.
\newblock {\em IEEE Transactions on Information Theory}, 59(1):472--481, 2012.

\bibitem[AFH{\etalchar{+}}11]{agarwal2011stochastic}
Alekh Agarwal, Dean~P Foster, Daniel Hsu, Sham~M Kakade, and Alexander Rakhlin.
\newblock Stochastic convex optimization with bandit feedback.
\newblock {\em arXiv preprint arXiv:1107.1744}, 2011.

\bibitem[AGH{\etalchar{+}}14]{anandkumar2014tensor}
Animashree Anandkumar, Rong Ge, Daniel Hsu, Sham~M Kakade, and Matus Telgarsky.
\newblock Tensor decompositions for learning latent variable models.
\newblock {\em Journal of machine learning research}, 15:2773--2832, 2014.

\bibitem[AJKS19]{agarwal2019reinforcement}
Alekh Agarwal, Nan Jiang, Sham~M Kakade, and Wen Sun.
\newblock Reinforcement learning: Theory and algorithms.
\newblock {\em CS Dept., UW Seattle, Seattle, WA, USA, Tech. Rep}, 2019.

\bibitem[AYBM14]{abbasi2014linear}
Yasin Abbasi-Yadkori, Peter~L Bartlett, and Alan Malek.
\newblock Linear programming for large-scale {M}arkov decision problems.
\newblock {\em arXiv preprint arXiv:1402.6763}, 2014.

\bibitem[AYPS11]{abbasi2011improved}
Yasin Abbasi-Yadkori, D{\'a}vid P{\'a}l, and Csaba Szepesv{\'a}ri.
\newblock Improved algorithms for linear stochastic bandits.
\newblock In {\em NIPS}, volume~11, pages 2312--2320, 2011.

\bibitem[AYPS12]{abbasi2012online}
Yasin Abbasi-Yadkori, David Pal, and Csaba Szepesvari.
\newblock Online-to-confidence-set conversions and application to sparse
  stochastic bandits.
\newblock In {\em Artificial Intelligence and Statistics}, pages 1--9. PMLR,
  2012.

\bibitem[AZL17]{allen2017first}
Zeyuan Allen-Zhu and Yuanzhi Li.
\newblock First efficient convergence for streaming k-pca: a global, gap-free,
  and near-optimal rate.
\newblock In {\em 2017 IEEE 58th Annual Symposium on Foundations of Computer
  Science (FOCS)}, pages 487--492. IEEE, 2017.

\bibitem[AZL19]{allen2019can}
Zeyuan Allen-Zhu and Yuanzhi Li.
\newblock What can resnet learn efficiently, going beyond kernels?
\newblock {\em arXiv preprint arXiv:1905.10337}, 2019.

\bibitem[BCB12]{bubeck2012regret}
Sebastien Bubeck and Nicolo Cesa-Bianchi.
\newblock Regret analysis of stochastic and nonstochastic multi-armed bandit
  problems.
\newblock {\em Foundations and Trends in Machine Learning}, 5(1), 2012.

\bibitem[BCR13]{bochnak2013real}
Jacek Bochnak, Michel Coste, and Marie-Fran{\c{c}}oise Roy.
\newblock {\em Real algebraic geometry}, volume~36.
\newblock Springer Science \& Business Media, 2013.

\bibitem[BDWY16]{balcan2016improved}
Maria-Florina Balcan, Simon~Shaolei Du, Yining Wang, and Adams~Wei Yu.
\newblock An improved gap-dependency analysis of the noisy power method.
\newblock In {\em Conference on Learning Theory}, pages 284--309. PMLR, 2016.

\bibitem[BL20]{bai2019beyond}
Yu~Bai and Jason~D. Lee.
\newblock Beyond linearization: On quadratic and higher-order approximation of
  wide neural networks.
\newblock In {\em International Conference on Learning Representations}, 2020.

\bibitem[BLE17]{bubeck2017kernel}
S{\'e}bastien Bubeck, Yin~Tat Lee, and Ronen Eldan.
\newblock Kernel-based methods for bandit convex optimization.
\newblock In {\em Proceedings of the 49th Annual ACM SIGACT Symposium on Theory
  of Computing}, pages 72--85, 2017.

\bibitem[CBL{\etalchar{+}}20]{chen2020towards}
Minshuo Chen, Yu~Bai, Jason~D Lee, Tuo Zhao, Huan Wang, Caiming Xiong, and
  Richard Socher.
\newblock Towards understanding hierarchical learning: Benefits of neural
  representations.
\newblock {\em Neural Information Processing Systems (NeurIPS)}, 2020.

\bibitem[CLM{\etalchar{+}}16]{cai2016optimal}
T~Tony Cai, Xiaodong Li, Zongming Ma, et~al.
\newblock Optimal rates of convergence for noisy sparse phase retrieval via
  thresholded wirtinger flow.
\newblock {\em The Annals of Statistics}, 44(5):2221--2251, 2016.

\bibitem[CLS15]{candes2015phase}
Emmanuel~J Candes, Xiaodong Li, and Mahdi Soltanolkotabi.
\newblock Phase retrieval via wirtinger flow: Theory and algorithms.
\newblock {\em IEEE Transactions on Information Theory}, 61(4):1985--2007,
  2015.

\bibitem[CM20]{chen2020learning}
Sitan Chen and Raghu Meka.
\newblock Learning polynomials in few relevant dimensions.
\newblock In {\em Conference on Learning Theory}, pages 1161--1227. PMLR, 2020.

\bibitem[DHK08]{dani2008stochastic}
Varsha Dani, Thomas~P Hayes, and Sham~M Kakade.
\newblock Stochastic linear optimization under bandit feedback.
\newblock {\em The 21st Annual Conference on Learning Theory}, 2008.

\bibitem[DHK{\etalchar{+}}20]{du2020few}
Simon~S Du, Wei Hu, Sham~M Kakade, Jason~D Lee, and Qi~Lei.
\newblock Few-shot learning via learning the representation, provably.
\newblock {\em arXiv preprint arXiv:2002.09434}, 2020.

\bibitem[DKL{\etalchar{+}}21]{du2021bilinear}
Simon~S Du, Sham~M Kakade, Jason~D Lee, Shachar Lovett, Gaurav Mahajan, Wen
  Sun, and Ruosong Wang.
\newblock Bilinear classes: A structural framework for provable generalization
  in {RL}.
\newblock {\em arXiv preprint arXiv:2103.10897}, 2021.

\bibitem[DLL{\etalchar{+}}19]{du2019gradient}
Simon~S Du, Jason~D Lee, Haochuan Li, Liwei Wang, and Xiyu Zhai.
\newblock Gradient descent finds global minima of deep neural networks.
\newblock In {\em International Conference on Machine Learning}, pages
  1675--1685, 2019.

\bibitem[DML21]{damian2021label}
Alex Damian, Tengyu Ma, and Jason Lee.
\newblock Label noise sgd provably prefers flat global minimizers.
\newblock {\em arXiv preprint arXiv:2106.06530}, 2021.

\bibitem[DYM21]{dong2021provable}
Kefan Dong, Jiaqi Yang, and Tengyu Ma.
\newblock Provable model-based nonlinear bandit and reinforcement learning:
  Shelve optimism, embrace virtual curvature.
\newblock {\em arXiv preprint arXiv:2102.04168}, 2021.

\bibitem[FKM04]{flaxman2004online}
Abraham~D Flaxman, Adam~Tauman Kalai, and H~Brendan McMahan.
\newblock Online convex optimization in the bandit setting: gradient descent
  without a gradient.
\newblock {\em arXiv preprint cs/0408007}, 2004.

\bibitem[FLYZ20]{fang2020modeling}
Cong Fang, Jason~D Lee, Pengkun Yang, and Tong Zhang.
\newblock Modeling from features: a mean-field framework for over-parameterized
  deep neural networks.
\newblock {\em arXiv preprint arXiv:2007.01452}, 2020.

\bibitem[GCL{\etalchar{+}}19]{gao2019convergence}
Ruiqi Gao, Tianle Cai, Haochuan Li, Liwei Wang, Cho-Jui Hsieh, and Jason~D Lee.
\newblock Convergence of adversarial training in overparametrized networks.
\newblock {\em Neural Information Processing Systems (NeurIPS)}, 2019.

\bibitem[GHM15]{garber2015online}
Dan Garber, Elad Hazan, and Tengyu Ma.
\newblock Online learning of eigenvectors.
\newblock In {\em International Conference on Machine Learning}, pages
  560--568. PMLR, 2015.

\bibitem[GLM18]{ge2018learning}
Rong Ge, Jason~D Lee, and Tengyu Ma.
\newblock Learning one-hidden-layer neural networks with landscape design.
\newblock {\em International Conference on Learning Representations (ICLR)},
  2018.

\bibitem[GMMM19]{ghorbani2019linearized}
Behrooz Ghorbani, Song Mei, Theodor Misiakiewicz, and Andrea Montanari.
\newblock Linearized two-layers neural networks in high dimension.
\newblock {\em arXiv preprint arXiv:1904.12191}, 2019.

\bibitem[GMMM21]{ghorbani2021linearized}
Behrooz Ghorbani, Song Mei, Theodor Misiakiewicz, and Andrea Montanari.
\newblock Linearized two-layers neural networks in high dimension.
\newblock {\em The Annals of Statistics}, 49(2):1029--1054, 2021.

\bibitem[GMZ16]{gopalan2016low}
Aditya Gopalan, Odalric-Ambrym Maillard, and Mohammadi Zaki.
\newblock Low-rank bandits with latent mixtures.
\newblock {\em arXiv preprint arXiv:1609.01508}, 2016.

\bibitem[HCJ{\etalchar{+}}21]{hu2021nearoptimal}
Jiachen Hu, Xiaoyu Chen, Chi Jin, Lihong Li, and Liwei Wang.
\newblock Near-optimal representation learning for linear bandits and linear
  rl, 2021.

\bibitem[HL16]{hazan2016optimal}
Elad Hazan and Yuanzhi Li.
\newblock An optimal algorithm for bandit convex optimization.
\newblock {\em arXiv preprint arXiv:1603.04350}, 2016.

\bibitem[HLSW21]{hao2021online}
Botao Hao, Tor Lattimore, Csaba Szepesv{\'a}ri, and Mengdi Wang.
\newblock Online sparse reinforcement learning.
\newblock In {\em International Conference on Artificial Intelligence and
  Statistics}, pages 316--324. PMLR, 2021.

\bibitem[HLW20]{hao2020high}
Botao Hao, Tor Lattimore, and Mengdi Wang.
\newblock High-dimensional sparse linear bandits.
\newblock {\em arXiv preprint arXiv:2011.04020}, 2020.

\bibitem[HP14]{hardt2014noisy}
Moritz Hardt and Eric Price.
\newblock The noisy power method: A meta algorithm with applications.
\newblock {\em Advances in neural information processing systems},
  27:2861--2869, 2014.

\bibitem[HWLM20]{haochen2020shape}
Jeff~Z. HaoChen, Colin Wei, Jason~D. Lee, and Tengyu Ma.
\newblock Shape matters: Understanding the implicit bias of the noise
  covariance.
\newblock {\em arXiv preprint arXiv:2006.08680}, 2020.

\bibitem[HZWS20]{hao2020low}
Botao Hao, Jie Zhou, Zheng Wen, and Will~Wei Sun.
\newblock Low-rank tensor bandits.
\newblock {\em arXiv preprint arXiv:2007.15788}, 2020.

\bibitem[JHG18]{jacot2018neural}
Arthur Jacot, Cl{\'e}ment Hongler, and Franck Gabriel.
\newblock Neural tangent kernel: Convergence and generalization in neural
  networks.
\newblock In {\em NeurIPS}, 2018.

\bibitem[JLM21]{jin2021bellman}
Chi Jin, Qinghua Liu, and Sobhan Miryoosefi.
\newblock Bellman eluder dimension: New rich classes of rl problems, and
  sample-efficient algorithms.
\newblock {\em arXiv preprint arXiv:2102.00815}, 2021.

\bibitem[JSB16]{johnson2016structured}
Nicholas Johnson, Vidyashankar Sivakumar, and Arindam Banerjee.
\newblock Structured stochastic linear bandits.
\newblock {\em arXiv preprint arXiv:1606.05693}, 2016.

\bibitem[JWWN19]{jun2019bilinear}
Kwang-Sung Jun, Rebecca Willett, Stephen Wright, and Robert Nowak.
\newblock Bilinear bandits with low-rank structure.
\newblock In {\em International Conference on Machine Learning}, pages
  3163--3172. PMLR, 2019.

\bibitem[KKS{\etalchar{+}}17]{katariya2017stochastic}
Sumeet Katariya, Branislav Kveton, Csaba Szepesvari, Claire Vernade, and Zheng
  Wen.
\newblock Stochastic rank-1 bandits.
\newblock In {\em Artificial Intelligence and Statistics}, pages 392--401.
  PMLR, 2017.

\bibitem[Kle04]{kleinberg2004nearly}
Robert Kleinberg.
\newblock Nearly tight bounds for the continuum-armed bandit problem.
\newblock {\em Advances in Neural Information Processing Systems}, 17:697--704,
  2004.

\bibitem[KN19]{kotlowski2019bandit}
Wojciech Kot{\l}owski and Gergely Neu.
\newblock Bandit principal component analysis.
\newblock In {\em Conference On Learning Theory}, pages 1994--2024. PMLR, 2019.

\bibitem[KTB19]{kileel2019expressive}
Joe Kileel, Matthew Trager, and Joan Bruna.
\newblock On the expressive power of deep polynomial neural networks.
\newblock {\em Advances in Neural Information Processing Systems}, 32, 2019.

\bibitem[LAAH19]{lale2019stochastic}
Sahin Lale, Kamyar Azizzadenesheli, Anima Anandkumar, and Babak Hassibi.
\newblock Stochastic linear bandits with hidden low rank structure.
\newblock {\em arXiv preprint arXiv:1901.09490}, 2019.

\bibitem[Lat20]{lattimore2020improved}
Tor Lattimore.
\newblock Improved regret for zeroth-order adversarial bandit convex
  optimisation, 2020.

\bibitem[LCLS10]{li2010contextual}
Lihong Li, Wei Chu, John Langford, and Robert~E Schapire.
\newblock A contextual-bandit approach to personalized news article
  recommendation.
\newblock In {\em Proceedings of the 19th international conference on World
  wide web}, pages 661--670, 2010.

\bibitem[LH21]{lattimore2021bandit}
Tor Lattimore and Botao Hao.
\newblock Bandit phase retrieval.
\newblock {\em arXiv preprint arXiv:2106.01660}, 2021.

\bibitem[LL18]{li2018learning}
Yuanzhi Li and Yingyu Liang.
\newblock Learning overparameterized neural networks via stochastic gradient
  descent on structured data.
\newblock In {\em Advances in Neural Information Processing Systems}, pages
  8157--8166, 2018.

\bibitem[LM13]{lecue2013minimax}
Guillaume Lecu{\'e} and Shahar Mendelson.
\newblock Minimax rate of convergence and the performance of erm in phase
  recovery.
\newblock {\em arXiv preprint arXiv:1311.5024}, 2013.

\bibitem[LMT21]{lu2021low}
Yangyi Lu, Amirhossein Meisami, and Ambuj Tewari.
\newblock Low-rank generalized linear bandit problems.
\newblock In {\em International Conference on Artificial Intelligence and
  Statistics}, pages 460--468. PMLR, 2021.

\bibitem[LS20]{lattimore2020bandit}
Tor Lattimore and Csaba Szepesv{\'a}ri.
\newblock {\em Bandit algorithms}.
\newblock Cambridge University Press, 2020.

\bibitem[MGW{\etalchar{+}}20]{moroshko2020implicit}
Edward Moroshko, Suriya Gunasekar, Blake Woodworth, Jason~D Lee, Nathan Srebro,
  and Daniel Soudry.
\newblock Implicit bias in deep linear classification: Initialization scale vs
  training accuracy.
\newblock {\em Neural Information Processing Systems (NeurIPS)}, 2020.

\bibitem[Mil17]{milneAG}
James~S. Milne.
\newblock Algebraic geometry (v6.02), 2017.
\newblock Available at www.jmilne.org/math/.

\bibitem[MM15]{musco2015randomized}
Cameron Musco and Christopher Musco.
\newblock Randomized block krylov methods for stronger and faster approximate
  singular value decomposition.
\newblock {\em arXiv preprint arXiv:1504.05477}, 2015.

\bibitem[NGL{\etalchar{+}}19]{nacson2019lexicographic}
Mor~Shpigel Nacson, Suriya Gunasekar, Jason Lee, Nathan Srebro, and Daniel
  Soudry.
\newblock Lexicographic and depth-sensitive margins in homogeneous and
  non-homogeneous deep models.
\newblock In {\em International Conference on Machine Learning}, pages
  4683--4692. PMLR, 2019.

\bibitem[RT10]{rusmevichientong2010linearly}
Paat Rusmevichientong and John~N Tsitsiklis.
\newblock Linearly parameterized bandits.
\newblock {\em Mathematics of Operations Research}, 35(2):395--411, 2010.

\bibitem[RTS18]{riquelme2018deep}
Carlos Riquelme, George Tucker, and Jasper Snoek.
\newblock Deep bayesian bandits showdown: An empirical comparison of bayesian
  deep networks for thompson sampling, 2018.

\bibitem[RVR13]{russo2013eluder}
Dan Russo and Benjamin Van~Roy.
\newblock Eluder dimension and the sample complexity of optimistic exploration.
\newblock In {\em Advances in Neural Information Processing Systems}, pages
  2256--2264, 2013.

\bibitem[SBRL19]{sanjabi2019does}
Maziar Sanjabi, Sina Baharlouei, Meisam Razaviyayn, and Jason~D Lee.
\newblock When does non-orthogonal tensor decomposition have no spurious local
  minima?
\newblock {\em arXiv preprint arXiv:1911.09815}, 2019.

\bibitem[Sha13]{shafarevich2013basic}
Igor~R Shafarevich.
\newblock {\em Basic Algebraic Geometry 1: Varieties in Projective Space}.
\newblock Springer Science \& Business Media, 2013.

\bibitem[Tro12]{tropp2012user}
Joel~A Tropp.
\newblock User-friendly tail bounds for sums of random matrices.
\newblock {\em Foundations of computational mathematics}, 12(4):389--434, 2012.

\bibitem[VKM{\etalchar{+}}13]{valko2013finite}
Michal Valko, Nathaniel Korda, R{\'e}mi Munos, Ilias Flaounas, and Nelo
  Cristianini.
\newblock Finite-time analysis of kernelised contextual bandits.
\newblock {\em arXiv preprint arXiv:1309.6869}, 2013.

\bibitem[WA16]{wang2016online}
Yining Wang and Animashree Anandkumar.
\newblock Online and differentially-private tensor decomposition.
\newblock {\em arXiv preprint arXiv:1606.06237}, 2016.

\bibitem[WGL{\etalchar{+}}19]{woodworth2019kernel}
Blake Woodworth, Suriya Genesekar, Jason Lee, Daniel Soudry, and Nathan Srebro.
\newblock Kernel and deep regimes in overparametrized models.
\newblock In {\em Conference on Learning Theory (COLT)}, 2019.

\bibitem[WLLM19]{wei2019regularization}
Colin Wei, Jason~D Lee, Qiang Liu, and Tengyu Ma.
\newblock Regularization matters: Generalization and optimization of neural
  nets vs their induced kernel.
\newblock In {\em Advances in Neural Information Processing Systems}, pages
  9709--9721, 2019.

\bibitem[WSY20]{wang2020reinforcement}
Ruosong Wang, Russ~R Salakhutdinov, and Lin Yang.
\newblock Reinforcement learning with general value function approximation:
  Provably efficient approach via bounded eluder dimension.
\newblock {\em Advances in Neural Information Processing Systems},
  33:6123--6135, 2020.

\bibitem[WWL{\etalchar{+}}20]{wang2020beyond}
Xiang Wang, Chenwei Wu, Jason~D Lee, Tengyu Ma, and Rong Ge.
\newblock Beyond lazy training for over-parameterized tensor decomposition.
\newblock {\em Neural Information Processing Systems (NeurIPS)}, 2020.

\bibitem[WX19]{wang2019generalized}
Yang Wang and Zhiqiang Xu.
\newblock Generalized phase retrieval: measurement number, matrix recovery and
  beyond.
\newblock {\em Applied and Computational Harmonic Analysis}, 47(2):423--446,
  2019.

\bibitem[XWZG20]{xu2020neural}
Pan Xu, Zheng Wen, Handong Zhao, and Quanquan Gu.
\newblock Neural contextual bandits with deep representation and shallow
  exploration.
\newblock {\em arXiv preprint arXiv:2012.01780}, 2020.

\bibitem[YHLD20]{yang2020provable}
Jiaqi Yang, Wei Hu, Jason~D Lee, and Simon~S Du.
\newblock Provable benefits of representation learning in linear bandits.
\newblock {\em arXiv preprint arXiv:2010.06531}, 2020.

\bibitem[ZLKB20]{zanette2020learning}
Andrea Zanette, Alessandro Lazaric, Mykel Kochenderfer, and Emma Brunskill.
\newblock Learning near optimal policies with low inherent {Bellman} error.
\newblock In {\em International Conference on Machine Learning}, 2020.

\end{thebibliography}
\bibliographystyle{alpha}

\newpage
\appendix

\section{Additional Preliminaries}
\label{appendix:prelim}
In this section we show that adapting the eluder UCB algorithms from \cite{russo2013eluder} would yield the sample complexity in \Cref{thm:prep}. Especially we give the rates in \Cref{tbl:main} for our stochastic settings.
\begin{algorithm}
	\caption{Eluder UCB}
	\label{alg:linUCB}	
	\begin{algorithmic}[1]
		\State {\bf \underline{Input:} } Function class $\Fc$, failure probability $\delta$, parameters $\alpha, N, C$.
		\State {\bf \underline{Initialization:} } $\Fc_0\leftarrow \Fc$.
    	\For {$t$ from $1$ to $T$}
		\State {\bf \underline{Select Action:} }
		\State $\va_t \in \argmax_{\va\in \Ac}\sup_{f_\vtheta\in \Fc_{t-1}} f_\vtheta(\va)$
        \State Play action $\va_t$ and observe reward $r_t$
        \State {\bf \underline{Update Statistics:} }
		\State $\hatvtheta_t \in \argmin_\vtheta \sum_{s=1}^t (f_\vtheta(\va_s)-r_s)^2$
        \State $\beta_t\leftarrow 8\log(N/\delta)+2\alpha t(8C+\sqrt{8\ln(4t^2/\delta)})$
        \State $\Fc_t \leftarrow \{f_\vtheta:\sum_{s=1}^t (f_{\vtheta}-f_{\hatvtheta_t})^2(\va_s)\leq \beta_t\}$
		\EndFor
    \end{algorithmic}
\end{algorithm}

\paragraph{The algorithm} \cite{russo2013eluder} consider \Cref{alg:linUCB} for the stochastic generalized linear bandit problem. Assume that $\vtheta^*$ is the true parameter of the reward model. The reward is $r_t=f_{\vtheta^*}(\va_t)+\eta_t$ for $f_{\vtheta^*}\in \Fc$. Let $N$ be the $\alpha$-covering-number (under $\norminf{\cdot}$) of $\Fc$, $d_E$ be the $\alpha$-eluder-dimension of $\Fc$ (see Definition 3,4 in \cite{russo2013eluder}). Let $C=\sup\limits_{f\in\Fc, a\in\Ac}|f(a)|$. We set $\alpha=\frac{1}{T^2}$ in the algorithm.
\paragraph{The regret analysis}
Choosing $\alpha=1/T^2$, proposition 4 in \cite{russo2013eluder} state that with probability $1-\delta$, for some universal constant $C$, the total regret $\mathfrak{R}(T)\leq \frac{1}{T}+C\min\{d_E,T\}+4\sqrt{d_E\beta_T T}\leq 1+C\sqrt{d_ET}+4\sqrt{d_E\beta_T T}=O(\sqrt{d_E(1+\beta_T) T})$. In our settings with $\alpha=1/T^2$, $\beta_T=8\log(N/\delta)+2(8C+\sqrt{8\ln(4T^2/\delta)})/T=O(\log(N/\delta))$ where $\log(N)=\Omega(1)$ for our action sets, and thus \[\mathfrak{R}(T)=\tilde{O}(\sqrt{d_ET\log N }).\]

\paragraph{Applications in our settings}
We show that in our settings \Cref{thm:prep} will obtain the rates listed in \Cref{tbl:main}.
\subparagraph{The covering numbers}
\begin{lemma}
    The log-covering-number (of radius $\alpha$ with $\alpha\ll 1$, under $\norminf{\cdot}$) of the function classes are: $\log N(\Fc_{\text{SYM}})=O(dk\log\frac{k}{\alpha})$, $\log N(\Fc_{\text{ASYM}})=O(dk\log\frac{k}{\alpha})$, $\log N(\Fc_{\text{EV}})= O(dk\log\frac{k}{\alpha})$, and $\log N(\Fc_{\text{LR}})=O(dk\log\frac{k}{\alpha})$.
\end{lemma}
\begin{proof}
    Let $S^d_\xi$ denote a minimal $\xi$-covering of $\mathbb{S}^{d-1}$ (under $\normtwo{\cdot}$) for $0<\xi<\frac{1}{10}$, and $|S^d_\xi|=O(d\log 1/\xi)$ (see for example \cite{russo2013eluder}). Then we can construct the coverings in our settings from $S^d_\xi$:
    \begin{itemize}
        \item $\Fc_{\text{SYM}}$: let $\xi=\frac{\alpha}{kp}$, and for $k$ copies of $S^d_\xi$, we can construct a covering of $\Fc_{\text{SYM}}$ with size $|S^d_\xi|^k$. Specifically, let the covering be $S_{\text{SYM}}=\{g(\va)=\sum_{j=1}^k \lambda_j (\vu_j^\top \va)^p:(\vu_1,\vu_2,\cdots ,\vu_k)\in S^d_\xi\times S^d_\xi\times\cdots\times S^d_\xi\}$, then for each $f(\va)=\sum_{j=1}^k \lambda_j (\vv_j^\top \va)^p\in \Fc_{\text{SYM}}$, as we can find $\vu_j\in S^d_{\xi}$ that $\normtwo{\vu_j-\vv_j}\leq\xi$,
            \[\sup_\va [f(\va)-g(\va)]\leq \sup_\va[\sum_{j=1}^k|\lambda_j| |\vu_j^\top \va-\vv_j^\top \va||\sum_{q=0}^{p-1} (\vu_j^\top \va)^q(\vv_j^\top \va)^{p-q-1}|]\leq pk\xi=\alpha;\]

        \item $\Fc_{\text{ASYM}}$: let $\xi=\frac{\alpha}{kp}$, and for $kp$ copies of $S^d_\xi$, let the covering be $S_{\text{ASYM}}=\{g(\va)=\sum_{j=1}^k \lambda_j \prod_{q=1}^p (\vu_j(q)^\top \va(q)):(\vu_1(1),\vu_1(2),\cdots,\vu_1(p),\vu_2(1),\cdots ,\vu_k(p))\in S^d_\xi\times S^d_\xi\times\cdots\times S^d_\xi\}$ with size $|S^d_\xi|^{kp}$. Then for each $f(\va)=\sum_{j=1}^k \lambda_j \prod_{q=1}^p (\vv_j(q)^\top \va(q))\in \Fc_{\text{ASYM}}$, as we can find $\vu_j(q)\in S^d_{\xi}$ that $\normtwo{\vu_j(q)-\vv_j(q)}\leq\xi$,
            \[\begin{array}{lcl}
            \sup_\va [f(\va)-g(\va)]& \leq & \sup_\va[\sum_{j=1}^k|\lambda_j| \sum_{q=1}^p|\vu_j(q)^\top \va-\vv_j(q)^\top \va|\cdot\\
            & & |\prod_{r<q}(\vu_j(r)^\top \va)\prod_{r>q}(\vv_j(r)^\top \va)|]\\
            & \leq & pk\xi=\alpha;
            \end{array}\]
        \item $\Fc_{\text{EV}}$: the construction follows that of $\Fc_{\text{SYM}}$ by taking $p=2$;
        \item $\Fc_{\text{LR}}$: taking the construction of $\Fc_{\text{SYM}}$ with $p=2$ and $\xi=\frac{\alpha}{2k}$, for $\mN=\sum_{j=1}^k \lambda_j \vu_j\vu_j^\top$ and $\mM=\sum_{j=1}^k \lambda_j \vv_j\vv_j^\top$ with $\normtwo{\vu_j-\vv_j}\leq\xi$, we know $\normF{\mN-\mM}\leq \normF{\mN-\sum_{j=1}^k \lambda_j \vu_j\vv_j^\top}+\normF{\sum_{j=1}^k \lambda_j \vu_j\vv_j^\top-\mM}\leq \sum_{j=1}^k 2|\lambda_j|\xi\leq \alpha$. Then $\sup_\mA [f_\mM(\mA)-f_\mN(\mA)]\leq \sup_\mA \normF{\mM-\mN}\cdot\normF{\mA}\leq \alpha$.
    \end{itemize}
    Then we can bound the covering numbers in \Cref{thm:prep}. Notice that in the settings the log-covering numbers are only different by constant factors.
\end{proof}


\subparagraph{The eluder dimensions}
\begin{lemma}
  The $\epsilon$-eluder-dimension ($\epsilon<1$) $d_E$ of the function classes are: $d_E(\Fc_{\text{SYM}})=\tilde{\Theta}(d^p)$ (for $k\geq p$), $d_E(\Fc_{\text{ASYM}})=\tilde{\Theta}(d^p)$, $d_E(\Fc_{\text{EV}})=\tilde{\Theta}(d^2)$, and $d_E(\Fc_{\text{LR}})=\tilde{\Theta}(d^2)$. In the settings WLOG we assume the top eigenvalue is $r^*=\lambda_1=1$ as we are mostly interested in the cases where $r^*>\epsilon$.
\end{lemma}
\begin{proof}
   The upper bounds for the eluder dimension can be given by the linear argument. \cite{russo2013eluder} show that the $d$-dimension linear model $\{f_{\vtheta}(\va)=\vtheta^\top \va\}$ has $\epsilon$-eluder-dimension $O(d\log\frac{1}{\epsilon})$. In all of these settings, we can find feature maps $\phi$ and $\psi$ so that $\Fc=\{f_\vtheta(\va),f_\vtheta(\va)=\phi(\vtheta)^\top \psi(\va), \normtwo{\phi(\vtheta)}\leq k, \normtwo{\psi(\va)}\leq k\}$. Then the eluder dimensions will be bounded by the corresponding linear dimension as an original $\epsilon$-independent sequence $\{\va_i\}$ will induce an $\epsilon$-independent sequence $\{\psi(\va_i)\}$ in the linear model. Therefore for matrices ($\Fc_{\text{LR}}$ and $\Fc_{\text{EV}}$) the eluder dimension is $O(d^2\log\frac{k}{\epsilon})$ and for the tensors ($\Fc_{\text{SYM}}$ and $\Fc_{\text{ASYM}}$) it is $O(d^p\log\frac{k}{\epsilon})$.

   Then we consider the lower bounds. We provide the following example of $O(1)$-independent sequences to bound the eluder dimension in our settings up to a $\log$ factor.
   \begin{itemize}
    \item $\Fc_{\text{SYM}}$: the sequence is $\{\va_i=(\ve_{i_1},\ve_{i_2},\cdots,\ve_{i_p}): i=(i_1,i_2,\cdots,i_p)\in [d]^p\}$. For $f_{j}(\va)=\prod_{q=1}^p \ve_{j_q}^\top \va(q)$, $f_{j}(\va_i)$ is only 1 when $i=j$ and 0 otherwise. Then each $\va_i$ is 1-independent to the predecessors on $f_i$ and zero, and thus the eluder dimension is lower bounded by $d^p$.

    \item $\Fc_{\text{ASYM}}$: for $p\leq d$ and $k\geq p$, the sequence is $\{\va_i=\frac{1}{\sqrt{p}}(\ve_{i_1}+\ve_{i_2}+\cdots +\ve_{i_p}): i=(i_1,i_2,\cdots,i_p)\in [d]^p, i_1<i_2<\cdots <i_p\}$. There are tensors $f_j$ and $g_j$ of CP-rank $k$ that $(f_j-g_j)(\va)=\prod_{q=1}^p (\ve_{j_q}^\top \va)$ where $j_1<j_2<\cdots <j_p$, $(f_{j}-g_j)(\va_i)$ is only 1 when $i=j$ and 0 otherwise. Then each $\va_i$ is 1-independent to the predecessors on $f_i$ and $g_i$, and thus the eluder dimension is lower bounded by $\binom{d}{p}$.

    \item $\Fc_{\text{EV}}$: the sequence is $\{\va_i=\frac{1}{\sqrt{2}}(\ve_{i_1}+\ve_{i_2}): i=(i_1,i_2)\in [d]^2, i_1\leq i_2\}$. For $f_{j}(\va)=\frac{1}{2}\va^\top (\ve_{j_1}+\ve_{j_2})(\ve_{j_1}+\ve_{j_2})^\top \va$ and $g_{j}(\va)=\frac{1}{2}\va^\top (\ve_{j_1}-\ve_{j_2})(\ve_{j_1}-\ve_{j_2})^\top \va$ with $j_1\leq j_2$, $(f_j-g_j)(\va_i)$ is only 1 when $i=j$ and 0 otherwise. Then each $\va_i$ is 1-independent to the predecessors on $f_i$ and $g_i$, and thus the eluder dimension is lower bounded by $\binom{d}{2}$.

    \item $\Fc_{\text{LR}}$: the sequence is $\{\mA_i=\frac{1}{2}\ve_{i_1}\ve_{i_2}^T+\ve_{i_1}\ve_{i_2}^T: i=(i_1,i_2)\in [d]^2, i_1\leq i_2\}$. For $f_{j}(\mA)=\langle\frac{1}{2}(\ve_{j_1}\ve_{j_2}^T+\ve_{j_2}\ve_{j_1}^T),\mA\rangle$ with $j_1\leq j_2$, $f_{j}(\mA_i)$ is only 1 when $i=j$  and 0 otherwise. Then each $\mA_i$ is 1-independent to the predecessors on $f_i$ and zero, and thus the eluder dimension is lower bounded by $\binom{d}{2}$.
   \end{itemize}
\end{proof}
Then we are all set for the results in the first line of \ref{tbl:main}. Notice that when we choose $\alpha=O(1/T^2)$ and $\epsilon=O(1/T^2)$ in our analysis of \Cref{alg:linUCB}, the regret upper bound would only expand by $\log(T)$ factors.

\section{Omitted Proofs for Quadratic Reward}

In this section we include all the omitted proof of the theorems presented in the main paper.

\subsection{Omitted Proofs of Main Results for Stochastic Bandit Eigenvector Problem}
\label{appendix:bandit_eigenvector}
\begin{proof}[Proof of Theorem \ref{thm:bilinear_npm}]
Notice in Algorithm \ref{alg:npm}, for each iterate $\va$, its next iterate $\vy $ satisfies
\begin{align*}
\vy = & \frac{1}{n_s} \sum_{i=1}^{n_s} (\va/2+\vz_i/2)^\top \mM(\va/2+\vz_i/2)\vz_i+ \eta_i\vz_i \\
= & \frac{m_s}{n_s} \sum_{i=1}^{n_s} (\frac{1}{4}\va^\top \mM\va + \frac{1}{2}\va^\top \mM \vz_i + \eta_i) \vz_i.
\end{align*}
Therefore $\E[\vy] = \frac{1}{2} \mM\va$. We can write $2\vy = \mM \va + \vg$ where $\vg:=  \frac{m_s}{n_s} \sum_{i=1}^{n_s} (\frac{1}{2}\va^\top \mM\va + 2\eta_i) \vz_i $. With Claim \ref{claim:gaussian_moment_concentration} and Claim \ref{claim:noise_concentration} we get that $\|\vg\|\leq C\sqrt{ \frac{m_s \log^2(n/\delta)\log(d/\delta)d }{n_s}}$. Therefore with our choice of $n_s\geq \tilde\Theta(\frac{d^2 }{ \varepsilon_s^2 (\lambda_1-|\lambda_2|)^2 })$ we guarantee $\|\vg\| \leq \varepsilon_s(\lambda_1-|\lambda_2|)$. Therefore it satisfies the requirements for noisy power method, and by applying Corollary \ref{coro:hardt}, we have with $L=O(\kappa \log(d/\varepsilon))$ iterations we will be able to find $\|\hat\va - \va^*\|\leq \varepsilon.$  By setting $\delta<0.1/L$ in the algorithm we can guarantee the whole process succeed with high probability. Altogether it is sufficient to take $Ln_s = \tilde O(\kappa d^2/(\varepsilon\Delta)^2) $ actions to get an $\varepsilon$-optimal arm.

Finally to get the cumulative regret bound, we apply Claim \ref{claim:pac2regret} with  $A = \frac{d^2\kappa}{\Delta^2}$ and $a=2$. Therefore we set $\varepsilon = A^{1/4} T^{-1/4} = \frac{d^{1/2}\kappa^{1/4}}{\Delta^{1/2}T^{1/4}}$ and get:
\begin{align*}
\Reg(T) \lesssim & T^{1/2} A^{1/2} r^* =  \sqrt{\frac{d^2\kappa}{\Delta^2} T    }r^*
=  \sqrt{d^2\kappa^3 T    }.
\end{align*}

\end{proof}

\begin{corollary}[Formal statement for Corollary \ref{lemma:gap_free_matrix}]
	In Algorithm \ref{alg:npm}, by setting $\alpha = 1-\varepsilon^2/2$, one can get $ \varepsilon$-optimal reward with a total of $\tilde O(d^2\lambda_1^2/\varepsilon^4)$ total samples to get $\va$ such that $r^* - f(\va)\leq \varepsilon  $. Therefore one can get an accumulative regret of $ \tilde O( \lambda_1^{3/5} d^{2/5}  T^{4/5} ).  $
\end{corollary}
\begin{proof}[Proof  of Lemma \ref{lemma:gap_free_matrix}]
	In order to find an arm with $\lambda_1 \varepsilon^2$-optimal reward,  one will want to recover an arm that is $\varepsilon/2$-close (meaning to find an $\va$ such that $\tan\theta(\mV_l, \va)\leq \varepsilon/2$) to the top eigenspace span$(\vv_1,\cdots \vv_l)$, where $l$ satisfies $\lambda_l\geq \lambda_1 - \tilde\varepsilon$ and $\lambda_{l+1}\leq \lambda_1-\tilde \varepsilon$. Here we set $\tilde\varepsilon:=\lambda_1\varepsilon^2/2$. We first show 1) this is sufficient to get an $\lambda_1 \varepsilon$-optimal reward, and next show 2) how to set parameter to achieve this.
	
	To get 1), we write $\mV_l = [\vv_1,\cdots \vv_l]\in \R^{d\times l}$ and $\mV_l^\perp = [\vv_{l+1},\cdots \vv_k]$. When $\tan\theta(\mV_l, \va_T) = \|\mV^\perp \va \|/\|\mV \va \| \leq \varepsilon/2$, from the proof of Claim \ref{claim:angle_to_regret}, we get $r^*-f(\va)  \leq \min\{\lambda_1,\lambda_1 2(\varepsilon/2)^2 + \tilde\varepsilon\}=\lambda_1 \varepsilon^2.$
	
	Now to get 2), we note that in each iteration we try to conduct the power iteration to find an action  $\tan\theta(\mV_l, \hat\va) \leq \varepsilon/2$ and with eigengap $\geq \tilde\varepsilon:=\lambda_1\varepsilon^2/2$. 	
	Therefore it is sufficient to let $\|\vg\| \leq 0.1 \tilde\varepsilon \varepsilon $ and $|\vv_1^\top \vg| \leq 0.1\tilde\epsilon \frac{1}{\sqrt{d}} $, and thus $n_s \geq \tilde\Theta(\frac{d^2}{\varepsilon^2 \tilde\varepsilon^2} ) \leq \tilde\Theta(d^2/\lambda_1^2\varepsilon^6) $. Together we need $\lambda_1/\tilde\epsilon \log(2d/\varepsilon)n_s   = \tilde\Theta(d^2  /\lambda_1^2 \varepsilon^8 ) $ samples to get an $\lambda_1 \varepsilon^2$-optimal reward. Namely we get $\tilde\varepsilon$-optimal reward with $\tilde O(d^2\lambda_1^2/\tilde\varepsilon^4)$ samples.
	
	Finally by applying Claim \ref{claim:pac2regret} we get:
	$$   \mathfrak{R}(T)  \lesssim (d^2\lambda_1^2)^{\frac{1}{5}} T^{\frac{4}{5}}\lambda_1^{\frac{1}{5}} \leq \tilde O( \lambda_1^{3/5} d^{2/5}  T^{4/5} ) . $$
	
	
\end{proof}

\begin{algorithm}
	\caption{Gap-free Subspace Iteration for Bilinear Bandit}
	\label{alg:gap-free_subspace_iteration}
	\begin{algorithmic}[1]
		\State {\bf \underline{Input:} }  Quadratic reward $f:\cX\rightarrow \R$ generating noisy reward, failure probability $\delta$, error $\varepsilon$.
		\State {\bf \underline{Initialization:} } Set $k'=2k$. Initial candidate matrix $\mX_0 \in\R^{d\times k'}$, $\mX_0(j)\in \R^{d}, j=1,2,\cdots k'$ is the $j$-th column of $X_0$ and are i.i.d sampled on the unit sphere $\mathbb{S}^{d-1}$ uniformly. Sample variance $m$, 
		\# sample per iteration $n$, 
		total iteration $L$.
		\For {Iteration $l$ from $1$ to $L$}
		\For {$s$ from $1$ to $k'$}
		\State {\bf \underline{Noisy subspace iteration:} }
		\State Sample $\vz_i\sim \cN(0,1/m I_d), i=1,2,\cdots n_s$.
		\State Calculate tentative rank-1 arms $\tilde \va_i = \frac{1}{2}(\mX_{l-1}(s)+\vz_i) $.
		\State Conduct estimation $\mY_l(s) \leftarrow 4m/n \sum_{i=1}^{n} (f(\tilde\va_i) +\eta_i)\vz_i $. ($\mY_l\in \R^{d\times k'}$)
		\EndFor
		\State Let $\mY_l = \mX_l \mR_l$ be a QR-factorization of $\mY_l$
		\State Update target arm $\va_l\leftarrow \argmax_{\|\va\|=1} \va^\top \mY_l\mX_{l-1}^\top \va$.
		\EndFor
		\State {\bf \underline{Output:} }$\va_L.$
	\end{algorithmic}
\end{algorithm}

\begin{theorem}[Formal statement of Theorem \ref{lemma:bilinear_subspace_iteration}]
In Algorithm \ref{alg:gap-free_subspace_iteration}, if we set $n = \tilde\Theta(\frac{d^2 \lambda_1^2 }{\varepsilon^2 \lambda_k^2} ), m=d\log(n/\delta), L =\Theta(\log(d/\varepsilon)), \delta=0.1/L$, we will be able to identify an action $\hat\va$  that yield at most $\varepsilon$-regret with probability $0.9$. Therefore by applying the standard PAC to regret conversion as discussed in Claim \ref{claim:pac2regret} we get a cumulative regret of $\tilde O(\lambda_1^{1/3} k^{1/3}(\tilde\kappa dT)^{2/3})$ for large enough $T$, where $\tilde \kappa=\lambda_1/|\lambda_k|$.

On the other hand, we set $n=\tilde\Theta( \frac{d^2k^2}{\varepsilon^2} )$ and keep the other parameters. If we play Algorithm \ref{alg:gap-free_subspace_iteration} $k$ times by setting $k'=2,4,6,\cdots 2k$ and select the best output among them, we can get a gap-free cumulative regret of $\tilde O(\lambda_1^{1/3}k^{4/3} (dT)^{2/3})$ for large enough $T$ with high probability.
\end{theorem}

\begin{proof}[Proof of Theorem \ref{lemma:bilinear_subspace_iteration}]
	First we show the first setting identify an $\varepsilon$-optimal reward with $\tilde O( \tilde\kappa^2 d^2 k \epsilon^{-2} )$ samples.
	
	Similarly as Theorem 	
	\ref{thm:low-rank_subspace_iteration}, when setting $n\geq \tilde \Theta(d^2/(\sigma_k^2\tilde\epsilon^2 )) $, we can find $\mX_L$ that satisfies $\|(\mX_L \mX_L^\top -I)\mU\| \leq \tilde\epsilon $, and therefore we recover an $ \mY_{L} = \mM \mX_{L-1} + \mG_L$ with $\|\mG_L\|\leq \sigma_k \tilde\varepsilon$ and $\|\mY_L\mX_{L-1}^\top - \mM\|_2 = \|\mM\mX_{L-1}\mX_{L-1}^\top - \mM + \mG_L\mX_{L-1}^\top\|_2  \leq (\lambda_1+|\lambda_k|)\tilde\varepsilon. $ Therefore by definition of $ \va_{L}, \va_L^\top \mY_L\mX_{L-1}^\top \va_L = \max_{\|\va\|=1}\va^\top (\mM\mX_{L-1}\mX_{L-1}^\top + \mG_{L}\mX_{L-1}^\top ) \va \geq \lambda_1 - (\lambda_1 + |\lambda_k|)\tilde\varepsilon$. Therefore $\va_L^\top \mM\va_L \geq \lambda_1 - 2(\lambda_1 + |\lambda_k|)\tilde\varepsilon$. Therefore we set $2(\lambda_1 + |\lambda_k|)\tilde\varepsilon =\epsilon $, i.e., $\tilde\varepsilon = 0.5\epsilon/(\lambda_1+|\lambda_k|)$ which will get a total sample of $T= \tilde\Theta(k n) = \tilde\Theta(d^2\tilde\kappa^2 k  \varepsilon^{-2})$. Then by applying Claim \ref{claim:pac2regret} we get the cumulative regret bound.

	Next we show how to estimate the action with $\tilde O(d^2 k^4 \varepsilon^{-2} )$ samples. To achieve this result, we need to slightly alter Algorithm \ref{alg:gap-free_subspace_iteration} where we respectively set $k'=2,4,6,\cdots 2k$ and keep the best arm among the $k$ outputs. We argue that among all the choices of $k'$, at least for one $l\in [k], k'=2l$, we have $|\lambda_l|-|\lambda_{l+1}|\geq \lambda_1 /k  $.
	Notice with  similar argument as above, when we set $n=\tilde\Theta(d^2 \lambda_{l}^{-2} \tilde\varepsilon^{-2}) \leq \tilde\Theta(d^2k^2 \lambda_{1}^{-2} \tilde\varepsilon^{-2}) $ we can get $\|\mG\|\leq \tilde\varepsilon \lambda_l$ as required by Corollary \ref{coro:hardt}, the total number of iterations $L = O(\sigma_l/(\sigma_l-\sigma_{l+1})\log(2d/\epsilon) = \tilde O(k)$.
	Finally by setting $\tilde\varepsilon = \epsilon/(4\lambda_1)$ 	we get the overall samples we required is $ \tilde O(k^2 n) = \tilde O(d^2k^4 \epsilon^{-2})$.

	For both settings, directly applying our arguments in the PAC to regret conversion: Claim \ref{claim:pac2regret} will finish the proof. 
\end{proof}

\subsection{Omitted Proofs of Main Results of Low-Rank Linear Reward}
\label{appendix:low-rank}

\begin{theorem}[Formal statement of Theorem 	\ref{thm:low-rank_subspace_iteration}]
	In Algorithm \ref{alg:subspace_iteration}, for large enough constants $C_n, C_L, C_m$, let $n=C_n d^2 \log^2(d/\delta)\sigma_k^{-2}\varepsilon^{-2}$, $m=C_m d\log(n/\delta)$, and $L = C_L\log(d/\varepsilon)$,  $\mX_L$ satisfies $\|(\mI - \mX_L\mX_L^\top )\mV\|\leq \varepsilon/4,$ and the output $\hat\mA$ satisfies $ \|\hat\mA-\mA^*\|_F\leq \|\mM\|_F \varepsilon$. 
	Altogether to get an $\varepsilon$-optimal action, it is sufficient to have total sample complexity of
	$
	T \leq \tilde O( d^2 k\lambda_k^{-2} \varepsilon^{-2}).$
\end{theorem}

\begin{proof}[Proof of Theorem \ref{thm:low-rank_subspace_iteration} ]
	Let $\mM=\mV\mSigma \mV^\top$. From Claim \ref{claim:bound__G_subspace_iteration} we get that for each noisy subspace iteration step we get $\mY_l = \mM\mX_l +\mG_l$ with $5\|\mG_l\|\leq \varepsilon \sigma_k$ and $\|\mV^\top \mG\|\leq \sigma_k \sqrt{k}/3\sqrt{d}\leq \sigma_k (\sqrt{2k}-\sqrt{k})/2\sqrt{d} $. Therefore we can apply Corollary \ref{coro:hardt}, and get $\|\mV (\mX_L\mX_L^\top -\mI)\|\leq \varepsilon/4$ with $O(\log 2d/\epsilon)$ steps. Therefore we have:
	\begin{align*}
	\| \mA_L-\mM \|_F = & \|( \mM\mX_L+\mG_L)\mX_L^\top - \mM \|_F	
	= \|\mV^\top\mSigma \mV(\mX_L\mX_L^\top -\mI)) + \mG_L\mX_L^\top \|_F	\\
	\leq & \|\mM\|_F \|\mV(\mX_L\mX_L^\top -\mI)\| + \|\mG_L\|\|\mX_L\|_F\\
	 \leq& (\|\mM\|_F + \sigma_k) \varepsilon/4 < \|\mM\|_F\varepsilon/2.
	\end{align*} 	
	Meanwhile, notice $\|\mA^*\|_F = 1, \|\mM\|_F =r^* $ and $\|\hat\mA\|_F =1$. $\|\mA_L/r^* - \mA^*\|_F \leq \varepsilon/2$.
	$\|\hat\mA-\mA^*\|_F = \|\mA_L/\|\mA_L\|_F - \mA^* \|_F = \|\vecz(\mA_L)/\|\vecz(\mA_L)\|_2 - \vecz(\mA^*) \|_2$.

Write  $\theta_A:=\theta(\vecz(\mA_L), \vecz(\mA^*)$. The worst case that makes $\|\vecz(\hat\mA) - \vecz(\mA^*) \|$ to be larger than $\|\vecz(\mA_L/r^*) - \vecz(\mA^*) \|$ is when $\|\vecz(\mA_L/r^*) - \vecz(\mA^*) \| = \sin\theta_A$ and $\|\vecz(\hat\mA) - \vecz(\mA^*) \|$ is always $2\sin(\theta_A/2)$. Notice trivially $2\sin(\theta_A/2)\leq 2\sin(\theta_A)$ Therefore we could get
	$\|\hat\mA-\mA^*\|_F \leq 2\|\mA_L/r^*-\mA^*\|_F \leq \varepsilon$.

\end{proof}

\begin{proof}[Proof of Theorem \ref{thm:gap-free_low-rank}]
We find an $l$ to be the smallest integer such that $\sum_{i=l+1}^k \sigma_i^2 \leq \epsilon^2 \|\mM\|_F^2$. Then we have $\sigma_l \geq \epsilon/\sqrt{k-l} > \epsilon/\sqrt{k}$.

Notice that in Algorithm \ref{alg:subspace_iteration}, we set $n\geq \tilde\Theta(\frac{d^2 k}{ (r^*)^2 \varepsilon^4 } )$ large enough such that $ \|\mG\|_2 \leq O(\|\mM\|_F \epsilon^2/\sqrt{k} ) \lesssim \epsilon(\sigma_l-0)$ and $\|\mU^\top \mG \|_2 \leq \|\mM\|_F\epsilon/\sqrt{k}\frac{\sqrt{ k'}-\sqrt{k-1} }{2\sqrt{d}}  $. (This comes from the argument proved in Claim \ref{claim:bound__G_subspace_iteration}.)

Therefore by conducting noisy power method we get with $O(nk)=\tilde O(\frac{d^2 k^2}{ (r^*)^2 \varepsilon^4 }  )$ samples we can get an action $\hat\mA$ that satisfies:
$$ \| \mM - \mX_L\mX_L^\top \mM\|_F^2 \leq \sum_{i=l+1}^k \sigma_i^2 + l\epsilon^2 \sigma_l^2 \leq 2 \|\mM\|^2_F \epsilon^2.   $$
Therefore we could get $\|\mA^* - \hat\mA\|\leq 2\epsilon,$ and with similar argument as \eqref{eqn:low_rank_action2regret} we have $r^* - f(\hat\mA) \leq \|\mM\|_F \epsilon^2$.

Therefore if we want to take a total of $T$ actions, we will set $\epsilon^6 = \tilde\Theta(\frac{d^2k^2}{(r^*)^2 T  })$ and we get:
\begin{align*}
\mathfrak{R}(T) = & \sum_{t=1}^{T_1} r^* - f(\mA_t) + \sum_{t=T_1+1}^T r^* - f(\hat\mA) \\
< & r^* T_1 + T r^*\varepsilon^2\\
\leq & \tilde O(d^{2/3} k^{2/3} (r^*)^{1/3}T^{2/3}).
\end{align*}
\end{proof}

\subsection{Technical Details for Quadratic Reward}


\paragraph{Noisy Power Method.}

\begin{corollary}[Adapted from Corollary 1.1 from \cite{hardt2014noisy}]
	\label{coro:hardt}
	Let $k'\geq l$. Let $\mU \in \R^{d\times l}$
	represent the top $l$ singular vectors of $\mM$ and let $\sigma_1 \geq \cdots \geq \sigma_k > 0$ denote its singular values. Suppose $X_0$ is an orthonormal basis of a random $k'$-dimensional
	subspace. Further suppose that at every step of NPM we have
	\begin{align*}
	5\|\mG\|\leq & \epsilon(\sigma_l - \sigma_{l+1}),\\
	\text{and } 5\|\mU^\top \mG\| \leq & (\sigma_{l} -\sigma_{l+1})\frac{\sqrt{k'}- \sqrt{l-1}  }{ 2\sqrt{d} }	
	\end{align*}
	for some fixed parameter $\epsilon < 1/2$. Then with all but $2^{-\Omega(k'+1-l)} + e^{\Omega(d)}$ probability, there
	exists an $L = O(\frac{\sigma_l}{\sigma_l-\sigma_{l+1}}	\log(2d /\epsilon))$ so that after $L$ steps we have that $	\|(\mI - \mX_L\mX_L^\top)\mU\|\leq \epsilon$.
	
\end{corollary}

\begin{theorem}[Adapted from Theorem 2.2 from \citep{balcan2016improved}]
	\label{thm:balcan}
	Let $\mU_l \in \R^{d\times l}$
	represent the top $l$  singular vectors of $\mM$ and let $\sigma_1 \geq \cdots \geq \sigma_k > 0$ denote its singular values. Naturally $l\leq k$. Suppose $X_0$ is an orthonormal basis of a random $k'$-dimensional
	subspace where $k'\geq k$. Further suppose that at every step of NPM we have
	\begin{align*}
	\|\mG\|\leq & O(\epsilon \sigma_l),\\
	\text{and } \|\mU_k^\top \mG\|_2 \leq & O(\sigma_{l}\frac{\sqrt{k'}- \sqrt{k-1}  }{ 2\sqrt{d} })	
	\end{align*}
	for small enough $\epsilon$. Then with all but $2^{-\Omega(k'+1-k)} + e^{\Omega(d)}$ probability, there
	exists an $L = O(\log(2d /\epsilon))$ so that after $L$ steps we have that $	\|(\mI - \mX_L\mX_L^\top)\mU_l\|\leq \epsilon$.	Furthermore:
	$$  \| \mM - \mX_L\mX_L^\top \mM\|_F^2 \leq \sum_{i=l+1}^k \sigma_i^2 + l\sigma^2 \sigma_l^2. $$
\end{theorem}

\paragraph{Concentration Bounds.}

\begin{claim}
\label{claim:bound__G_subspace_iteration}	
Write the eigendecomposition for $\mM$ as $\mM=\mU\mSigma \mU^\top$. In Algorithm \ref{alg:subspace_iteration}, when $n\geq \tilde \Theta ( d^2/(\lambda_k^2 \varepsilon^2) )$, the noisy subspace iteration step can be written as:
$ \mY_l = \mM \mX_{l-1} + \mG_l  $, where the noise term satisfies:
\begin{align*}
5\|\mG_l\|\leq & \varepsilon |\lambda_k| \\
5\|\mU^\top \mG_l\| \leq & \varepsilon|\lambda_k| \frac{\sqrt{k}}{3\sqrt{d}}.
\end{align*}
with high probability for our choice of $n$.
\end{claim}
\begin{proof}

For compact notation, write vector $\veta_i:=[\eta_{i,1},\eta_{i,2},\cdots \eta_{i,k'}]^\top \in \R^{k'}$. We have:
\begin{align*}
\mG_l(s)= & \frac{m}{n}\sum_{i=1}^n (\vz_i^\top \mM \mX_l(s)) \vz_i + \frac{m}{n}\sum_{i=1}^n \eta_{i,s} \vz_i - \mM\mX_l(s), \text{ therefore}\\
\mG_l = & (\frac{m}{n}\sum_{i=1}^n[\vz_i\vz_i^\top ] - I )\mM \mX_l + \frac{m}{n}\sum_{i=1}^n  \vz_i {\veta}_i^\top.
\end{align*}
First note that for orthogonal matrix $\mX_l$, $\|\mM \mX_l\|\leq \lambda_1 $, and $\|\frac{m}{n}\sum_{i=1}^n[\vz_i\vz_i^\top ] - I \|\leq O(\sqrt{\frac{d+\log(1/\delta)}{n}})$. The bottleneck is from the second term and we will use Matrix Bernstein to concentrate it.
Write $\mS_i=\frac{m}{n}\vz_i\veta_i^\top$. We have $\|\mS_i\|\leq O(\frac{\sqrt{mk'}\log(n/\delta) }{n})$ with probability $1-\delta$ and $\E[\sum_{i}\mS_i\mS_i^\top] = \frac{mk'}{n}I_d$ and $\E[\sum_{i}\mS_i^\top \mS_i] = \frac{md}{n}I_{k'} $. Therefore with matrix Bernstein we can get that $\|\sum_i \mS_i\|_i \leq  O(\sqrt{\frac{md}{n}}\log(d/\delta)) $ with probability $1-\delta$. 

Therefore for $n\geq \tilde \Omega(d^2/(\lambda_k^2\varepsilon^2)$, we can get that $5\|\mG_l\|\leq \varepsilon|\lambda_k|$.

Similarly since $\mU^\top \vz_i \sim \cN(0, \frac{1}{m}I_{k'} )$, with the same argument one can easily get that $\|\mU^\top \mG_l\| \leq O(\sqrt{\frac{mk'}{n}}\log(d/\delta))$. Therefore with the same lower bound for $n$ one can get $15\|\mU^\top\mG_l\|\leq \varepsilon|\lambda_k|\sqrt{\frac{k}{d}}$.
\end{proof}

\subsection{Omitted Proof for RL with Quadratic Q function}

\begin{algorithm}[h]\caption{Learn policy complete polynomial with simulator.}\label{alg:simulator_tensor_policy_complete}
	\begin{algorithmic}[1]
		\State {\bf Initialize:} Set $n=\tilde\Theta(\tilde\kappa^2 d^2H^3/\varepsilon^2)$, Oracle to estimate $\hat T_h$ from noisy observations.
		\For{$h=H,\ldots 1$}
		\State Sample $\phi(s^i_h,a^i_h), i \in [n]$ from standard Gaussian $N(0,I_d)$
		\For{$i \in [n]$}
		\State Query $(s^i_h,a^i_h)$ 
		and use $\pi_{h+1}, \dots,\pi_{H}$ as the roll-out to get estimation ${\hat Q}_h^{\pi_{h+1}, \dots,\pi_{H}}(s^i_h,a^i_h)$
		\EndFor
		\State Retrieve $\hat \mM_h$ from estimation ${\hat Q}_h^{\pi_{h+1}, \dots,\pi_{H}}(s^i_h,a^i_h), i \in [n]$
		\State Set $\hat Q_h(s,a) \leftarrow  f_{\hat T_h}$
		\State Set $\pi_h(s) \leftarrow \argmax_{a \in \states}~ \hat Q_h(s,a)$
		\EndFor
		\State {\bf Return} $\pi_1,\dots,\pi_H$
	\end{algorithmic}
\end{algorithm}

\begin{proof}[Proof of Theorem \ref{thm:quadratic_rl}]
With the oracle, at horizon $H$, we can estimate $\hat \mM_H $ that is $\epsilon/H$ close to $\mM^*_H$ in spectral norm through noisy observations from the reward function with $\tilde O(\tilde\kappa^2 d^2H^2/\varepsilon^2)$ samples.
Next, for each horizon $h=H-1,H-1,\cdots,1$, sample $s'_i\sim \PP(\cdot |s,a)$, we define $\eta_i = \max_{a'}f_{\hat \mM_{h+1}}(s'_i, a') - \E_{s'\sim \PP(\cdot|s,a)} \max_{a'}f_{\hat \mM_{h+1}}(s',a')$. $\eta_i$ is mean-zero and
$O(1)$-sub-gaussian since it is bounded. Denote $\mM_h$ as the matrix that satisfies $f_{\mM_h}:=\cT f_{\hat \mM_{h+1}},$ which is well-defined due to Bellman completeness. We estimate $\hat \mM_h$ from the noisy observations $y_i = r_h(s,a) + \max_{a'}f_{\hat \mM_{h+1}}(s'_i, a') = \cT f_{\hat \mM_{h+1}} + \eta_i =: f_{\mM_{h}} + \eta_i$. Therefore with the oracle, we can estimate $\hat \mM_h$ such that $\|\hat \mM_h - \mM_{h}\|_2\leq \epsilon/H$ with $\Theta(\tilde\kappa^2 d^2k^2H^2/\epsilon^2)$ bandits.
Together we have:
\begin{align*}
\|f_{\hat\mM_h}-f_{\hat\mM^*_h}  \|_{\infty} = & \|\hat \mM_{h} -\mM^*_{h} \| \\
\leq &\|\hat \mM_{h}-\mM_{h} \| + \|\mM_h - \mM^*_h\| \\
\leq  &\epsilon/H + \|\T f_{\hat \mM_{h+1} } - \T f_{\mM^*_{h+1}} \|_{\infty} \\
\leq &  \epsilon/H + \|f_{\hat \mM_{h+1} } - f_{\mM^*_{h+1}} \|_{\infty}\\
\leq & 2\epsilon/H + \|f_{\hat \mM_{h+2} } - f_{\mM^*_{h+2}} \|_{\infty}\\
\leq & \cdots\\
\leq & (H-h)\epsilon/H.
\end{align*}
Finally for $h=1$ we have $\|\hat \mM_1 - \mM^*\|\leq \epsilon $ if we sample $n = \tilde\Theta(\tilde\kappa^2 d^2k^2H^2/\epsilon^2 )$ for each $h\in [H]$.
Therefore for all the $H$ timesteps we need $\Theta(\tilde\kappa^2 d^2k^2H^3/\epsilon^2)$.

\end{proof}

\section{Technical details for General Tensor Reward}

\subsection{Technical Details for Symmetric Setting}
\label{appendix:symmetric_tensor} 
\begin{lemma}[Zeroth order optimization for noiseless setting]
	\label{lemma:population_progress}
	For $p\geq 3$, suppose $0.5 \va^\top \vv_1>|\va^\top \vv_j|$ for all $j\geq 2$, we have: 
	\begin{align*}
	\tan\theta(G(\va), \vv_1) \leq  \frac{1}{2}  \tan \theta(\va,\vv_1). 
	\end{align*}
\end{lemma}

\begin{proof}
	We first simplify $G(\va) = \sum_{j=1}^r \lambda_j \vv_j \cdot S_j$, where 
	\begin{align*}
	G(\va)= &  \sum_{s=0}^{\floor{ (p-3)/2}} \frac{(1-\frac{1}{2p})^{p-2s-1}(\frac{1}{2p})^{2s+1}}{m^s } \binom{p}{2s+1} T(I^{\otimes s+1}\otimes \va^{\otimes p-2s-1}) \\
	= & \sum_{s=0}^{\floor{(p-3)/2}} \frac{(1-\frac{1}{2p})^{p-2s-1}(\frac{1}{2p})^{2s+1}}{m^s } \binom{p}{2s+1} \sum_{j=1}^k \lambda_j (\vv_j^\top \va)^{p-2j-1} \vv_j \\
	= & \sum_{j=1}^k  v_j  \cdot \overbrace{\lambda_j \sum_{s=0}^{\floor{ (p-3)/2}} \frac{(1-\frac{1}{2p})^{p-2s-1}(\frac{1}{2p})^{2s+1}}{m^s } \binom{p}{2s+1}  (\vv_j^\top \va)^{p-2s-1} }^{S_j:=}\\
	= & \sum_{j=1}^k S_j \vv_j. 
	\end{align*}
	Notice for even $p$, 
	\begin{align*}
	S_j = & \lambda_j (\vv_j^\top \va)^3  \cdot \sum_{s=0}^{p/2-2} \frac{(1-\frac{1}{2p})^{p-2s-1}(\frac{1}{2p})^{2s+1}}{m^s } \binom{p}{2s+1}  (\vv_j^\top \va)^{p-2s-4}\\
	= & \lambda_j (\vv_j^\top \va)^3  \cdot \sum_{r=0}^{p/2-2} \frac{(1-\frac{1}{2p})^{2r+3}(\frac{1}{2p})^{p-3-2r}}{m^{p/2-2-r} } \binom{p}{p-2r-3}  (\vv_j^\top \va)^{2r}.   \tag{let $2r = p-4-2s$}  \\
	\frac{S_j}{ \lambda_j (\vv_j^\top \va)^3} = & \sum_{r=0}^{p/2-2} \frac{(1-\frac{1}{2p})^{2r+3}(\frac{1}{2p})^{p-3-2r}}{m^{p/2-2-r}} \binom{p}{p-2r-3}  (\vv_j^\top \va)^{2r}  \tag{Divide both sides by $\lambda_j(\vv_j^\top \va)^3 $} \\
	\leq & \sum_{r=0}^{p/2-2} \frac{(1-\frac{1}{2p})^{2r+3}(\frac{1}{2p})^{p-3-2r}}{m^{p/2-2-r}} \binom{p}{p-2r-3}  (\vv_1^\top \va)^{2r}. \tag{Since the first term is constant and $|\vv_j^\top \va| \leq \vv_1^\top \va$ for $r\geq 1$} \\
	= & \frac{S_1}{ \lambda_1 (\vv_1^\top \va)^3 }.
	\end{align*} 
	
	Therefore for even $p\geq 4$:
	\begin{align}
	\label{eqn:S_relations_even}
	|S_j| \leq \frac{|\lambda_j|}{\lambda_1} \frac{|\vv_j^\top \va|^3}{|\vv_1^\top \va|^3} S_1 \leq \frac 1 4 \frac{|\vv_j^\top \va|}{|\vv_1^\top \va|} S_1, \forall j\geq 2.
	\end{align}
	Similarly for odd $p$, we have:
	\begin{align*}
	S_j = & \lambda_j (\vv_j^\top \va)^2  \cdot \sum_{s=0}^{ (p-3)/2} \frac{(1-\frac{1}{2p})^{p-2s-1}(\frac{1}{2p})^{2s+1}}{m^s} \binom{p}{2s+1}  (v_j^\top a)^{p-2s-3} \\
	= & \lambda_j (\vv_j^\top \va)^2  \cdot \sum_{r=0}^{ (p-3)/2} \frac{(1-\frac{1}{2p})^{2r+2}(\frac{1}{2p})^{p-2-2r}}{m^{(p-3)/2-r}} \binom{p}{p-2-2r}  (\vv_j^\top \va)^{2r}  \tag{Let $r = (p-3)/2- s $},\\
	\frac{S_j}{\lambda_j  (\vv_j^\top \va)^2} = & \sum_{r=0}^{ (p-3)/2} \frac{(1-\frac{1}{2p})^{2r+2}(\frac{1}{2p})^{p-2-2r}}{m^{(p-3)/2-r}} \binom{p}{p-2-2r}  (\vv_j^\top \va)^{2r}  \tag{Divide both sides by $\lambda_j(\vv_j^\top \va)^2$} \\
	\leq &  \sum_{r=0}^{ (p-3)/2} \frac{(1-\frac{1}{2p})^{2r+2}(\frac{1}{2p})^{p-2-2r}}{m^{(p-3)/2-r}} \binom{p}{p-2-2r}  (\vv_1^\top \va)^{2r} \tag{Since the first term is constant and $|\vv_j^\top \va| \leq \vv_1^\top \va$ for $r\geq 1$} \\
	= & \frac{S_1}{\lambda_1 (\vv_1^\top \va)^2}. 
	\end{align*}
	Therefore for odd $p$ we have:
	\begin{align}
	\label{eqn:S_relations_odd}
	|S_j| \leq \frac{|\lambda_j|}{\lambda_1} \frac{|\vv_j^\top \va|^2}{|\vv_1^\top \va|^2} S_1\leq \frac 1 2 \frac{|\vv_j^\top \va|}{|\vv_1^\top \va|} S_1, \forall  j\geq 2.
	\end{align}
	
	Write $\mV=[\vv_2,\vv_3,\cdots, \vv_k]\in \R^{d\times k}$ be the complement for $\vv_1$. Therefore for any $\vx$ without normalization, one can conveniently represent  $|\tan\theta(\vx,\vv_1)|$ as $\|\mV^\top \vx\|_2/|\vv_1^\top x|$. 
	
	\begin{align}
	\|\mV^\top G(\va) \|^2 = & \sum_{j=2}^k S_j^2 \\
	\leq &	\sum_{j=2}^k \frac{ |\vv_j^\top \va|^2 }{4 |\vv_1^\top \va|^2 } S_1^2 \tag{from \eqref{eqn:S_relations_odd},\eqref{eqn:S_relations_even}}  \\
	= & 
	\frac{1}{4}  \tan^2\theta(\vv_1,\va) (\vv_1^\top G(\va))^2. 
	\end{align}
	Therefore for $p\geq 3$, $\tan\theta(G(\va),\vv_1) \leq \frac{1}{2} \tan\theta(\va,\vv_1) $. 
\end{proof}


\begin{proof}[Proof of Lemma \ref{lemma:bound_g_tensor}]
	We first estimate $G_n(\va) - \E[G_n(\va)]$, which is want we want for even $p$. For odd $p$ we will need to analyze an extra bias term that is orthogonal to $\vv_1$, $\ve: = \frac{(\frac{1}{2p})^{p-1}(1-\frac{1}{2p}) p }{ m^{p/2-1} } \sum_{j=2}^k \lambda_j (\vv_j^\top \va)\vv_j	
	$; and we have $G_n(\va) - \E[G_n(\va)] = G_n(\va) - G(\va) +\ve$. 
	
	We decompose $G_n(\va)$ as $G_n(\va)=\sum_{s=1}^k G_n^{(s)} + N$, where $G_n^{(s)}: = \frac{m}{n}\sum_{i=1}^n \binom{p}{s}T(((1-0.5/p)\va)^{\otimes p-s}\otimes (\vz_i/(2p))^{\otimes s}) \vz_i$. 
	The noise term $N:=\frac{m}{n}\sum \epsilon_i \vz_i$.

	\begin{align*}
	G_n^{(s)}: = &\frac{m}{n}\sum_{i=1}^n \binom{p}{s}T(((1-0.5/p)\va)^{\otimes p-s}\otimes (\vz_i/(2p))^{\otimes s}) \vz_i \\
	= & \frac{m}{n} (1-\frac{1}{2p})^{p-s}(\frac{1}{2p})^s \binom{p}{s} \sum_{i=1}^n   \sum_{j=1}^k  \lambda_j   (\va^\top \vv_j)^{p-s} (\vz_i^\top \vv_j)^{s} \vz_i. \\ 
	\E[G_n^{(s)} ]= &  m  (1-\frac{1}{2p})^{p-s}(\frac{1}{2p})^s \binom{p}{s}  \sum_{j=1}^k  \lambda_j  (\va^\top \vv_j)^{p-s} \E [(\vz^\top \vv_j)^{s} \vz]   \\
	= & \left\{  \begin{array}{ll}
	(1-\frac{1}{2p})^{p-s}(\frac{1}{2p})^s m \binom{p}{s}  \sum_{j=1}^k \lambda_j  (\va^\top \vv_j)^{p-s} \frac{1}{m^{(s+1)/2}}(s)!! \vv_j, &\text{ for odd }s,   \\
	0, & \text{ for even }s 
	\end{array}\right.\\
	= & \left\{  \begin{array}{ll}  (1-\frac{1}{2p})^{p-s}(\frac{1}{2p})^s
	\frac{s!!}{m^{(s-1)/2}} \binom{p}{s}  \sum_{j=1}^k \lambda_j  (\va^\top \vv_j)^{p-s}  \vv_j, &\text{ for odd }s,   \\
	0, & \text{ for even }s 
	\end{array}\right.
	\end{align*}
	
	\begin{align*}
	G_n^{(s)} - \E[G_n^{(s)}]  = m (1-\frac{1}{2p})^{p-s}(\frac{1}{2p})^s \binom{p}{s} \sum_{j=1}^k  \lambda_j   (\va^\top \vv_j)^{p-s}  \vg_{n,s}(j),
	\end{align*}
	where $\vg_{n,s}(j) := \frac{1}{n}\sum_{i=1}^n (\vz_i^\top \vv_j)^{s} \vz_i  -\E[(\vz^\top \vv_j)^{s} \vz] $.

	Notice the scaling in each $G_n^{(s)}$ is $(1-\frac{1}{2p})^{p-s}(\frac{1}{2p})^s \binom{p}{s} \leq (\frac{1}{2p})^s p^s/(s!) < 2^{-s}$ decays exponentially. 
	In Claim \ref{claim:gaussian_moment_concentration} we give bounds for $g_{n,s}(j).$	We note the bound for each $g_{n,s}$ also decays with $s$. 

	Therefore the bottleneck of the upper bound mostly depend on $\vg_{n,0}$ and $\vv_1^\top \vg_{n,0}$, and we get: 
	\begin{align*}
	\|G_n-\E[G_n]\|\leq & C_1 \lambda_1\sqrt{\frac{(d+\log(1/\delta))d\log(n/\delta)}{n}} +N,\\
	|\vv_1^\top G_n - \E[\vv_1^\top G_n] | \leq & C_2 \lambda_1 \sqrt{\frac{d\log(n/\delta)(1+\log(1/\delta))}{n}} +\vv_1^\top N. 
	\end{align*}
	Next from Claim \ref{claim:noise_concentration}, the noise term  
	$$N\leq C_3\sqrt{ \frac{m\log(n/\delta)(d+\log(n/\delta))\log(d/\delta) }{n} },$$
	 $$|v_1^\top N|\leq C_4 \sqrt{m \frac{\log^2(n/\delta)\log(d/\delta) }{n} },$$
	
	Finally $\ve$ is very small: $\|\ve\|\leq \frac{1}{m^{p/2-1}(2p)^(p-1)} \lambda_2 \|\mV^\top \va\| = \lambda_2\frac{1}{m^{p/2-1}(2p)^(p-1)} \sin\theta(\va,\vv_1)$. $|\ve^\top \vv_1|=0$. 
	
	Together we can bound $G_n(\va) - G(\va)$ and finish the proof. 	
\end{proof}

\begin{proof}[Proof of Lemma \ref{lemma:iterative_progress}]
	From Lemma \ref{lemma:population_progress} we have: $|\tan\theta(G(\va),\vv_1)|\leq 1/2 |\tan\theta(\va,\vv_1)|$. Let $\mV = [\vv_2,\cdots \vv_k]$. 
	For any $p\geq 2$, we have:
	\begin{align*}
	|\tan\theta(\va^+, \vv_1)| =  & \frac{\|\mV^\top \va^+\|_2 }{|\vv_1^\top \va^+|} \\
	= & \frac{ \|\mV^\top (G(\va) + \vg)\|  }{|\vv_1^\top (G(\va)+\vg)|} \\
	\leq & \frac{ \|\mV^\top G(\va)\| + \|\mV^\top \vg\|  }{ |\vv_1^\top G(\va)| - |\vv_1^\top \vg|  }\\
	\leq & \frac{1/2 |\tan\theta(\va,\vv_1)| |\vv_1^\top G(\va)| +\|\vg\| }{ |\vv_1^\top G(\va)| - |\vv_1^\top \vg|  } \\
	= & \alpha |\tan\theta(\va,\vv_1)| \frac{S_1}{S_1 - \|\vv_1^\top \vg\|} + \frac{\|\vg\|}{ S_1 - \|\vv_1^\top \vg\|},
	\end{align*}
	where $S_1:=\vv_1^\top G(\va) = \vv_1^\top G(\va) =  \lambda_1  \sum_{s=0}^{\floor{ (p-3)/2}} \frac{(1-\frac{1}{2p})^{p-2s-1}(\frac{1}{2p})^{2s+1}}{m^s} \binom{p}{2s+1}  (\vv_1^\top \va)^{p-2s-1}\geq \lambda_1(1-\frac{1}{2p})^{p-1}(\frac{1}{2p}) p (\vv_1^\top \va)^{p-1} \geq  \frac{\lambda_1}{4} (\vv_1^\top \va)^{p-1}$. The inequality comes from keeping only the first term where $s=0$.   
	With the assumption that $|\vv_1^\top \vg|\leq 0.05 \lambda_1 (\vv_1^\top \va)^{p-1}$, we have $|\vv_1^\top \vg|\leq 0.2 S_1$. Therefore  
	\begin{align*}
	|\tan\theta(\va^+,\vv_1)| \leq & 1.25/2 |\tan\theta(\va,\vv_1)| + \frac{\|\vg\|}{ S_1 - |\vv_1^\top \vg|} \\
	\leq & 1.25/2 |\tan\theta(\va,\vv_1)| +  5/4 \frac{\|\vg\|}{S_1}\\
	\leq & 1.25/2 |\tan\theta(\va,\vv_1)| + 5 \frac{\|\vg\|}{  \lambda_1 (\vv_1^\top \va)^{p-1} }. 
	\end{align*}
	Notice when $5 \frac{\|\vg\|}{  \lambda_1 (\vv_1^\top \va)^{p-1} } \leq \max\{0.125 |\tan\theta(\va,\vv_1)|, \tilde \epsilon \}$, which will ensure $  |\tan\theta(G(\va),\vv_1) | \leq (1.25/2+0.125) |\tan\theta(G(\va),\vv_1)| +\tilde\epsilon  $. (We will prove this condition is satisfied when $\|\vg\| \leq \min\{ \frac{0.025}{p} \lambda_1 (\vv_1^\top \va)^{p-2}, 0.1 \lambda_1 \tilde \varepsilon   \}$. We will handle the additional term in the upper bound of $\|\vg\|$ later.) We divide this requirement into the following two cases. 
	On one hand, when $|\vv_1^\top \va|\leq 1-1/(p-1)$, $\|\mV^\top \va\|\geq \sqrt{1-(1-1/(p-1))^2}>1/p$, therefore $|\tan\theta(\va,\vv_1)| \geq 1/p|\vv_1^\top \va| $. Therefore 
	\begin{align*}
	& 5 \frac{\|\vg\|}{  \lambda_1 (\vv_1^\top \va)^{p-1} } \leq 0.125 |\tan\theta(\va,\vv_1)| \\
	\Leftarrow & 5 \frac{\|\vg\|}{  \lambda_1 (\vv_1^\top \va)^{p-1} } \leq   0.125/\left(p|\vv_1^\top \va|\right) \\
	\Leftrightarrow  &  \|\vg\| \leq 0.025\lambda_1 (\vv_1^\top \va)^{p-2}/p. 
	\end{align*}
	On the other hand, when $|\vv_1^\top \va|\geq 1-1/(p-1)$, $|\vv_1^\va|^{p-1}\geq 1/4$ when $p=3$. Therefore
	$5 \frac{\|\vg\|}{  \lambda_1 (\vv_1^\top \va)^{p-1} } \leq 20 \|\vg\| /\lambda_1 $. Therefore we will need $ \|\vg\|\leq 0.05\lambda_1 \tilde \epsilon $, and then the requirement that $
	5 \frac{\|\vg\|}{  \lambda_1 (\vv_1^\top \va)^{p-1} } \leq \tilde \epsilon$ is satisfied. 
	
	Altogether in both cases we have:  $ |\tan\theta(G(\va),\vv_1) | \leq 0.75 |\tan\theta(G(\va),\vv_1)| +\tilde\epsilon  $. Finally if we additionally increase  $\|\vg\|$ by $0.05\lambda_1 (\vv_1^\top \va)^{p-1}$ we will have:
	$ |\tan\theta(G(\va),\vv_1) | \leq 0.8 |\tan\theta(G(\va),\vv_1)| +\tilde\epsilon  $. 
\end{proof}
\begin{proof}[Proof of Corollary \ref{coro:tensor_regret_bound_symmetric} ]
	As shown in Theorem \ref{thm:tensor_staged_progress} at least one action $\va$ in $\cA_S, |\cA_S|\leq \tilde O(k) $ satisfies $\tan\theta(\va,\va^*)\leq \varepsilon$ with a total of $\tilde O(\frac{d^pk}{\lambda_1^2 \varepsilon^2}  )$ steps. Therefore with Claim \ref{claim:angle_to_regret} we have to get $\tilde \varepsilon$-optimal reward we need $\tilde O(\frac{d^pk}{\lambda_1 \tilde\varepsilon}  )$ steps. Notice the eluder dimension for symmetric polynomials is $d^p$ and the size of $\cA_S$ is at most $\tilde O(k)$. Then by applying Corollary \ref{coro:eluderUCB} we get that the total regret is at most $\tilde O(\sqrt{d^p k T} + \sqrt{ |\cA_S| T} ) =  \tilde O(\sqrt{d^p k T})$. 
\end{proof}

\subsection{PAC to Regret Bound Relation.}

\begin{claim}[Connecting angle to regret]
	\label{claim:angle_to_regret}
	When $0<\tan\theta(\va,\vv_1)\leq \zeta,$ we have regret $ r^* - r(\va)\leq r^* \min\{2, p\zeta^2\}  $. 
\end{claim}
\begin{proof}
	
	\begin{align*}
	|\cos\theta(\va,\vv_1)| = &  |\va^\top \vv_1|=:b ,\\ 
	|\tan\theta(\va,\vv_1)| = & \frac{\sqrt{1-b^2}}{b} \leq \zeta 
	\Leftrightarrow  b \geq  \frac{1}{\sqrt{\zeta^2+1}}. \\
	\Rightarrow  r^* - r(\va) \leq &  \lambda_1 -  \lambda_1 b^p \\
	\leq & \lambda_1(1-(\zeta^2+1)^{-p/2})\\
	= & \lambda_1 \frac{(\zeta^2+1)^{p/2} -1 }{(\zeta^2+1)^{p/2} } \\
	\leq & \lambda_1 ((\zeta^2+1)^{p/2} -1)  \tag{since denominator $(\zeta^2+1)^{p/2}\geq 1 $} \\
	\leq & \lambda_1 p\zeta^2, \text{ when }\zeta^2\leq 1/p. 
	\end{align*}	
	Additionally by definition $r^* - r(\va)\leq \lambda_1-(-\lambda_1) = 2\lambda_1 $ and thus $r^* - r(a)\leq \lambda_1\min\{2,p\zeta^2\} $. 
	We now derive the last inequality. When $\zeta\geq 1/p$ it is trivially true. When $\zeta\leq 1/p$, we have $(1+\zeta^2)^{p/2}\leq 1+p\zeta^2$ for any $p\geq 2$. Since the LHS is a convex function for $\zeta$ when $p\geq 2$ and when $\zeta=0$ LHS=RHS and when $\zeta^2=1/p$ LHS is always smaller than RHS (=2). 

Notice the argument is straightforward to extend to the setting where the angle is between $\va$ and subspace $V_1$ that satisfies $\forall \vv \in \mV_1, T(\vv)\geq \lambda_1-\epsilon $, then one also get $ r^*-r(\va) \leq \lambda_1-(\lambda_1-\epsilon)b^p \leq \min\{\lambda_1,\lambda_1p\zeta^2 + \epsilon b^p\} \leq \min\{\lambda_1,\lambda_1p\zeta^2 + \epsilon\}. $ 	
\end{proof}

\begin{claim}[Connecting PAC to Cumulative Regret]
	\label{claim:pac2regret}
	Suppose we have an algorithm \textit{alg}$(\zeta)$ that finds $\zeta$-optimal action $\hat \va$ that satisfies $0<\tan\theta(\va,\vv_1)\leq \zeta$ by taking $A \zeta^{-a} $ actions. Here $A$ can depend on any parameters such as $d,\lambda_1,$ probability error $\delta$, etc., that are  not $\zeta$. Then for large enough $T$, by calling \textit{alg} with $\zeta=A^{\frac{1}{a+2}  }T^{-\frac{1}{a+2} } p ^{-\frac{1}{a+2} }  $ and playing its output action $\hat\va$ for the remaining actions,
	 one can get a cumulative regret of:
	$$ \mathfrak{R}(T)  \lesssim T^{\frac{a}{a+2}}p^{\frac{a}{a+2}} A^{\frac{2}{a+2}} r^* .$$
	
Similarly, if an oracle finds $\varepsilon$-optimal action $\hat\va$ that satisfies $ r^*-r(\va) \leq \varepsilon$ with $B \varepsilon^{-b}  $ samples, then by setting $\varepsilon = (Br^*/T)^{\frac{1}{1+b}}$, and playing the output arm for the remaining actions, one can get cumulative regret of:
$$ \mathfrak{R}(T)  \lesssim B^{\frac{1}{1+b}} T^{\frac{b}{1+b}}r^{\frac{1}{1+b}}. $$

\end{claim}
\begin{proof}
	For the chosen $ \zeta$, write $T_1 = A\zeta^{-a}$ be the number of actions that finds $\zeta$-optimal action. Therefore $T_1= A^{\frac{2}{a+2}  }T^{\frac{a}{a+2} } p ^{\frac{a}{a+2} }$. First, when $T\geq A p^{a/2}$, $\zeta^2 \leq 1/p$, namely $r^*-r(\va) \leq r^* p \zeta^2$. We have:
	\begin{align*}
	\mathfrak{R}(T) \leq &   \sum_{t=1}^{T_1}  2r^* + \sum_{t=T_1+1}^T r^* p \zeta^2 \\
	\leq & 2r^* T_1 + T r^*p\zeta^2 \\
	\leq & 3 T^{\frac{a}{a+2}}p^{\frac{a}{a+2}} A^{\frac{2}{a+2}} r^*. 
	\end{align*}
	When $T< A p^{a/2}$, it trivially holds that $\mathfrak{R}(T) \leq 2r^*T <  2 T^{\frac{a}{a+2}}p^{\frac{a}{a+2}} A^{\frac{2}{a+2}} r^*$. 
\end{proof}

\begin{algorithm}
	\caption{UCB (Algorithm 1 in Section 5 of \cite{agarwal2019reinforcement}) }
	\label{alg:UCB}	
	\begin{algorithmic}[1]
		\State {\bf \underline{Input:} } Stochastic reward function $f$, failure probability $\delta$, action set $\cA$ with finite size $K$.
		\For {$t$ from $1$ to $T-1-K$}
		\State Execute arm $I_t = \argmax_{i\in [K]} \left(\hat\mu^t(i) + \sqrt{\frac{\log(TK/\delta)}{N^t(i)} }  \right) $. Here $N^t(\va) = 1 +\sum_{i=1}^t \mathbf{1}\{I_i=\va \}$; and $\hat\mu^t(\va) = \frac{1}{N^t(\va)} \left(r_a + \sum_{i=1}^t\mathbf{1}\{I_i=\va\}r_i \right).$
		\State Observe $r_{I_t}$
		\EndFor
	\end{algorithmic}
\end{algorithm}
\begin{theorem}[Theorem 5.1 from \cite{agarwal2019reinforcement}]
\label{thm:UCB}
	With UCB algorithm on action set with size $K$, we have with probability $1-\delta$, 
	\begin{align*}
	\mathfrak{R}(T)= \tilde O( \min\{ \sqrt{KT} \} + K  ).
	\end{align*}
\end{theorem}

\begin{corollary}
	\label{coro:eluderUCB}
With the same setting of Claim \ref{claim:pac2regret}, except that now the algorithm \textit{alg}$(\varepsilon)$ finds \textbf{a set} $\cA$ of size $S$ where at least one action $\va\in \cA$ satisfies $r^*-f(\va)\leq \varepsilon$. Then all argument in Claim \ref{claim:pac2regret} still hold by adding $\tilde O(\sqrt{ST})$ on the RHS of each regret bound.
\end{corollary}
\begin{proof}
Suppose we run \textit{alg} for $T_1$ steps and achieve $\varepsilon$-optimal reward. 

Let $r_{\varepsilon} :=\max_{\va\in \cA}f(\va) $. Therefore with UCB on mutiarm bandit we have:
$\sum_{t=T_1+1}^T r_{\varepsilon} - f(\va_t)  \leq \tilde O(\sqrt{ST})$ by Theorem \ref{thm:UCB}.

From the statement $r_{\varepsilon} \geq r^*-\zeta$. Therefore $\sum_{t=T_1+1}^T r^* - f(\va_t)  \leq \tilde O(\sqrt{S T}) + \varepsilon (T-T_1)$. Therefore
\begin{align*}
\mathfrak{R}(T) \leq &   \sum_{t=1}^{T_1}  2r^* + \varepsilon(T-T_1) +  \tilde O(\sqrt{S T}).
\end{align*}
With the same choices of $T_1$ in Claim \ref{claim:pac2regret}, the same conclusion still holds with an additional term of $\tilde O(\sqrt{S T})$. 
\end{proof}
For symmetric tensor problems the set size is $\tilde O(k)$ and therefore we will have an additional $\sqrt{k T}$ term which will be subsumed in our regret bound. 

\subsection{Variance and Noise Concentration}
\begin{lemma}[Vector Bernstein; adapted from Theorem 7.3.1 in \cite{tropp2012user}]
	\label{lemma:vector_bernstein}
	Consider a finite sequence $\{ \vx_k \}_{k=1}^n$ be i.i.d randomly generated samples, $x_k\in \R^d$, and assume that $\E[\vx_k] = 0$, $\|\vx_k\|\leq L$, and covariance matrix of $x_k$ is $\Sigma$. Then it satisfies that when $n\geq \log d/\delta$, we have:
	\begin{align*}
	\left\|	\frac{\sum_{i=1}^n \vx_i}{n}\right\| \leq C \sqrt{\frac{ (\|\Sigma\| + L^2) \log d/\delta}{n}},
	\end{align*}
	with probability $1-\delta$. 
\end{lemma}

\begin{claim}[Noise concentration]
	\label{claim:noise_concentration} 
	Let independent samples $ \vz_i\sim \cN(0,1/m I_d) $ and $\epsilon_i \sim \cN(0,1)$. With probability $1-\delta, \delta\in (0,1)$:
	\begin{align*}
	\left\|\frac{m}{n} \sum_{i=1}^n \epsilon_i \vz_i \right\| \leq  & C\sqrt{ \frac{m\log(n/\delta)(d+\log(n/\delta))\log(d/\delta) }{n} }\\
\left|\frac{m}{n} \sum_{i=1}^n \epsilon_i \vz_i^\top \vv_1 \right| \leq  & 
 C'\sqrt{ \frac{m\log^2(n/\delta)\log(d/\delta) }{n} }.
	\end{align*}	
\end{claim}
\begin{proof}
	We use the Vector Bernstein Lemma \ref{lemma:vector_bernstein}.  
	The covariance matrix for $\vx_i = \epsilon_i \vz_i$ satisfies $\E[\vx_i\vx_i^\top] = 1/m I_d$. $\vx_i$ is mean zero. $\|\epsilon_i \vz_i\|^2 = \epsilon_i^2 \|\vz_i\|^2$. Notice $\epsilon_i^2 \sim \chi(1) \lesssim 1+\log(1/\delta)$ and $m \vz_i^\top \vz_i\sim \chi(d) \lesssim d+\log(1/\delta)$. Therefore by directly applying Vector Bernstein $  \|\epsilon_i \vz_i\|\leq c \sqrt{\frac{(1+\log(1/\delta))(d+\log(1/\delta)) }{m}  } $ with probability $1-\delta$. By union bound we have: for all $i$, $\|\epsilon_i \vz_i\|\leq c \sqrt{\frac{\log(n/\delta)(d+\log(n/\delta)) }{m}  }$ with probability $1-\delta$. Therefore 
	\begin{align*}
	\left\|\frac{1}{n} \sum_{i=1}^n \epsilon_i \vz_i \right\| \leq   C\sqrt{ \frac{\log(n/\delta)(d+\log(n/\delta))\log(d/\delta) }{mn} },
	\end{align*}
	with probability $1-\delta$. 
	Similarly 
	\begin{align*}
	\left|\frac{1}{n} \sum_{i=1}^n \epsilon_i \vz_i^\top \vv_1 \right| \leq  & C\sqrt{ \frac{\log(n/\delta)(1+\log(n/\delta))\log(d/\delta) }{mn} }\\
	= & C'\sqrt{ \frac{\log^2(n/\delta)\log(d/\delta) }{mn} },
	\end{align*}

\end{proof}

\begin{claim}
\label{claim:gaussian_moment_concentration}
	Let $\{\vz_i\}_{i=1}^n$ be i.i.d samples from $\cN(0,1/m I_d)$. 
	Let $g_{n,s}(j) := \frac{1}{n}\sum_{i=1}^n (\vz_i^\top \vv_j)^{s} \vz_i  -\E[(\vz^\top \vv_j)^{s} \vz] $. We have:
\begin{align*}
\|g_{n,0}(j)\|\lesssim & \sqrt{\frac{d+\log(1/\delta)}{nm} },\\
|\vv_1^\top g_{n,0}(j)| \lesssim & \sqrt{\frac{1+\log(1/\delta)}{nm} }, \\
|\vv_1^\top g_{n,1}(j)|\leq  \|g_{n,1}(j)\| \lesssim & \sqrt{\frac{ d+\log(1/\delta)}{m^2n} }, \text{ when }n\geq d\log(1/\delta), \\
|\vv_1^\top g_{n,s}(j)|\leq  \|g_{n,s}(j)\| \lesssim & \sqrt{\frac{ \log(d/\delta)}{d^sn} }, \text{ when }n\geq \log(d/\delta),  m\geq c_0d\log(n/\delta), s\geq 2. 
\end{align*}	
	For any $j\in [k]$. 
\end{claim}
We mostly care about the correct concentration for smaller $s$. For larger $s$ a very loose bound will already suffice our requirement. 
\begin{proof}[Proof of Claim \ref{claim:gaussian_moment_concentration}]
For $s=0$, $nm\|\frac{1}{n}\sum_{i=1}^n \vz_i\|^2\sim \chi(d)$, therefore $\|\frac{1}{n}\sum_{i=1}^n \vz_i\| \lesssim \sqrt{\frac{d+\log(1/\delta)}{nm} }$.
$nm(\frac{1}{n}\sum_{i=1}^n \vz_i^\top \vv_1)^2\sim \chi(1)$. Therefore 
$|\frac{1}{n}\sum_{i=1}^n \vz_i^\top \vv_1| \lesssim \sqrt{\frac{1+\log(1/\delta)}{nm} }$.

For $s=1$, due to standard concentration for covariance matrices (see e.g. \cite{tropp2012user,du2020few} ), we have:
\begin{align*}
m\|(\frac{1}{n}\sum_{i=1}^n \vz_i\vz_i^\top - \E[\vz\vz^\top]) \| \leq \max\{\sqrt{\frac{ d+\log(2/\delta) }{n} },  \frac{ d+\log(2/\delta) }{n}  \}. 
\end{align*}
Therefore when $n\geq d\log(1/\delta)$, both results
\begin{align*}
\|g_{n,1}(j)\| \lesssim & \sqrt{\frac{ d+\log(1/\delta) }{m^2 n} }\|\vv_j\|,\\
= & \sqrt{\frac{ d+\log(1/\delta) }{m^2 n} }, \text{ and}\\
 \|\vv_1^\top g_{n,1}(j)\| \lesssim & \sqrt{\frac{ d+\log(1/\delta) }{m^2 n} } \|\vv_1\|\|\vv_j\|\\
 = & \sqrt{\frac{ d+\log(1/\delta) }{m^2 n} }
\end{align*}
hold.

For larger $s\geq 2$, 
with probability $1-\delta$, $|\vz_i^\top \vv_j|\leq C \sqrt{\log(n/\delta)/m}= Cc_0/\sqrt{d} \leq 1/\sqrt{d} $. When $m\geq c_0 d\log(n/\delta)$, for small enough $c_0$ we have $|\vz_i^\top \vv_j|\leq 1/\sqrt{d}$ and $\|\vz_i\|\leq 1$ for all $i\in [n]$. Therefore $\|(\vz_i^\top \vv_j)^{s} \vz_i\|\leq d^{-s/2} $ We can use vector Bernstein, i.e., Lemma \ref{lemma:vector_bernstein} to get:
\begin{align*}
\|g_{n,s}(j)\|\leq & C_1 \sqrt{\frac{\log(d/\delta) }{d^{s}n}}.
\end{align*}
Therefore we have:
\begin{align*}
|g_{n,s}(j)^\top \vv_1| \leq & C_1 \sqrt{\frac{\log(d/\delta) }{d^{s}n}}.
\end{align*}

\end{proof}

\subsection{Omitted Details for Asymmetric Tensors} 
\label{sec:proof_asymmetric_tensor}

\begin{algorithm}
	\caption{Phased elimination with alternating tensor product.}
	\label{alg:asymmetric_tensor} 
	\begin{algorithmic}[1]
		\State {\bf \underline{Input:} }  Stochastic reward $r:(B_1^{d})^{\otimes p}\rightarrow \R$ of polynomial degree $p$, failure probability $\delta$, error $\varepsilon$. 
		\State {\bf \underline{Initialization:} }  $L_0 = C_L k\log(1/\delta) $;  Total number of stages $S = C_S\lceil \log(1/\varepsilon)\rceil+1 $,   $\cA_0 = \{\va_0^{(1)}, \va_0^{(2)}, \cdots \va_0^{(L_0)} \}\subset (B_1^{d})^{\otimes p}$ where each $\va_0^{(l)}(j),j\in [p]$ is uniformly sampled on the unit sphere $\mathbb{S}^{d-1}$.  $\tilde\varepsilon_0 = 1$. 
		\For {$s$ from $1$ to $S$}
		\State $ \tilde\varepsilon_s \leftarrow \tilde \varepsilon_{s-1}/2$, $n_s \leftarrow C_n d^p\log(d/\delta)/\tilde\varepsilon_s^2 $, $n_s\leftarrow n_s\cdot \log^3(n_s/\delta)$, $m_s\leftarrow C_m d\log(n/\delta)$, $\cA_s=\varnothing$.
		\For {$ l$ from  $1$ to $L_{s-1}$ }
		\State {\bf \underline{Tensor product update:} }
		\State Locate current arm $\tilde \va = \va_{s-1}^{(l)}$.
		\For {$\lceil (\lambda_1/\Delta) \log(2d) \rceil$ times }
		\For {$j$ from $1$ to $p$}
		\State Sample $\vz_i\sim \cN(0,1/m_s I_d), i=1,2,\cdots n_s$. 
		\State Calculate tentative arm $\va_i\leftarrow \tilde \va, \va_i(j) = (1-\tilde\varepsilon_s )\tilde \va(j) + \tilde\varepsilon_s \vz_i$
		\State Conduct estimation $\vy \leftarrow 1/n_s \sum_{i=1}^{n_s} r_{\epsilon_i}(\va_i)\vz_i $. 
		\State Update the current arm $\tilde \va(j)\leftarrow \vy/\|\vy\|$. 
		\EndFor
		\EndFor  
		\State Estimate the expected reward for $\tilde \va$ through $n_s$ samples: $r_n = 1/n_s \sum_{i=1}^{n_s} r_{\epsilon_i}(\tilde \va) $.
		\State {\bf \underline{Candidate Elimination:} }
		\If {$r_n\geq \lambda_1(1-p\tilde \varepsilon_s)$}
		\State Keep the arm $ \cA_s\leftarrow \cA_s\cup \{\tilde \va \}$ 
		\EndIf 
		\EndFor 
		\State Label the arms: $L_s = |\cA_s|, \cA_t =: \{\va_s^{(1)},\cdots  \va_s^{(L_s)} \}$.
		\EndFor
	\State Run Algorithm \ref{alg:UCB} with $\cA_S$.
	\end{algorithmic} 
\end{algorithm} 

\begin{lemma}[Asymmetric Tensor Initialization]
	With probability $1-\delta$, with $L=\tilde\Theta((2k)^p \log^p(p/\delta))$ random initializations $\cA_0 = \{ \va_0^{(0)},\va_0^{(1)},\cdots \va_0^{(L)} \}$, there exists an initialization $\va_0 \in \cA_0$ that satisfies:
	\begin{align}
\label{eqn:tensor_asymmetric_biased}
	\alpha \va_0(s)^\top \vv_1^{(s)}\geq |\va_0(s)^\top \vv_j^{(s)}|, \forall j\geq 2 \& j\in [k], \forall s\in [p], \\
	\notag 
	\va_0(s)^\top \vv_1^{(s)} \geq 1/\sqrt{d}. 
	\end{align}
	with some constant $\alpha<1$.
\end{lemma}
\begin{proof}
	This lemma simply comes from applying Lemma \ref{lemma:tensor_initialization} for $p$ times and we need $\geq 2k \log_2(p\delta)$ to ensure the condition for each $\va_0(s),s\in [p]$ holds. Therefore together we will need $(2k \log_2(p/\delta))^p$ samples. 
\end{proof}

\begin{lemma}[Asymmetric tensor progress]
	\label{lemma:asymmetric_tensor_progress}
For each $a$ that satisfies Eqn. \eqref{eqn:tensor_asymmetric_biased} with constant $\alpha<1$, we have:
\begin{align*}
\tan\theta(\mT(\va(1),\cdots \va(s-1),\mI,\va(s+1),\cdots \va(p)),\vv_1^{(s)}) \leq \alpha \tan\theta(\va_j, \vv_1^{(j)}),   
\end{align*}	
for any $ j$ that is in $[p]$ but is not $s.$ When $n\geq \Theta( d^p\log(d/\delta)\log^3(n/\delta)/\tilde\varepsilon^2 ) $ and $m = \Theta(d\log(n/\delta))$, we have:
\begin{align*}
& \tan\theta(\mT(\va(1),\cdots \va(s-1),\mI,\va(s+1),\cdots \va(p)),\vv_1^{(s)})\\
 \leq & (1+\alpha)/2 \tan\theta(\va_j, \vv_1^{(j)}) + \tilde \varepsilon, \forall j\in [p] \& j\neq s.  
\end{align*}
\end{lemma} 
The remaining proof is a simpler version for the symmetric tensor setting on conducting noisy power method with the good initialization and iterative progress.  


Finally due to the good initialization that satisfies \eqref{eqn:tensor_asymmetric_biased} and together with Lemma \ref{lemma:asymmetric_tensor_progress} we can finish the proof for Theorem \ref{thm:asymmetric_tensor_bound}.

\section{Proof of Theorem \ref{thm:poly_noiseless}}
\renewcommand{\citet}{\citep}
\subsection{Additional Notations}

Here, we briefly introduce complex and real algebraic geometry. This section is based on \citep{milneAG, shafarevich2013basic, bochnak2013real, wang2019generalized}. 

An (affine) \textbf{algebraic variety} is the common zero loci of a set of polynomials, defined as $V = Z(S) = \{\vx \in \sC^n : f(\vx) = 0, \forall f \in S\} \subseteq \sA^n = \sC^n$ for some $S \subseteq \sC[x_1, \cdots, x_n]$. A \textbf{projective variety} $U$ is a subset of $\sP^n = (\sC^{n+1} \setminus \{0\}) / \sim$, where $(x_0, \cdots, x_n) \sim k(x_0, \cdots, x_n)$ for $k \ne 0$ and $S$ is a set of homogeneous polynomials of $(n+1)$ variables.

For an affine variety $V$, its \textbf{projectivization} is the variety $\sP(V)= \{[\vx] : x \in V\} \subseteq \sP^{n-1}$, where $[\vx]$ is the line corresponding to $\vx$.

The \textbf{Zariski topology} is the topology generated by taking all varieties to be the closed sets.

A set is \textbf{irreducible} if it is not the union of two proper closed subsets.

A \textbf{variety is irreducible} if and only if it is irreducible under the Zariski topology.

The \textbf{algebraic dimension} $d = \dim V$ of a variety $V$ is defined as the length of the longest chain $V_0 \subset V_1 \subset \cdots \subset V_d = V$, such that each $V_i$ is irreducible.

A variety $V$ is said to be \textbf{admissible} to a set of linear functions $\{\ell_\alpha : \sC^d \to \sC\}_{\alpha \in I}\}$, if for every $\ell_\alpha$, we have $\dim (V\cap \{\vx \in \sC^d : \ell_\alpha(\vx)= 0\}) < \dim V$. 

A map $f = (f_1, \cdots, f_m) : \sA^n \to \sA^m$ is \textbf{regular} if each $f_i$ is a polynomial.

A \textbf{algebraic set} is the common real zero loci of a set of polynomials.

For a complex variety $V \subseteq \sA^n$, its real points form a algebraic set $V_\sR$. 

For an algebraic set $V_\sR$, its real dimension $d = \dim_\sR V_\sR$ is the maximum number $d$ such that $V_\sR$ is locally semi-algebraically homeomorphic to the unit cube $(0, 1)^d$, details can be found in \citep{bochnak2013real}.

\subsection{Proof of Sample Complexity}




\begin{lemma}[\citet{wang2019generalized}, Theorem 3.2] \label{lem:32} For $i = 1, \ldots, T$, let $L_i : \sC^n \times \sC^m \to \sC$ be bilinear functions and $V_i$ be varieties given by homogeneous polynomials in $\sC^n$. Let $V = V_1 \times \cdots \times V_T \subseteq (\sC^n)^N$. Let $W \subseteq \sC^m$ be a variety given by homogeneous polynomials. In addition, we assume $V_i$ is admissible with respect to the linear functions $\{f^{\vw}(\cdot) = L_i(\cdot, \vw) : \vw \in W\setminus \{0\}\}$. When $T \ge \dim W$, let $\delta = T - \dim W + 1 \ge 1$.  Then there exists a subvariety $Z \subseteq V$ with $\dim Z \le \dim V - \delta$ such that for any $(\vx_1, \ldots, \vx_T) \in V \setminus Z$ and $\vw \in W$, if $L_1(\vx_1, \vw) = \cdots = L_T(\vx_T, \vw) = 0$, then $\vw = 0$.
\end{lemma}

\begin{lemma}[\citet{wang2019generalized}, Lemma 3.1] Let $V$ be an algebraic variety in $\sC^d$. Then $\dim_\sR V_\sR \le \dim V$.
\end{lemma}

\begin{lemma} \label{lem:inj} Let $W$ be a vector space. For vectors $\vx_1, \cdots, \vx_T$, if the map $f : \vw \mapsto (\langle \vx_1, w \rangle, \ldots, \langle \vx_T, \vw \rangle)$ is not injective over $W-W := \{\vw_1 - \vw_2 : \vw_1, \vw_2 \in W\}$, then there exists $\vv \in W$ such that $f(\vv) = 0$.
\end{lemma}

\begin{proof} Suppose $f(\vw_1) = f(\vw_2)$. Let $\vv = \vw_1 - \vw_2$. Then $\vv \in W-W$ and $f(\vv) = f(\vw_1) - f(\vw_2) = 0$.
\end{proof}

\begin{definition}[Tensorization] Let $f$ be a polynomial of $x_1, \cdots, x_d$ with degree $\deg f \le p$. Then every $p$-tensor $\tW_f$ satisfying $\langle \tW_f, \tX_\vx\rangle = f(\vx)$ is said to be a \emph{tensorization} of the polynomial $f$, where $\tX_\vx$ is the tensorization of $\vx$ itself: 
    \begin{align}
        \tX_\vx = \mqty(1 \\ \vx)^{\otimes p}.
    \end{align}

    Let $\gF$ be a class of polynomials. A variety of tensorization of $\gF$ is defined to be an irredicuble closed variety defined by homogeneous polynomials $W$, such that for every $f\in \gF$, there is a tensorization $\tW_f$ of $f$, such that $W \ni \tW_f$ contains its tensorization. Note that neither tensorization of $f$ nor variety of tensorization of $\gF$ is unique.

    We define the variety of tensorization of $\vx$ as follows. (Note that this is uniquely defined.) Consider the regular map 
    \begin{align}
        \varphi_1: \sC^d \to \sC^{(d+1)^p}, \qquad \vx \mapsto \mqty(1 \\ \vx)^{\otimes p}, \label{eq:phi1}
    \end{align}
    the tensorization of $\vx$ is defined as $V_i = \sP(\overline{\Im \varphi_1})$. 
    \label{defn:tensor}
\end{definition}

Note that $V_i$ is irreducible because $\varphi_1$ is regular and $\sC^d$ is irreducible. By \citet[Theorem 9.9]{milneAG}, its dimension is given by 
\begin{align}
    \dim V_i \le \dim \overline{\Im \varphi_1} + 1 \le \dim \sC^d + 1 = d + 1.
\end{align}

\begin{lemma} \label{lem:admin} For any non-zero polynomial $f \ne 0$ with $\deg f \le p$. Let $W_f$ be a tensorization of $f$. Then $V_i$ is admissible with respect to $\{L_i(\cdot) = \langle \cdot, W_f\rangle\}$. 
\end{lemma}

\begin{proof} Since $V_i$ is irredicuble and $L_i$ is a linear function, it suffices to verify that $\langle \tX_\vx, W_f\rangle \ne 0$ \citep{wang2019generalized}. But according to Definition \ref{defn:tensor}, $\langle \tX_\vx, W_f\rangle \ne 0$ is equivalent to 
    \begin{align}
        f(\vx) = \left \langle \tW_f, \mqty(1 \\ \vx)^{\otimes p} \right \rangle \ne 0.
    \end{align}
    Since $f \ne 0$, we must have $f(\vx) \ne 0$ for some $\vx$, which gives a non-zero $\tX_\vx \ne 0$ for the above equation: $\langle \tX_x, W_f \rangle \ne 0$, and we conclude that $V_i$ is admissible.
\end{proof}

\begin{lemma} \label{lem:leb} Let $V \subset \sC^n$ be a (Zariski) closed proper subset, $V \ne \sC^n$. Then $V$ is a null set, i.e. it has (Lebesgue) measure zero.
\end{lemma}

\begin{proof} Suppose $V = Z(S)$ is the vanishing set for some $S \subseteq \sC[x_1, \cdots, x_n]$. Since $V \ne \sC^n$, let $f \in S$, we have $V \subseteq Z(f)$, so it suffices to show $\mathrm{Leb}(Z(f)) = 0$, which is because $Z(f) = f^{-1}(0), \mathrm{Leb}(\{0\}) = 0, f$ is a continuous function (under Euclidean topology), and $\mathrm{Leb}(\{\vx : \nabla{f}(\vx) = 0\}) = 0$.
\end{proof}

\begin{theorem} \label{thm:main0} Assume that the reward function class is a class of polynomials $\gF$. Let $W$ be (one of) its variety of tensorization. If we sample $T \ge \dim W$ times, and the sample points satisfying $(\vx_1, \cdots, \vx_T)\in (\sC^d)^T \setminus Z$ for some null set $Z$. Then we can uniquely determine the reward function $f$ from the observed rewards $(f(\vx_1), \cdots, f(\vx_T))$.
\end{theorem}

\begin{proof} Let $n = m = (d+1)^p, L_i(\vx, \vw) = \langle \vx, \vw \rangle, V = V_1 \times \cdots \times V_T$, where $V_i$ is as in Definition \ref{defn:tensor}. By \citet[Example 1.33]{shafarevich2013basic}, we have $\dim V \le (d+1)T$. Since $W$ is a vareity of tensorization, by Lemma \ref{lem:admin}, $V_i$ is admissible with respect to $\{L_i(\cdot, \tW) : \tW \in W\}$. 

    We are now ready to apply Lemma \ref{lem:32}, which gives that when $T \ge \dim W$, there exists subvariety $Z \subset V$ with $\dim Z < \dim V \le rT$, and for any $(\tX_1, \cdots, \tX_T) \in V \setminus Z$ and any $\tW \in W$, if $\langle \tX_1, \tW \rangle = \cdots = \langle \tX_T, \tW \rangle = 0$, then $\tW = 0$. By Lemma \ref{lem:inj}, we have for every $(\tX_1, \cdots, \tX_T) \in V \setminus Z$, the map $\tW \mapsto (\langle \tX_1, \tW \rangle, \cdots, \langle \tX_T, \tW \rangle)$ is injective, so $\tW_f$ and thus $f$ can be uniquely recovered from the observed rewards. 

    Finally, we show that $(\varphi_1^{-1} \times \cdots \times \varphi_1^{-1})(Z)$ is a null set, where $\varphi_1$ is as in (\ref{eq:phi1}). According to the proof of Lemma \ref{lem:32} by \citet{wang2019generalized}, we find that $Z$ is also defined by homogeneous polynomials. We take the slice $Z' = \{\vx \in Z : x_{11} = \cdots = x_{T1} = 1\}, V' = \{\vx \in V : x_{11} =\cdots = x_{T1} = 1\}$, (here $x_{ij}$ is the $j$-th coordinate of $\vx_i$), then $Z', V'$ are varieties. Since $\dim Z < \dim V$, we have $\dim Z' = \dim Z - T < \dim V - T = \dim V'$ and $Z' \subset V'$. 

    Now consider the regular map $\varphi_1' : V' \to (\sC^d)^T$, 
    \begin{align}
        \left(\mqty(1 \\ \vx_1)^{\otimes p}, \cdots, \mqty(1 \\ \vx_1)^{\otimes p} \right) \mapsto (\vx_1, \cdots, \vx_T).
    \end{align}
    Then $\varphi'_1(Z'), \varphi'_1(V')$ are both varieties. By \citet[Lemma 9.9]{milneAG}, we have $\dim \overline{\varphi'_1(Z')} \le \dim Z$. Since $\varphi'_1(V) = (\sC^d)^T$ and $\dim \varphi'_1(V') \le \dim V' = \dim V - T \le (d+1)T - T$, we have $\dim V = \dim \varphi'_1(V) = dT$ and as a result, $\dim \overline{\varphi_1'(Z)} \le \dim Z < \dim V = dT$. By Lemma \ref{lem:leb}, $\overline{\varphi'_1(Z)}$ is a null set. Since $(\vx_1, \cdots, \vx_T)\notin \overline{\varphi'_4(Z)}$ implies that $(\varphi_1(\vx_1), \cdots, \varphi_1(\vx_T))\notin Z$, we conclude the proof.
\end{proof}

Theorem \ref{thm:main0} is stated for complex sample points. Next we extend it to the real case.

\begin{lemma} \label{lem:coro33} In Lemma \ref{lem:32}, if we assume in addition that $\dim_\sR V_\sR = \dim V$, then the conclusion can be enhanced to ensure that $Z$ is a real subvariety and $\dim_\sR Z < \dim_\sR V_\sR$. 
\end{lemma}

\begin{lemma} \label{lem:lebreal} Let $V \subset \sR^n$ be a (Zariski) closed proper subset, $V \ne \sR^n$. Then $V$ is a null set.
\end{lemma}

The proof of Lemma \ref{lem:lebreal} is the same as that of Lemma \ref{lem:leb}. 

\begin{theorem} \label{thm:main0real} We can additionally assume $\vx_i \in \sR^d$ in Theorem \ref{thm:main0}. 
\end{theorem}

\begin{proof} We verify that $\dim V = \dim_\sR V_\sR$, where $V$ is defined in the proof of Theorem \ref{thm:main0}, but this follows clearly by \citet[Corollary 2.8.2]{bochnak2013real}. We conclude the proof by applying Lemma \ref{lem:coro33}.
\end{proof}

Finally, we apply Theorem \ref{thm:main0real} to two concrete classes of polynomials, namely Examples \ref{ex:nn} and \ref{ex:qux}. For Example \ref{ex:nn}, we construct its variety of tensorization of $\gR_{\gV}$ as follows. We first construct the tensorization of each polynomial. We define 
\begin{align}
    \tW_f = \sum_{i = 1}^r a_i \mqty(1 \\ \vw_i)^{\otimes p_i} \otimes \mqty(1 \\ 0)^{\otimes (p-p_i)}.
\end{align} 
Next we construct the variety of tensorization $W$. Consider the map $\varphi_2 : (\sC^d)^r \to \sC^{(d+1)^p}$,
\begin{align}
    \varphi_2(\vw_1, \cdots, \vw_r) = \sum_{i = 1}^r \mqty(1 \\ \vw_i)^{\otimes p_i} \otimes \mqty(1 \\ 0)^{\otimes (p-p_i)},
\end{align}
and let $Y = \sP(\overline{\Im \varphi_2})$. Similar to $V_i$, we can prove that $Y$ is an irredicuble closed variety defined by homogeneous polynomials with $\dim Y \le dr + 1$. Next consider the map $\varphi'_2 : (\sC^d)^{2r} \to \sC^{(d+1)^p}$, 
\begin{align}
    \varphi'_2(\vw_1, \cdots, \vw_{2r}) = \varphi_2(\vw_1, \cdots, \vw_r) - \varphi_2(\vw_{r+1}, \cdots, \vw_{2r})
\end{align}
and let $W = \sP(\overline{\Im \varphi_2'})$. Similar to $Y$, we can prove that $W$ is an irredicuble closed variety defined by homogeneous polynomials with $\dim W \le 2dr+1$. Together with Theorem \ref{thm:main0real}, we can conlude that the optimal action for Example \ref{ex:nn} can be uniquely determined using at most $2dr+1$ samples.

For Example \ref{ex:qux}, we construct $W$ as follows. Let 
\begin{align*}
    \mU = \mqty(\vw_1 & \cdots & \vw_k), \qquad q = \sum_{I\subseteq [k] : \abs{I} \le p} a_I x^I,
\end{align*}
then we construct the tensorization of each polynomial by 
\begin{align}
    \tW_f = \sum_{I \subseteq [k] : \abs{I} \le p} a_I \bigotimes_{i \in I} \mqty(1 \\ \vw_i) \otimes \mqty(1 \\ 0)^{\otimes (p - \abs{I})}. \label{eq:quxwf}
\end{align}
Then we have $f(\vx) = \langle \tW_f, \tX_\vx\rangle$. To reduce the dimension of $W$ and get better sample complexity bound, we construct in a manner slightly different from what we did for Example \ref{ex:nn}. Consider the map $\varphi_3 : (\sC^d)^k \times \sC^{(k+1)^p} \to \sC^{(d+1)^p}$,
\begin{align}
    (\vw_1, \cdots, \vw_k) \times (a_I : I \subseteq [k], \abs{I} \le p) \mapsto \tW_f, 
\end{align}
where $\tW_f$ is as defined in (\ref{eq:quxwf}). Let $Y = \sP(\overline{\Im \varphi_3})$ and $W = \sP(\overline{\Im \varphi_3 - \Im \varphi_3})$. We end up with $\dim Y =\le dk + (k+1)^p + 1, \dim W \le 2(dk + (k+1)^p) + 1$. So we conlude that the optimal action for Example \ref{ex:qux} can be uniquely determined using at most $2 dk + 2(k+1)^p + 1$ samples.

\section{Omitted Proof for Lower Bounds with UCB Algorithms}


In this section, we provide the proof for the lower bounds for learning with UCB algorithms in Subsection~\ref{sec:noiseless:lb}.

\paragraph{Notation} Recall that we use $\Lambda$ to denote the subset of the $p$-th multi-indices $\Lambda = \{ (\alpha_1,\dots,\alpha_p) | 1 \leq \alpha_1 < \dots < \alpha_p \leq d \}$. For an $\alpha = (\alpha_1,\dots,\alpha_p) \in \Lambda$, denote $\mM_\alpha = \ve_{\alpha_1} \otimes \dots \otimes \ve_{\alpha_p}$, $\mA_\alpha =  (\ve_{\alpha_1} + \dots + \ve_{\alpha_p})^{\otimes p}$. The model space $\cM$ is a subset of rank-1 $p$-th order tensors, which is defined as $\cM =   \Big\{ \mM_\alpha | \alpha \in \Lambda \Big\}$. We define the core action set $\cA_0$ as $\cA_0 = \{ \ve_{\alpha_1} + \dots + \ve_{\alpha_p} | \alpha \in \Lambda \}$. The action set $\cA$ is the convex hull of $\cA_0$: $\cA = \mathrm{conv}(\cA_0)$. 
Assume that the ground-truth parameter is $\mM^* = \mM_{\alpha^*} \in \cM$. At round $t$, the algorithm chooses an action $\va_t \in \cA$, and gets the \textbf{noiseless} reward $r_t = r(\mM^*, \va_t) =  \langle \mM^*, (\va_t)^{\otimes p} \rangle = \prod_{i=1}^p \langle \ve_{\alpha^*_i}, \va_t\rangle$. 

\subsection{Proof for Theorem~\ref{thm:lb_noiseless_ucb}}

We introduce a lemma showing that if the action set is \textbf{restricted} to the core action set $\cA_0$, then at least $ |\cA_0| - 1 = {d \choose p}  - 1$ actions are needed to identify the ground-truth.

\begin{lemma}
\label{lem:restricted:action}
If the actions are restricted to $\cA_0$, then for the noiseless degree-$p$ polynomial bandits, any algorithm needs to play at least ${d \choose p}  - 1 $ actions to determine $\mM^*$ in the worst case. Furthermore, the worst-case cumulative regret at round $T$ can be lower bounded by 
\[
    \mathfrak{R}(T) \geq \min\{ T, {d \choose p}  - 1\}.
\]
\end{lemma}

\begin{proof}[proof of Lemma~\ref{lem:restricted:action}]
For any $\alpha$ and $\alpha'$, the reward of playing $\ve_{\alpha_1} + \dots + \ve_{\alpha_p}$ when the ground-truth model is $\mM_\alpha'$ is 
\begin{align*}
    \langle \mM_\alpha', (\ve_{\alpha_1} + \dots + \ve_{\alpha_p})^{\otimes p} \rangle  &= \prod_{i=1}^p \langle \ve_{\alpha'_i}, \ve_{\alpha_1} + \dots + \ve_{\alpha_p}\rangle \\
    &= \prod_{i=1}^p \indict \{ \alpha'_i \in \alpha \} \\
    &= \left\{
        \begin{array}{ll}
        1, & \text{ if }  \alpha = \alpha ' \\
        0, & \text{ otherwise }.
        \end{array}
    \right.
\end{align*}

Hence, no matter how the algorithm adaptively chooses the actions, in the worst case ${d \choose p} -1$ actions are needed to determine $\mM^*$. 
Also notice that the reward for $\ve_{\alpha_1} + \dots + \ve_{\alpha_p}$ is zero if $\alpha \ne \alpha^*$. Therefore the regret lower bound follows.
\end{proof}

Next, we show that even when the action set is unrestricted, any UCB algorithm fails to explore in an unrestricted way. This is because the optimistic mechanism forbids the algorithm to play an informative action that is known to be low reward for all models in the confidence set. We first recall the definition of UCB algorithms.

\paragraph{UCB Algorithms} The UCB algorithms sequentially maintain a confidence set $\cC_t$ after playing actions $\va_1,\dots, \va_t$. Then UCB algorithms play
\[
    \va_{t+1} \in \argmax_{\va\in \cA} \mathrm{UCB}_t(\va),
\]
where
\[
    \mathrm{UCB}_t(\va) = \max_{\mM \in \cC_t} \langle \mM, (\va)^{\otimes p} \rangle.  
\]

\begin{proof} [proof of Theorem~\ref{thm:lb_noiseless_ucb}]
We prove that even if the action set is unrestrcited, the optimistic mechanism in the UCB algorithm above forces it to choose actions in the restricted action set $\cA_0$.

Assume $\mM^* = \mM_{\alpha^*}$. Next we show that for all $\va \in \cA - \cA_0$ (where the minus sign should be understood as set difference), we have 
\[
    \mathrm{UCB}_t(\va) < 1.
\]
For all $\va \in \cA $,  since $\cA = \mathrm{conv}(\cA_0)$, we can write 
\[
    \va = \sum_{\alpha \in \Lambda} p_\alpha (\ve_{\alpha_1} + \dots + \ve_{\alpha_p}),
\]
where $\sum_{\alpha \in \Lambda } p_{\alpha} = 1$ and $p_{\alpha} \geq 0$. Therefore, 
\begin{align*}
    \mathrm{UCB}_t(\va) &= \max_{\mM \in \cC_t}\langle \mM, (\va)^{\otimes p}  \rangle\\
    &\leq \max_{\mM \in \cM} \langle \mM, (\va)^{\otimes p}\rangle \\
    & = \max_{\alpha'}  \langle \mM_{\alpha'} , (\va)^{\otimes p}  \rangle \\
    & = \max_{\alpha'}  \prod_{i=1}^p  \langle \ve_{\alpha'_i},  \va  \rangle  .
\end{align*}
Plug in the expression of $\va$, we have
\begin{align*}
    \langle \ve_{\alpha'_i}, \va \rangle &= \sum_\alpha p_\alpha  \langle  \ve_{\alpha'_i} ,\ve_{\alpha_1} + \dots + \ve_{\alpha_p} \rangle  \\
    &= \sum_\alpha p_\alpha  \indict{\{\alpha'_i \in \alpha \}} \\
    &\leq  \sum_\alpha p_\alpha = 1.
\end{align*}
Therefore, for any fixed $\alpha' = (\alpha'_1, \dots ,\alpha'_p)$,
\begin{align*}
   \prod_{i=1}^p  \langle  \ve_{\alpha'_i}, \va \rangle  &= 
    \Big( \sum_\alpha p_\alpha  \indict{\{{\alpha'_1}  \in \alpha \}} \Big) \cdots \Big( \sum_\alpha p_\alpha  \indict{\{{\alpha'_p}  \in \alpha \}} \Big) \\
    &\leq 1,
\end{align*}
where the equality holds if and only if for any $p_\alpha > 0$, $\alpha = \alpha'$, which is equivalent to $\va = \ve_{\alpha'_1} + \dots + \ve_{\alpha'_p} $. Therefore, if $\va \in \cA - \cA_0$, for any $\alpha' \in \Lambda$, we have $\prod_{i=1}^p  \langle \ve_{\alpha'_i},  \va  \rangle < 1$. This means
\[
    \mathrm{UCB}_t(\va) < 1.
\]

Meanwhile, we can see that for the action $\va^* = \ve_{\alpha^*_1} + \dots + \ve_{\alpha^*_p} \in \cA_0$, 
\begin{align*}
    \mathrm{UCB}_t(\va^*) &= \max_{\mM \in \cC_t} \langle \mM ,    (\va^*)^{\otimes p} \rangle \\
    & \geq \langle \mM^* ,  (\va^*)^{\otimes p} \rangle  \tag{$\mM^* \in \cC_t$} \\
    & = \langle \mM^* ,  \mA_{\alpha^*} \rangle = 1. 
\end{align*}

Therefore, we see that $(\cA - \cA_0) \cap \argmax_{\va\in \cA} \mathrm{UCB}_t(\va) = \varnothing $, which means $\va_{t+1} \in \cA_0$ for all $t\geq0$. Therefore, by Lemma~\ref{lem:restricted:action}, the theorem holds.

\end{proof}

\subsection{$O(d)$ Actions via Solving Polynomial Equations}

Firstly, we verify that the model falls into the category of Example~\ref{ex:qux} with $k=p$. For every $\alpha \in \Lambda$, the reward of playing $\va$ when the ground-truth model is $\mM_\alpha$ is 
\[
    \langle \mM_\alpha, (\va)^{\otimes p} \rangle = \prod_{i=1}^p \langle \ve_{\alpha_i}, \va\rangle,
\]
which can be written as $q_0(\mU_\alpha \va)$, where $q_0(x_1,\dots, x_p) = x_1  x_2 \cdots x_p$ and $\mU_\alpha \in \mathbb{R}^{p \times d}$ is a matrix with $\ve_{\alpha_i}$ as the $i$-th row. 

Secondly, we show that since the ground-truth model is $p$-homogenous, we can extend the action set to $ \mathrm{conv}(\gA, \mathbf{0})$. This is because for every action of the form $ c \va$, where $0\leq c \leq 1$ and $\va \in \gA$, the reward is $c^p$ times the reward at $\va$. Therefore, to get the reward at $c\va$, we only need to play at $\va$ and multiply the reward by $c^p$.  

Notice that $\mathrm{conv}(\gA, \mathbf{0})$ is of positive Lebesgue measure. By Theorem~\ref{thm:poly_noiseless}, we know that only $2(dk+(p+1)^p) = O(d)$ actions are needed to determine the optimal action almost surely. 

\section{Proof of Section~\ref{sec:lb}}\label{sec:proof_lb}

We present the proof of Theorem~\ref{thm:lb_asym_tensor} in the following.
\begin{proof}
We overload the notation and use $[d]$ to denote the set $\{e_1,e_2,\dots,e_d\}$. The hard instances are chosen in $\Delta \cdot [d]^p$, i.e. $(\vtheta_1,\dots,\vtheta_p) = \Delta \cdot (\hat \vtheta_1,\dots,\hat \vtheta_p)$ where $(\hat \vtheta_1,\dots,\hat \vtheta_p) \in [d]^p$. For a group of vectors $\vtheta_1,\dots,\vtheta_p \in [d]$, we use
$$\supp(\vtheta_1,\dots,\vtheta_p) := (\max_{i\in [p]} (\vtheta_i)_1,\dots, \max_{i\in [p]} (\vtheta_i)_d) \in \{0,1\}^d$$
to denote the support of these vectors. We use $\va^{(t)} \in \R^{d}$ to denote the action in $t$-th episode.

We use $\PP_{(\vtheta_1,\dots,\vtheta_p)}$ to denote the measure on outcomes induced by the interaction of the fixed policy and the bandit paramterised by $r = \prod_{i=1}^p (\vtheta_i^\top \va) + \epsilon$. Specifically, We use $\PP_{0}$ to denote the measure on outcomes induced by the interaction of the fixed policy and the pure noise bandit $r = \epsilon$.
\begin{align*}
    &~ \mathfrak{R}(d,p,T)\\
    \geq &~ \frac{1}{d^p}\sum_{(\vtheta_1,\dots,\vtheta_p) \in \Delta \cdot [d]^p} \E_{(\vtheta_1,\dots,\vtheta_p)}\left[ T\Delta^p/p^{p/2} - \sum_{t=1}^T \prod_{i=1}^p (\vtheta_i^\top \va^{(t)}) \right]\\
    = &~ \frac{\Delta^p}{d^p}\sum_{(\vtheta_1,\dots,\vtheta_p) \in \Delta \cdot[d]^p} \left( T/p^{p/2} - \E_{(\vtheta_1,\dots,\vtheta_p)} \left[\sum_{t=1}^T \prod_{i=1}^p (\hat \vtheta_i^\top \va^{(t)})\right] \right)\\
    \geq &~  \frac{\Delta^p}{d^p}\sum_{(\vtheta_1,\dots,\vtheta_p) \in \Delta \cdot [d]^p} \left( T/p^{p/2} - \E_{0} \left[\sum_{t=1}^T \prod_{i=1}^p (\hat \vtheta_i^\top \va^{(t)})\right] - T \|\PP_{0}-\PP_{(\vtheta_1,\dots,\vtheta_p)}\|_{\mathrm{TV}} \right)\\
    \geq &~  \frac{\Delta^p}{d^p}\sum_{(\vtheta_1,\dots,\vtheta_p) \in \Delta \cdot [d]^p} \left( T/p^{p/2} - \E_{0} \left[\sum_{t=1}^T \prod_{i=1}^p (\hat \vtheta_i^\top \va^{(t)})\right] - T\sqrt{\KL(\PP_{0}||\PP_{(\vtheta_1,\dots,\vtheta_p)})} \right)\\
    \geq &~  \frac{\Delta^p}{d^p}\sum_{(\vtheta_1,\dots,\vtheta_p) \in\Delta \cdot [d]^p} \left( T/p^{p/2} - \E_{0} \left[\sum_{t=1}^T \prod_{i=1}^p (\hat \vtheta_i^\top \va^{(t)})\right] - T\sqrt{\Delta^{2p}\E_{0} \left[\sum_{t=1}^T \prod_{i=1}^p (\hat \vtheta_i^\top \va^{(t)})^2\right]} \right)\\
    \geq &~  \frac{\Delta^p}{d^p} \left( \frac{d^pT}{p^{\frac{p}{2}}} - \E_{0} \left[\sum_{(\hat \vtheta_1,\dots,\hat \vtheta_p) \in [d]^p}\sum_{t=1}^T \prod_{i=1}^p (\hat \vtheta_i^\top \va^{(t)})\right] - Td^{\frac{p}{2}}\Delta^{p}\sqrt{\E_{0} \left[\sum_{(\hat \vtheta_1,\dots,\hat \vtheta_p) \in [d]^p}\sum_{t=1}^T \prod_{i=1}^p (\hat \vtheta_i^\top \va^{(t)})^2\right]} \right)\\
\end{align*}
where the first step comes from $$\mathrm{Regret} \geq \E_{(\vtheta_1,\dots,\vtheta_p)} \left[T\Delta^p/p^{p/2} - \sum_{t=1}^T \prod_{i=1}^p (\vtheta_i^\top \va^{(t)})\right] $$ (the optimal action in hindsight is $\va = \supp(\vtheta_1,\dots,\vtheta_p)/\sqrt{p}$); the second step comes from $(\vtheta_1,\dots,\vtheta_p) = \Delta \cdot (\hat \vtheta_1,\dots,\hat \vtheta_p)$ and algebra; the third step comes from $\left|\sum_{t=1}^T \prod_{i=1}^p (\hat \vtheta_i^\top \va^{(t)})\right| \leq T$; the fourth step comes from Pinsker's inequality; the fifth step comes from 
\begin{align*}
    \KL(\PP_0||\PP_{\vtheta_1,\dots,\vtheta_p}) = &~ \E_0\left[\sum_{t=1}^T \KL\left(N(0,1)||N(\prod_{i=1}^p (\vtheta_i^\top \va^{(t)}),1)\right)\right]\\
    =&~ \Delta^{2p}\E_{0} \left[\sum_{t=1}^T \prod_{i=1}^p (\hat \vtheta_i^\top \va^{(t)})^2\right]
\end{align*}
and the final step comes from Jensen's inequality and algebra.

Notice that 
\begin{align*}
    \E_{0} \left[\sum_{(\hat \vtheta_1,\dots,\hat \vtheta_p) \in [d]^p}\sum_{t=1}^T \prod_{i=1}^p (\hat \vtheta_i^\top \va^{(t)})\right] = &~ \E_{0} \left[\sum_{(j_1,\dots,j_p) \in [d]^p}\sum_{t=1}^T \prod_{i=1}^p (\va^{(t)}_{j_i})\right]\\
    = &~ \E_{0} \left[\sum_{t=1}^T \prod_{i=1}^p (\sum_{j= 1}^d \va^{(t)}_j)\right]\\
    \leq &~ \E_{0} \left[\sum_{t=1}^T \prod_{i=1}^p \| \va^{(t)}\|_1\right]\\
    \leq &~ d^{p/2}T
\end{align*}
and
\begin{align*}
    \E_{0} \left[\sum_{(\hat \vtheta_1,\dots,\hat \vtheta_p) \in [d]^p}\sum_{t=1}^T \prod_{i=1}^p (\hat \vtheta_i^\top \va^{(t)})^2\right] = &~ \E_{0} \left[\sum_{(j_1,\dots,j_p) \in [d]^p}\sum_{t=1}^T \prod_{i=1}^p (\va^{(t)}_{j_i})^2\right]\\
    = &~ \E_{0} \left[\sum_{t=1}^T \prod_{i=1}^p \|\va^{(t)}\|_2^2\right]\\
    \leq &~ T
\end{align*}
where we used $\|\va^{(t)}\|_2 \leq 1, \forall t \in [T]$.
Therefore plugging back we have
\begin{align*}
    \mathfrak{R}(d,p,T) \geq &~\frac{\Delta^p}{d^p} \left( \frac{d^pT}{p^{\frac{p}{2}}} - d^{\frac{p}{2}}T - Td^{\frac{p}{2}}\Delta^{p}\sqrt{T} \right)
\end{align*}
and finally letting $\Delta^p = \sqrt{\frac{d^p}{4Tp^p}}$ leads to
\begin{align*}
    \mathfrak{R}(d,p,T) \geq &~O(\sqrt{d^pT}/p^{p}).
\end{align*}
\end{proof}

\begin{remark}
Better result $O(\sqrt{d^pT})$ holds for bandits $r = \prod_{i=1}^p (\vtheta_i^\top \va_i) + \epsilon$ where $\va_i \in \R^d, \|\va_i \|_2 \leq 1$.
\end{remark}

For completeness, we show the proof of the above remark.
\begin{proof}
We overload the notation and use $[d]$ to denote the set $\{e_1,e_2,\dots,e_d\}$. The hard instances are chosen in $\Delta \cdot [d]^p$, i.e. $(\vtheta_1,\dots,\vtheta_p) = \Delta \cdot (\hat \vtheta_1,\dots,\hat \vtheta_p)$ where $(\hat \vtheta_1,\dots,\hat \vtheta_p) \in [d]^p$.
We use $\va^{(t)}_i \in \R^{d}$ to denote the $i$-th action in $t$-th episode, where $i \in [p], t \in [T]$.

We use $\PP_{(\vtheta_1,\dots,\vtheta_p)}$ to indicate the measure on outcomes induced by the interaction of the fixed policy and the bandit paramterised by $r = \prod_{i=1}^p (\vtheta_i^\top \va_i) + \epsilon$. Specifically, We use $\PP_{0}$ to indicate the measure on outcomes induced by the interaction of the fixed policy and the pure noise bandit $r = \epsilon$.
\begin{align*}
    &~ \mathfrak{R}(d,p,T)\\
    \geq &~ \frac{1}{d^p}\sum_{(\vtheta_1,\dots,\vtheta_p) \in \Delta \cdot [d]^p} \E_{(\vtheta_1,\dots,\vtheta_p)}\left[ T\Delta^p - \sum_{t=1}^T \prod_{i=1}^p (\vtheta_i^\top \va_i^{(t)}) \right]\\
    = &~ \frac{\Delta^p}{d^p}\sum_{(\vtheta_1,\dots,\vtheta_p) \in \Delta \cdot[d]^p} \left( T - \E_{(\vtheta_1,\dots,\vtheta_p)} \left[\sum_{t=1}^T \prod_{i=1}^p (\hat \vtheta_i^\top \va^{(t)}_i)\right] \right)\\
    \geq &~  \frac{\Delta^p}{d^p}\sum_{(\vtheta_1,\dots,\vtheta_p) \in \Delta \cdot [d]^p} \left( T - \E_{0} \left[\sum_{t=1}^T \prod_{i=1}^p (\hat \vtheta_i^\top \va^{(t)}_i)\right] - T \|\PP_{0}-\PP_{(\vtheta_1,\dots,\vtheta_p)}\|_{\mathrm{TV}} \right)\\
    \geq &~  \frac{\Delta^p}{d^p}\sum_{(\vtheta_1,\dots,\vtheta_p) \in \Delta \cdot [d]^p} \left( T - \E_{0} \left[\sum_{t=1}^T \prod_{i=1}^p (\hat \vtheta_i^\top \va^{(t)}_i)\right] - T\sqrt{\KL(\PP_{0}||\PP_{(\vtheta_1,\dots,\vtheta_p)})} \right)\\
    \geq &~  \frac{\Delta^p}{d^p}\sum_{(\vtheta_1,\dots,\vtheta_p) \in\Delta \cdot [d]^p} \left( T - \E_{0} \left[\sum_{t=1}^T \prod_{i=1}^p (\hat \vtheta_i^\top \va^{(t)}_i)\right] - T\sqrt{\Delta^{2p}\E_{0} \left[\sum_{t=1}^T \prod_{i=1}^p (\hat \vtheta_i^\top \va^{(t)}_i)^2\right]} \right)\\
    \geq &~  \frac{\Delta^p}{d^p} \left( {d^pT} - \E_{0} \left[\sum_{(\hat \vtheta_1,\dots,\hat \vtheta_p) \in [d]^p}\sum_{t=1}^T \prod_{i=1}^p (\hat \vtheta_i^\top \va^{(t)}_i)\right] - Td^{\frac{p}{2}}\Delta^{p}\sqrt{\E_{0} \left[\sum_{(\hat \vtheta_1,\dots,\hat \vtheta_p) \in [d]^p}\sum_{t=1}^T \prod_{i=1}^p (\hat \vtheta_i^\top \va^{(t)}_i)^2\right]} \right)\\
\end{align*}
where the first step comes from $$\mathrm{Regret} \geq \E_{(\vtheta_1,\dots,\vtheta_p)} \left[T\Delta^p - \sum_{t=1}^T \prod_{i=1}^p (\vtheta_i^\top \va^{(t)}_i)\right] $$ (the optimal action in hindsight is $\va_i = \hat \vtheta_i$); the second step comes from $(\vtheta_1,\dots,\vtheta_p) = \Delta \cdot (\hat \vtheta_1,\dots,\hat \vtheta_p)$ and algebra; the third step comes from $\left|\sum_{t=1}^T \prod_{i=1}^p (\hat \vtheta_i^\top \va^{(t)}_i)\right| \leq T$; the fourth step comes from Pinsker's inequality; the fifth step comes from 
\begin{align*}
    \KL(\PP_0||\PP_{\vtheta_1,\dots,\vtheta_p}) = &~ \E_0\left[\sum_{t=1}^T \KL\left(N(0,1)||N(\prod_{i=1}^p (\vtheta_i^\top \va^{(t)}_i),1)\right)\right]\\
    =&~ \Delta^{2p}\E_{0} \left[\sum_{t=1}^T \prod_{i=1}^p (\hat \vtheta_i^\top \va^{(t)}_i)^2\right]
\end{align*}
and the final step comes from Jensen's inequality and algebra.

Notice that 
\begin{align*}
    \E_{0} \left[\sum_{(\hat \vtheta_1,\dots,\hat \vtheta_p) \in [d]^p}\sum_{t=1}^T \prod_{i=1}^p (\hat \vtheta_i^\top \va^{(t)}_i)\right] = &~ \E_{0} \left[\sum_{(j_1,\dots,j_p) \in [d]^p}\sum_{t=1}^T \prod_{i=1}^p \left((\va^{(t)}_i)_{j_i}\right)\right]\\
    = &~ \E_{0} \left[\sum_{t=1}^T \prod_{i=1}^p \left(\sum_{j= 1}^d (\va^{(t)}_i)_j\right)\right]\\
    \leq &~ \E_{0} \left[\sum_{t=1}^T \prod_{i=1}^p \| \va^{(t)}_i\|_1\right]\\
    \leq &~ d^{p/2}T
\end{align*}
and
\begin{align*}
    \E_{0} \left[\sum_{(\hat \vtheta_1,\dots,\hat \vtheta_p) \in [d]^p}\sum_{t=1}^T \prod_{i=1}^p (\hat \vtheta_i^\top \va^{(t)}_i)^2\right] = &~ \E_{0} \left[\sum_{(j_1,\dots,j_p) \in [d]^p}\sum_{t=1}^T \prod_{i=1}^p \left( (\va^{(t)}_i)_{j_i}\right)^2\right]\\
    = &~ \E_{0} \left[\sum_{t=1}^T \prod_{i=1}^p \|\va^{(t)}_i\|_2^2\right]\\
    \leq &~ T
\end{align*}
where we used $\|\va^{(t)}_i\|_2 \leq 1, \forall t \in [T]$.
Therefore plugging back we have
\begin{align*}
    \mathfrak{R}(d,p,T) \geq &~\frac{\Delta^p}{d^p} \left( {d^pT} - d^{\frac{p}{2}}T - Td^{\frac{p}{2}}\Delta^{p}\sqrt{T} \right)
\end{align*}
and finally letting $\Delta^p = \sqrt{\frac{d^p}{4T}}$ leads to
\begin{align*}
    \mathfrak{R}(d,p,T) \geq &~O(\sqrt{d^pT}).
\end{align*}
\end{proof}
We present the proof of Theorem~\ref{thm:lb_gap_dep} in the following.
\begin{proof}
Denote the optimal action in hindsight as $\va^\ast = \supp(\vtheta_1,\dots,\vtheta_p)/\sqrt{p}$. 
From the proof of Theorem~\ref{thm:lb_asym_tensor} we know that if $T \leq \frac{1}{4p^p}\cdot\frac{d^p}{\Delta^{2p}}$, then
\begin{align*}
    &~ \frac{1}{d^p}\sum_{(\vtheta_1,\dots,\vtheta_p) \in \Delta \cdot [d]^p}\E_{(\vtheta_1,\dots,\vtheta_p)}\left[ \prod_{i=1}^p (\vtheta_i^\top \va^\ast) -  \prod_{i=1}^p (\vtheta_i^\top \va^{(t)}) \right]\\
    \geq &~ \frac{\Delta^p}{d^p} \left( \frac{d^p}{p^{\frac{p}{2}}} - d^{\frac{p}{2}} - d^{\frac{p}{2}}\Delta^{p}\sqrt{T} \right)\\
    \geq &~ \frac{\Delta^p}{4p^{\frac{p}{2}}}\\
    \geq &~ \frac{1}{4}\cdot\frac{1}{d^p}\sum_{(\vtheta_1,\dots,\vtheta_p) \in \Delta \cdot [d]^p}\E_{(\vtheta_1,\dots,\vtheta_p)}\left[ \prod_{i=1}^p (\vtheta_i^\top \va^\ast) \right]
\end{align*}
which indicates the following 
\begin{align*}
    \inf_{\pi} \sup_{(\vtheta_1,\dots,\vtheta_p)} \E_{(\vtheta_1,\dots,\vtheta_p)}\left[ \frac{3}{4}\cdot \prod_{i=1}^p (\vtheta_i^\top \va^\ast) -  \prod_{i=1}^p (\vtheta_i^\top \va^{(t)}) \right] \geq 0.
\end{align*}
\end{proof}


\end{document}